\documentclass[conference]{IEEEtran}
\usepackage{url}
\usepackage{custom_math}
\usepackage{lipsum}
\usepackage[hidelinks]{hyperref}
\usepackage{cleveref}
\usepackage{booktabs}
\usepackage{graphicx}
\usepackage{xcolor}

\usepackage{enumitem}
\usepackage{amssymb}
\usepackage{algorithm,algpseudocode}
\usepackage[caption=false,font=footnotesize]{subfig}
\usepackage{caption}
\usepackage{multirow}
\usepackage{listings}
\usepackage{stackengine}
\usepackage{placeins}
\usepackage{microtype}
\usepackage{todonotes}

\def\BibTeX{{\rm B\kern-.05em{\sc i\kern-.025em b}\kern-.08em
    T\kern-.1667em\lower.7ex\hbox{E}\kern-.125emX}}

\renewcommand\paragraph[2][.]{\noindent{\textbf{#2#1}}}

\newcommand{\bx}{\mat{x}}

\newcommand{\bz}{\mat{z}}
\newcommand{\bdelta}{\boldsymbol{\delta}}
\newcommand{\btau}{\boldsymbol{\tau}}

\newcommand{\PP}{\mathbb{P}}
\newcommand{\XX}{\mathbb{X}}
\newcommand{\CC}{\mathbb{C}}

\newcommand{\ProcedureDist}{\mathcal{P}}

\newcommand{\Attack}{\mathcal{A}}
\newcommand{\AttackDB}{\mathbb{A}}

\newcommand{\TransMat}{A}
\newcommand{\EmissionProb}{\boldsymbol{\phi}}
\newcommand{\FP}{\mathtt{F}}
\newcommand{\FF}{\mathbb{F}}
\newcommand{\Metric}{\mathcal{M}}

\newcommand{\PROC}[1]{\texttt{#1}}
\newcommand{\SEA}{\texttt{SEA}}

\begin{document}

\author{\IEEEauthorblockN{Yue Gao}
\IEEEauthorblockA{\textit{University of Wisconsin--Madison}\\
Madison, USA \\
gy@cs.wisc.edu}
\and
\IEEEauthorblockN{Ilia Shumailov}
\IEEEauthorblockA{\textit{University of Oxford}\\
Oxford, UK\\
ilia.shumailov@chch.ox.ac.uk}
\and
\IEEEauthorblockN{Kassem Fawaz}
\IEEEauthorblockA{\textit{University of Wisconsin--Madison}\\
Madison, USA \\
kfawaz@wisc.edu}
}

\title{\texttt{SEA}: Shareable and Explainable Attribution for \\ Query-based Black-box Attacks}

\maketitle

\begin{abstract}
Machine Learning (ML) systems are vulnerable to adversarial examples, particularly those from query-based black-box attacks. Despite various efforts to detect and prevent such attacks, ML systems are still at risk, demanding a more comprehensive approach to security that includes logging, analyzing, and sharing evidence. While traditional security benefits from well-established practices of forensics and threat intelligence sharing, ML security has yet to find a way to profile its attackers and share information about them. In response, this paper introduces \texttt{SEA}, a novel ML security system to characterize black-box attacks on ML systems for forensic purposes and to facilitate human-explainable intelligence sharing. \texttt{SEA} leverages Hidden Markov Models to attribute the observed query sequence to known attacks. It thus understands the attack's progression rather than focusing solely on the final adversarial examples. Our evaluations reveal that \texttt{SEA} is effective at attack attribution, even on the second incident, and is robust to adaptive strategies designed to evade forensic analysis. \texttt{SEA}'s explanations of the attack's behavior allow us even to fingerprint specific minor bugs in widely used attack libraries. For example, we discover that the SignOPT and Square attacks in ART v1.14 send over 50\% duplicated queries. We thoroughly evaluate \texttt{SEA} on a variety of settings and demonstrate that it can recognize the same attack with more than 90\% Top-1 and 95\% Top-3 accuracy. Finally, we demonstrate how \texttt{SEA} generalizes to other domains like text classification.
\end{abstract}


\section{Introduction}
\label{sec:intro}

Machine learning (ML) systems are vulnerable to adversarial examples, where imperceptible perturbations to the input data can change its prediction~\cite{advex-1,advex-2}. Among the existing attacks, query-based black-box attacks are particularly concerning. To generate an adversarial example, they only need to interact with the model through a sequence of queries without any knowledge of its architecture or gradients~\cite{hsj,geoda,square,signopt}. Despite substantial efforts to detect~\cite{stateful-detection,blacklight} and prevent~\cite{noise-empirical,noise-theoretical,post-process} such attacks, logging, analyzing, and sharing the full evidence of the attack remains under-explored. This gap motivates a critical question: \textbf{what can we do when an attack has bypassed all detection and protection mechanisms?}

Traditional security addresses this gap at the \textbf{respond} stage in the NIST cybersecurity framework, including \textit{forensics}, i.e., retrospective analysis of attack incidents to aid future detection and recovery~\cite{forensics1,forensics2,forensics3}, and \textit{intelligence sharing}, i.e., the dissemination of attack incident details to impede future attack success in different contexts~\cite{share1,share2}. Such security mechanisms come into play \textit{after realizing a breach}, helping to retrospectively analyze how the attacker succeeded. While traditional security benefits from well-established techniques like public CVE trackers~\cite{cve}, CERT systems~\cite{cert}, and open attack frameworks like Metasploit, the study of forensics and intelligence sharing is limited in the context of ML security.

Existing research primarily focuses on the adversarial attribution problem~\cite{attr-multitask}, which seeks to understand the hidden signatures in adversarial examples, often in the context of less concerning white-box attacks~\cite{pgd,cw}. These works model the attribution problem as a deep learning classification task, attempting to extract all information of the attacker from a single adversarial example~\cite{attr-multitask,attr-whitebox,attr-ssl,attr-fingerprint}. However, this approach requires vast amounts of adversarial examples, meaning that the same attack must happen thousands of times before it can be characterized for attribution. Furthermore, these studies overlook a critical aspect of forensic analysis, namely, that attribution outcomes must be shareable and explainable. The need to distribute large models and signatures for knowledge sharing is neither practical nor efficient.

\begin{figure*}[t]
    \centering
    \includegraphics[width=\linewidth]{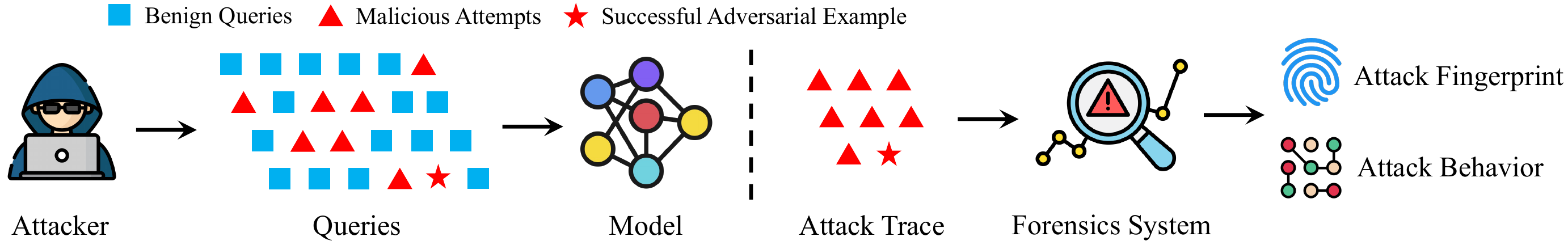}
    \caption{The general scenario for our forensic system. \textbf{Left:} The attacker sends malicious queries to the model to construct an adversarial example. \textbf{Right:} After the attack has triggered a security alarm, the forensic system extracts the attack's trace to attribute, explain, and share the attack's behavior.}
    \label{fig:demo:pipeline}
\end{figure*}

In this paper, we introduce \SEA{}, a novel ML security system designed to analyze black-box attacks against ML systems for forensic purposes and enable human-explainable intelligence sharing. \SEA{} can characterize an attack on its \emph{first} incident and accurately recognize subsequent incidents of the same attack. In particular, the attribution it provides can be easily interpreted and understood by both machines and humans. It thus fills the long missing respond stage in the NIST cybersecurity framework for ML security. \Cref{fig:demo:pipeline} illustrates its pipeline.

\SEA{} achieves these objectives by examining the \textit{progression} of an attack incident, i.e., the sequence of queries extracted from logged history queries, rather than merely the final adversarial example. By analyzing the attack's progression, we can describe its behavior at a finer granularity. As we will show in \Cref{sec:hmm:pattern}, the attack's most noticeable identifiable information is embedded within its query sequence. Building upon this intuition, we model black-box attacks using a unified framework based on Hidden Markov Models (HMMs). This model allows analysts to attribute the observed query sequence to internal attack procedures. Leveraging the well-established theoretical framework of HMMs, \SEA{} achieves data-efficient, explainable, and shareable attack fingerprints.

We thoroughly evaluate \SEA{} and demonstrate its ability to characterize and identify a variety of state-of-the-art attacks. In particular, we show that:
\begin{itemize}[leftmargin=*]
    \item \SEA{} can characterize an attack's behavior and produce a fingerprint right after its \emph{first} incident. The fingerprint is a small easy-to-share matrix that describes the attack's overall programmatic behavior during its progression.
    \item \SEA{} can recognize the same attack's \emph{second} incident with more than 90\% Top-1 and 95\% Top-3 accuracy on multiple image datasets and models. The performance can improve further when more incidents of that attack are observed.
    \item \SEA{} can generalize to other domains like text classification. Even if existing textual attacks are similar, \SEA{} can recognize them with over 70\% Top-1 and 90\% Top-2 accuracy.
    \item \SEA{} can explain the attack's behavior at a per-query granularity. This explainability leads us to discover implementation bugs in attack libraries. For example, we discover that the \texttt{SignOPT-2} and \texttt{Square-2} attacks implemented by the \texttt{ART} library (v1.14)~\cite{art} send over 50\% duplicated queries, making it easy to distinguish from other libraries.
    \item \SEA{} is robust to several adaptive attacks that aim to evade forensic analysis. While employing adaptive strategies has changed the nature of the original attack, \SEA{} can treat them as different attacks and eventually fingerprint and recognize such strategies with a high accuracy.
\end{itemize}

\section{Background and Related Work}
\label{sec:background}

In this section, we introduce the background and related work on black-box adversarial examples and forensics.

\subsection{Black-box Adversarial Examples}
\label{sec:background:adv}

Given a clean input $\bx\in\DD\coloneqq[0,1]^d$ and a classifier $f$, the adversarial example $\bx^\prime$ is visually similar to $\bx$ but misclassified, i.e., $f(\bx^\prime)\neq f(\bx)$~\cite{advex-1,advex-2}. Most attacks achieve visual similarity by bounding the perturbation with a budget $\epsilon$ in \LL{p} norm, where $\norm{\bx - \bx^\prime}_p\leq\epsilon$. We use boldface to denote vectors in this paper.

\paragraph{Black-box Attacks}
A black-box attack $\Attack$ only has query access to the classifier without knowledge of the classifier's architecture or gradients. To generate an adversarial example $\bx^\prime$, the attack interacts with the classifier through a sequence of $n$ queries $X=\s{\bx_t}_{t=1}^n$, where $\bx_n$ is the final adversarial example. We refer to this query sequence as the \textit{attack's trace} or an \textit{incident} of the attack.

Existing query-based black-box attacks can be categorized into \emph{score-based} and \emph{decision-based} attacks based on their assumptions of the classifier's outputs. Score-based attacks assume the classifier returns the confidence score for each query, which facilitates estimation of the classifier's gradients~\cite{nes,eco,square,zoo,simba}. Decision-based attacks assume the classifier returns only the predicted label. In this challenging yet more practical setting, the attacks usually walk near the decision boundary~\cite{hsj,geoda,boundary,rays,qeba} to find the minimally perturbed adversarial example or optimize for a particular objective function~\cite{opt,signopt,policy}.

With additional assumptions of the classifier's architecture or data distribution, query-free black-box attacks can avoid repeatedly querying the model by leveraging the transferability of adversarial examples~\cite{transfer-attack}. Such attacks train a substitute model approximating the victim classifier's decision boundary and mainly attempt to improve transferability~\cite{transfer1,transfer2,transfer3}. Detailed analysis of query-free attacks remains challenging as they can be reduced to the white-box threat model~\cite{pgd,cw} with full access to the classifier's architecture and gradients.

In this paper, we cover a wide range of recent attacks, including 3 score-based attacks~\cite{square,eco,nes} and 5 decision-based attacks~\cite{hsj,geoda,signopt,boundary,rays}. We also include the \LL{2} and \LL{\infty} variants of each attack if it supports and adopts different strategies, totaling 11 distinct attack variants. We explain their details in \Cref{sec:exp:setup,app:exp:attack}.

\paragraph{Black-box Defenses}
Defenses against black-box attacks fall into three broad categories: detection-based, pre-processing, and post-processing defenses. 

Detection defenses identify queries generated by black-box attacks and reject or return incorrect outputs. SD~\cite{stateful-detection} inspects each query's mean pairwise distance to each account's history queries. PRADA~\cite{prada} inspects the query's distributional shifts among different accounts; it was originally designed for model stealing attacks~\cite{model-stealing} but is effective for evasion attacks~\cite{stateful-detection}. Blacklight~\cite{blacklight} detects overly similar queries in the input space. It introduces a hash function that is sensitive to small changes in the image using a segment-based probabilistic fingerprint.

Pre-processing defenses~\cite{random-rescaling,input-transformation,bart,diffpure} aim to increase the cost of finding adversarial examples with randomized input transformations. While their efficacy in the white-box setting is limited both empirically~\cite{bpda,adaptive,demystify} and theoretically~\cite{limitations-preprocessing}, they remain effective in the black-box setting~\cite{noise-empirical,noise-theoretical,boundary-analysis}.
Post-processing defenses modify the model outputs to interfere with the attack~\cite{post-process,advmind}.

Our forensic system \textbf{\SEA{} is complementary to the above defenses}. For example, upon receiving a report about an active black-box attack (at its first few queries), instead of immediately blocking the attack, the defense can turn into a honeypot and allow the attack to proceed. Then, our system would analyze the attack's behavior to expose more information about the attacker. In some cases, identifying attacks may be more valuable than immediate blocking, as it allows for strategic responses.

\subsection{Digital Forensics}

Forensics offers a complementary approach for investigating an attack after it has occurred, especially when existing defenses do not work. They can uncover valuable information about the attacker, such as their identity and unique behaviors.

\paragraph{Forensics Research in Machine Learning}
Existing forensics research in ML primarily focuses on the adversarial attribution problem (AAP)~\cite{attr-multitask}, which aims to understand the hidden signatures in adversarial examples, such as the attack algorithm and hyper-parameters~\cite{attr-multitask,attr-whitebox,attr-ssl,attr-fingerprint}. While our work tackles a similar problem, it offers significant contributions beyond existing work:

\begin{itemize}[leftmargin=*]
	\item \textbf{Explainability.} Prior works cannot provide explainability for their attribution results. Yet, explaining the attribution results is critical in forensics, as it helps human analysts understand and explain the attack's behavior. A standalone attribution only provides limited information in practice. We leverage our modeling of black-box attacks to explain the attack's behavior at the per-query granularity.
	\item \textbf{Sharing.} Prior works model the attribution problem as a deep-learning classification task. To share the knowledge of an attack, they have first to collect vast amounts of data (i.e., the same attack is observed thousands of times) and then distribute the deep learning model. This scenario is neither practical nor efficient. Our system fingerprints the attack on its first incident and can immediately recognize its subsequent incidents with high accuracy.
\end{itemize}

\paragraph{Distinction from Black-box Defenses}
As complementary approaches, forensics and defenses share aspects of the threat model: the attacker aims to compromise the system, and the victim aims to prevent or investigate the attack. However, there is a critical distinction in their threat models that provide an advantage to the forensic system.

The adversarial machine learning literature has shown that designing robust defenses against adversarial examples is challenging both empirically and theoretically~\cite{adaptive,bpda,limitations-preprocessing}. The root cause of this problem, like other security problems, is that the attack can always adapt its technique to the deployed defense until a successful circumvention~\cite{adaptive}.

However, when it comes to forensics, although the attack can still adapt its strategy~\cite{anti-forensics-1,anti-forensics-2,anti-forensics-3}, it is the forensic system that makes the ultimate adaptation~\cite{poison-forensics}. As long as the attack has left a trace, which cannot be changed later, the forensic system can keep adapting its techniques until a successful investigation. In contrast, defenses cannot adapt after the attack has succeeded. \emph{As a result, a forensic system is only broken when it cannot recover any meaningful information about the attack from the trace even after adapting to the attack.} Moreover, actively adapting to a forensic system usually comes with a higher cost, or requires a full system compromise to alter the system logs. We evaluate several adaptive attacks in \Cref{sec:exp:adaptive} and show they cannot prevent \SEA{} from analyzing their traces.

\section{Evasion Forensics}
\label{sec:overview}

\begin{figure*}[t]
    \centering
    \subfloat[\texttt{HSJ-2} Queries ($\bx_t$)]{\includegraphics[width=0.32\textwidth]{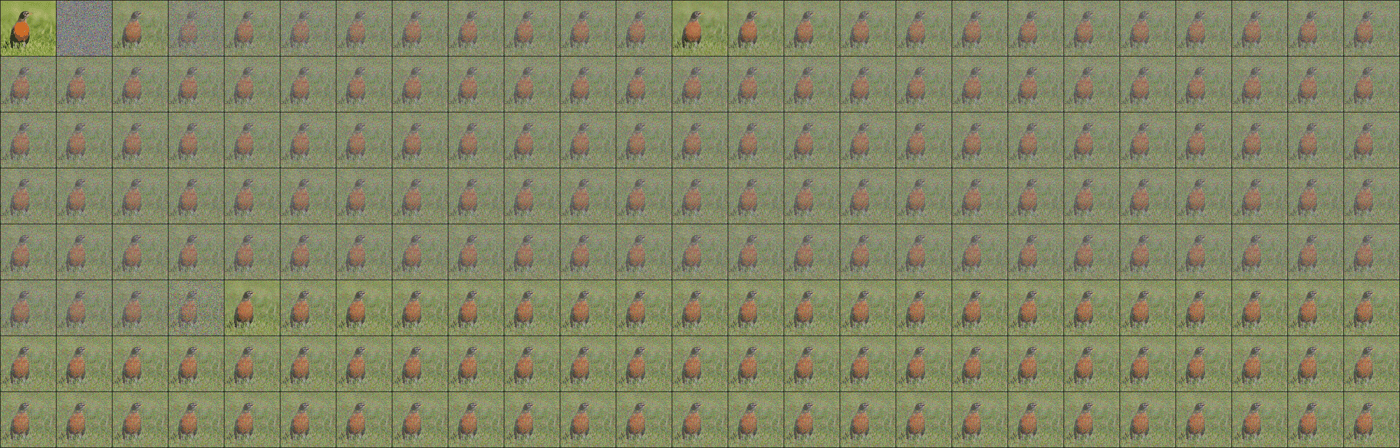}\label{fig:demo:hsj-raw}}
    \hfill
    \subfloat[\texttt{HSJ-2} Steps ($\bx_t-\bx_{t-1}$)]{\includegraphics[width=0.32\textwidth]{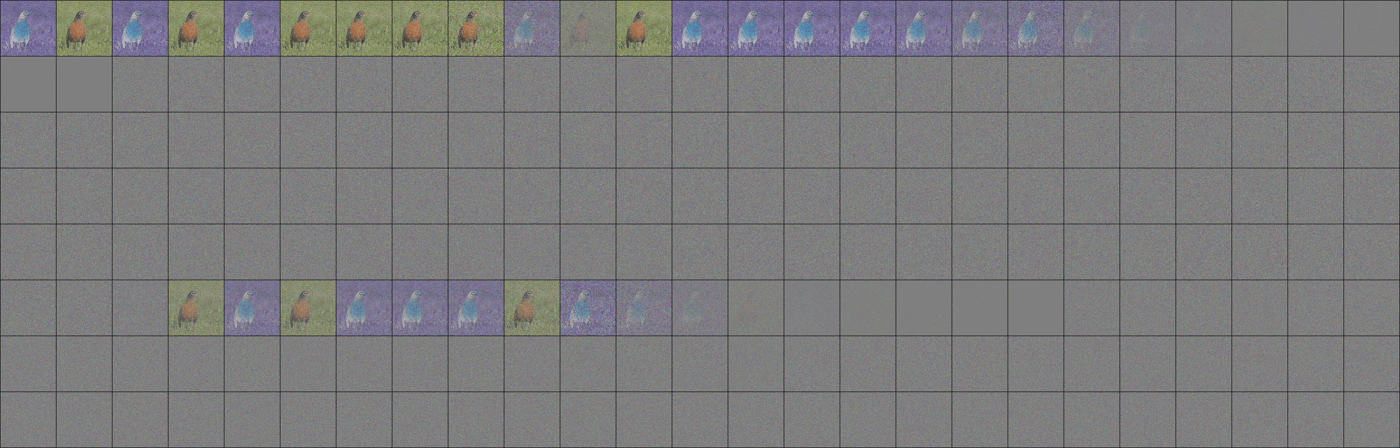}\label{fig:demo:hsj-diff}}
    \hfill
    \subfloat[\texttt{HSJ-2} Steps in Spectrum ($\bx_t-\bx_{t-1}$)]{\includegraphics[width=0.32\textwidth]{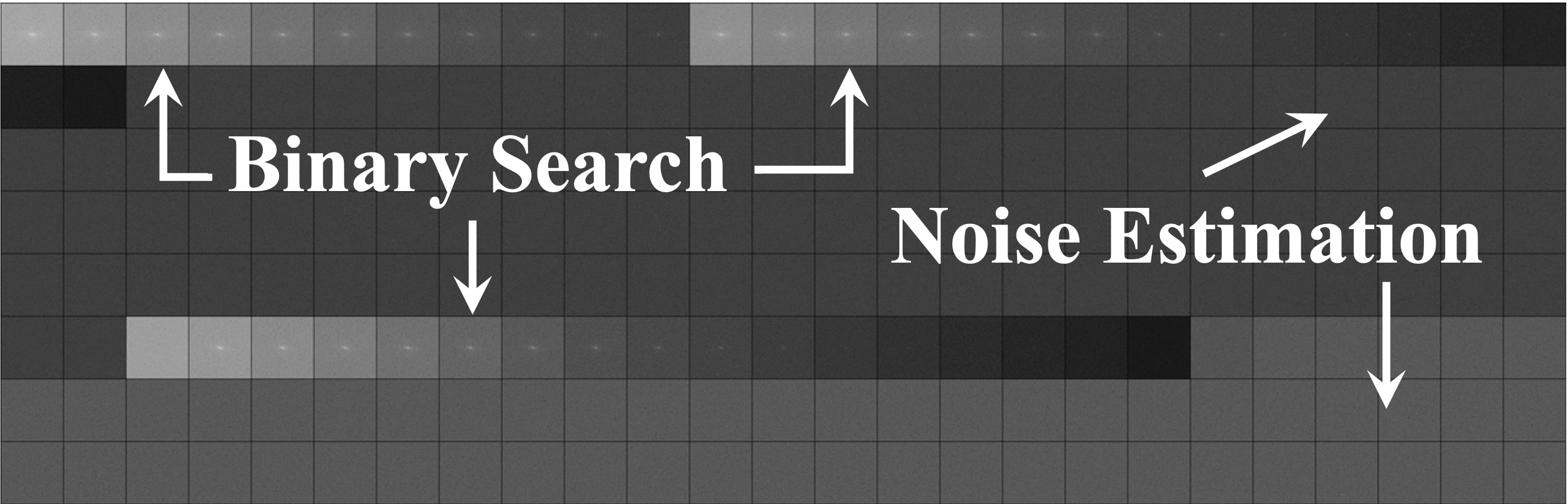}\label{fig:demo:hsj-spec}}
    
    \subfloat[\texttt{GeoDA-2} Queries ($\bx_t$)]{\includegraphics[width=0.32\textwidth]{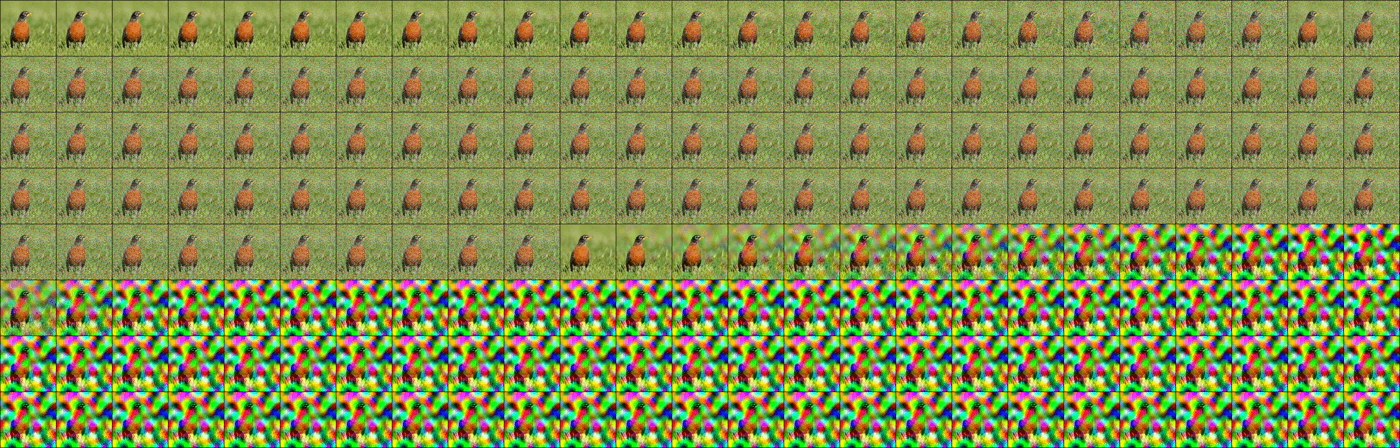}\label{fig:demo:geoda-raw}}
    \hfill
    \subfloat[\texttt{GeoDA-2} Steps ($\bx_t-\bx_{t-1}$)]{\includegraphics[width=0.32\textwidth]{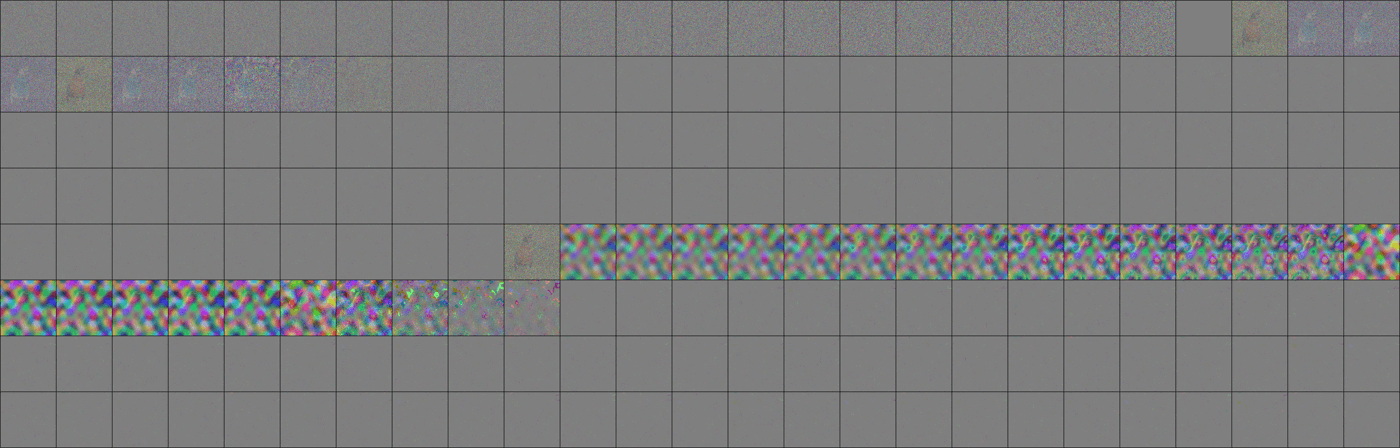}\label{fig:demo:geoda-diff}}
    \hfill
    \subfloat[\texttt{GeoDA-2} Steps in Spectrum ($\bx_t-\bx_{t-1}$)]{\includegraphics[width=0.32\textwidth]{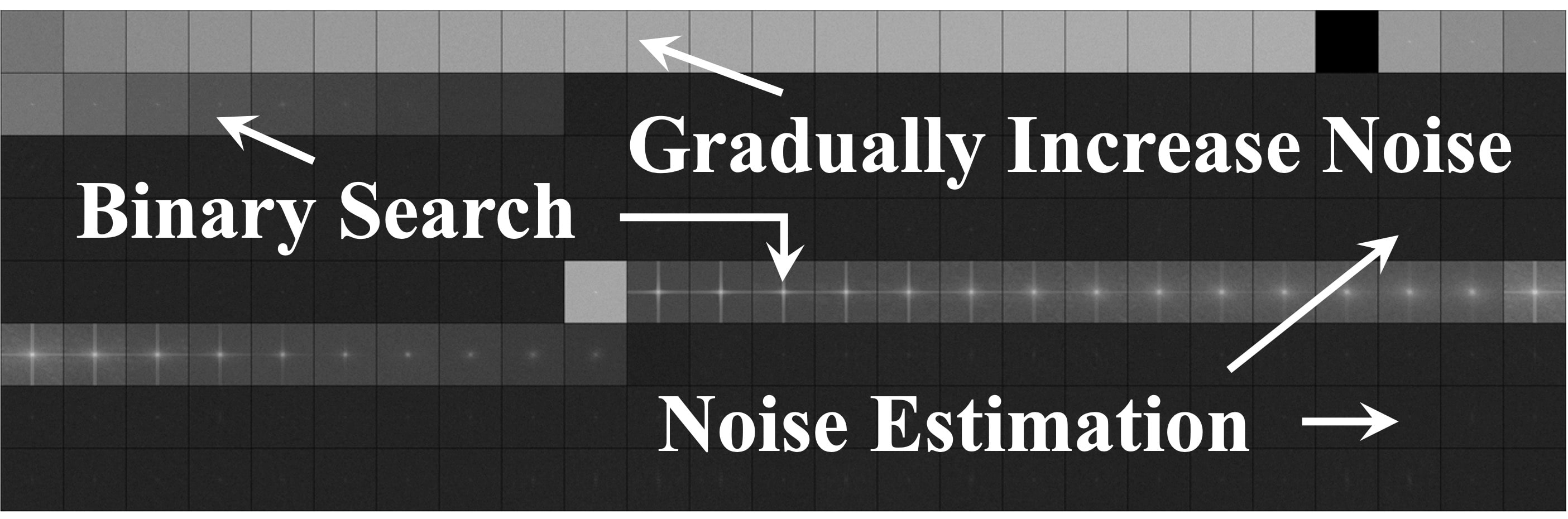}\label{fig:demo:geoda-spec}}
    \caption{The first 200 queries of \texttt{HSJ-2} and \texttt{GeoDA-2} attacks against the same clean image. \textbf{Left:} It is hard to precisely identify each attack's actions in each query. \textbf{Medium and Right:} The per-query changes between successive queries reveal detailed, human-identifiable behaviors of the two attacks, especially in the spectrum space.}
    \label{fig:demo:traces}
\end{figure*}

In this section, we introduce the pipeline and settings of \SEA, which is illustrated in \Cref{fig:demo:pipeline}.

\subsection{Threat Model}
\label{sec:overview:threat}

\paragraph{Attacker}
We focus on the standard threat model of query-based black-box attacks, where the attacker issues a sequence of $n$ queries $X=\s{\bx_t}_{t=1}^n$ to the model to generate an adversarial example, $\bx^\prime = \bx_n$. We refer to this query sequence as the attack's trace, which constitutes an attack incident. The model returns the final predicted label, potentially with a confidence score. The attacker does not have any knowledge of the classifier's architecture or gradients. We also assume that the final adversarial example is bounded by a perturbation budget $\epsilon$ under some distance metric, such as the \LL{p} norm distance.

In this work, we focus on attacks for image classification, but it is straightforward to generalize \SEA{} to other domains like text classification, as we will demonstrate in \Cref{sec:exp:text}.

\paragraph{Forensic System}
\SEA{} aims to identify the most probable attack that generated a given trace $X$ and characterize its behavior. \SEA{} has access to the adversarial example $\bx^\prime$ as well as the history queries $Q$, which contain the attack's trace $X \subseteq Q$, but does not know the account issuing the queries. \SEA{} assumes access to history queries because forensic analysis is only possible with access to the attack's trace, implying an inevitable storage and privacy concern. We thus consider scenarios where \textit{security outweighs storage and privacy}. We study the storage trade-off in \Cref{sec:exp:ablation}, and its real-world examples and how to obtain the adversarial examples in \Cref{app:discuss}.

\paragraph{Non-Goals}
\SEA{} does not aim to identify the precise attacker or determine the exact attack algorithm used, and its primary function is not to serve as a classifier of attacks or a standalone tool for forensic analysis. Instead, \SEA{} continually updates its database with fingerprints of newly observed attack behaviors. This ongoing effort allows \SEA{} to associate new incidents with existing fingerprints, thereby facilitating a more comprehensive forensic analysis when combined with other auxiliary information. This approach is akin to the methods employed in analyzing and tracking new malware, where the focus is on identifying behavior patterns rather than the direct identification of attackers or specific malware signatures.

\subsection{Problem Definition}
\label{sec:overview:definition}

Given an adversarial example, \SEA{} aims to attribute its attack trace to a known attack (if possible), provide the rationale behind its attribution, and share attack knowledge with other parties. Specifically, \SEA{} has the following three objectives: 

\begin{enumerate}[leftmargin=*]
    \item \textbf{Sharing.} The system should produce a shareable fingerprint that concisely represents the underlying attack. This process would help link potentially related incidents to the observed attack. It is preferable to achieve high data efficiency: the system should ideally learn the fingerprint from the attack's first incident (i.e., a single trace).
    \item \textbf{Explanation.} The system should be able to explain the rationale behind its attribution results. A good explanation should generally describe the underlying attack's behavior, or at least the factors that result in the attribution.
    \item \textbf{Attribution.} The system should attribute a given trace to a known attack with high accuracy. The attribution should adapt to the attack's change of behavior, yet it should generalize to different incidents of the same attack. 
\end{enumerate}

\paragraph{Theoretical Limitation}
We acknowledge a theoretical limitation in the task we aim to solve, where pursuing a perfect solution could be intractable. \SEA{} attempts to solve an inverse problem: observing a sequence of queries, it tries to find the attack that generated them. Such a problem is hard to solve as infinitely many programs can produce a given trace. Hence, it is difficult to provide a guarantee for generalization across datasets, models, and attack incidents. Attributing the observed trace to some attack only allows us to argue that it is the most likely source among a set of known attacks. Note that the same limitations apply to classic software forensics.

\subsection{Attack Identity}
\label{sec:overview:identity}

Attack \textit{identity} is a key concept in the attribution task. Given the above limitations, \textbf{we define attack identity based on the attack's observed behavior, instead of its exact algorithm or specifications}. That means, two attacks will have the same identity as long as they have the same behavior, even if their actual algorithms or hyper-parameters are different. We model the attribution as an $M$-ary hypothesis testing problem, where each attack identity is a hypothesis and $M$ is the number of attacks. Each attack $\Attack_i$ is associated with a probability distribution $\mathcal{P}_{\Attack_i}$, with density $P_{\Attack_i}=\Prob{X\given\Attack_i}$, over all possible traces. We assume that all attacks happen with the same probability so that, given an attack trace $X$, the optimal decision rule is choosing the attack with the maximum likelihood $P_{\Attack_i}(X)$; i.e., $\mathcal{A}^* = \argmax_{i\in[M]} P_{\Attack_i}(X)$.

With this definition, \SEA{} chooses the optimal hypothesis for a given trace $X$ in the Bayesian setting. As the space of traces is intractable, it is not feasible to determine the decision rule ahead of time. Instead, \SEA{} computes the probabilities of each attack generating the observed trace, and then determines the most probable attack. We show in the next section how we model attacks as a Hidden Markov Model to obtain $P_{\Attack_i}(X)$. The advantage of this approach, compared to neural networks, is that it provides an explainable final decision. An analyst will have access to how \SEA{} attributes the trace to an attack.

\section{Characterizing Black-box Attacks}
\label{sec:hmm}

In this section, we introduce a novel modeling of black-box attacks using Hidden Markov Models. This modeling allows us to characterize the \textit{behavior} of black-box attacks in a unified framework, encompassing the many attacks, approaches, and strategies proposed in the black-box attack literature.

\subsection{Visualizing Black-box Attack Behaviors}
\label{sec:hmm:pattern}

We begin with a preliminary study to illustrate how attacks can exhibit distinct patterns, even when a human analyst is \textit{manually inspecting} their traces. Building on this insight, we then model black-box attacks using Hidden Markov Models (HMMs) in \Cref{sec:hmm:attack}. This approach is comprehensible to humans, enabling them to understand the progression of an attack by examining its trace on a query-by-query basis.

We first visualize the traces of HSJ~\cite{hsj} and GeoDA~\cite{geoda} attacks to determine if they exhibit unique behaviors. In \Cref{fig:demo:hsj-raw,fig:demo:geoda-raw}, we observe some differences between the two traces. However, it is hard to precisely identify each attack's actions in each query. One may also notice the similarity among queries, which is important for detecting black-box attacks~\cite{blacklight,stateful-detection}, but this feature cannot differentiate between attack types. In particular, it is not immediately clear to a human observer how to break down the trace into fine-grained attack phases.

To gain deeper insights, we then focus on the \emph{per-query changes} between successive queries $\bdelta_t=\bx_{t} - \bx_{t-1}$ and visualize them in \Cref{fig:demo:hsj-diff,fig:demo:geoda-diff}. This analysis reveals more detailed behaviors of the two attacks, including identifiable phases (e.g., line search and gradient estimation) and their distinct patterns (e.g., the various spectrum peaks).

The above visualizations demonstrate the potential of applying forensic analysis to black-box attack traces. While this visual interpretation provides forensic analysts with intuitive insights into different attack behaviors, it lacks a formalized structure for broader application and sharing among different parties. Thus, we still need a systematic and quantitative framework to characterize these attack behaviors.

\subsection{Modeling Black-box Attacks as HMMs}
\label{sec:hmm:attack}

Black-box attacks are commonly described as programs that repetitively query a model~\cite{breaking-eggs}. Prior research has identified that black-box attacks typically follow a fixed set of procedures, such as gradient estimation and line search~\cite{scaling}. These procedures can be further broken down into scalar and vector operations~\cite{autoda}. However, an external observer, like the victim model, cannot directly discern the internal state or the specific procedures being executed by the attack. The best approach for the victim is to infer the internal procedures over time based on the observed behavior of the queries. \emph{This situation aligns perfectly with the purpose of Hidden Markov Models (HMMs).}

An HMM is characterized by a sequence of \emph{hidden states} and \emph{observations}. The hidden states represent the underlying, unobservable processes of a system, while observations are the visible outcomes of such processes. The primary goal of HMMs is to infer the hidden states based on the observations and their probabilistic connections.

\begin{figure}
    \centering
    \includegraphics[width=\linewidth]{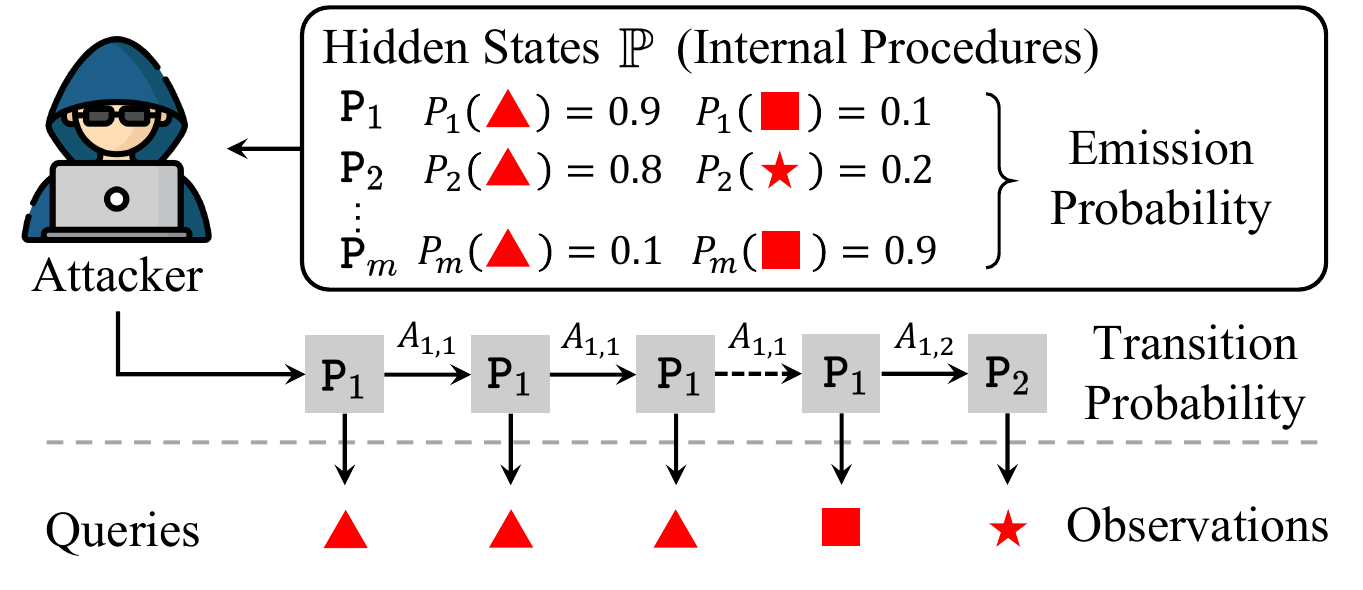}
    \caption{The analogy between black-box attacks and HMMs.}
    \label{fig:hmm}
\end{figure}

This framework draws a compelling analogy to the black-box attack scenario. From an external viewpoint, a black-box attack can be modeled as a stochastic process --- at each step, the attack selects an internal \emph{procedure} which then produces a \emph{query}. \Cref{fig:hmm} provides a visual representation of how each attack step corresponds to the components of an HMM. While one may argue that black-box attacks are deterministic (given specific inputs, target model, and seed), this determinism only holds when the underlying attack is already known. When analyzing an unknown attack, its behavior is more accurately represented as a stochastic process.

\paragraph{Attack Model}
Building on the HMM analogy, we propose modeling black-box attacks using an HMM framework that includes \emph{procedures} and \emph{queries}. This allows us to apply the HMM's extensive theoretical tools to unify various black-box attacks. Our formalization in HMM notation is as follows:

\begin{itemize}[leftmargin=*]
    \item \textbf{Observations:} Each observation corresponds to a query from the attack. The space of observations is $\DD\coloneqq[0, 1]^d$ as we defined earlier in \Cref{sec:background:adv}, which is discrete with size $|\DD|=256^d$ due to the 8-bit representation of images.
    \item \textbf{Hidden States:} Each hidden state $\PROC{P}_j$ represents an internal procedure of the attack, where $\mathbb{P}\coloneqq\s{\PROC{P}_j}_{j=1}^m$ is the space of all hidden states and $m$ is the total number of states.
    \item \textbf{Emission Probability:} The probability of a state $\PROC{P}_j$ producing a query $\bx_t$ given the previous query $\bx_{t-1}$. It is characterized by the distribution $\mathcal{P}_j$ with PMF $P_j(\bx_t|\bx_{t-1})$, which is conditional as we analyze the changes between consecutive queries.
    \item \textbf{Transition Probability:} The probability of an attack transitioning from one state to another over time. It is a matrix $\TransMat\in[0, 1]^{m\times m}$, where $A_{i,j}\coloneqq\Prob{\PROC{P}^{(t)}=\PROC{P}_j \given \PROC{P}^{(t-1)}=\PROC{P}_i}$ is the probability of transitioning from state $\PROC{P}_i$ to state $\PROC{P}_j$.
\end{itemize}

\paragraph{Attack Representation}
With the above formalization, we represent each attack as an HMM parameterized by $\Attack_i=(\TransMat, \EmissionProb)$, where $\EmissionProb\coloneqq\s{P_j}_{j=1}^m$ is the global emission probability shared by all attacks, and $\TransMat$ is the transition matrix specific to each attack. This representation also instantiates the attack's identity in \Cref{sec:overview:identity}; given the attack trace, $X=\s{\bx_t}_{t=1}^n$, the probability $P_{\Attack_i}(X)$ of attack $\Attack_i$ producing trace $X$ is
\begin{equation}
\small
\label{eq:hmm_inference}
\begin{aligned}
\small
&P_{\Attack_i}(X) = \Prob{X\given\Attack_i} = \sum_{j=1}^{m}{\alpha_n(j)} , \\
&\text{where } \alpha_{t+1}(j) = \left[\sum_{k=1}^{m}{\alpha_t(k)\times A_{k,j}}\right]\times P_j(\bx_{t+1}) \\
&\text{and } \alpha_{1}(j) = \frac{1}{m} \times P_j(\bx_{1}),
\end{aligned}
\end{equation}
where $\alpha_t(k)$ is the forward probability of procedure $\PROC{P}_k$ at step $t$. This is known as the Forward algorithm~\cite{hmmbook}.

\begin{figure*}[t]
  \centering
  \stackunder[5pt]{\includegraphics[width=0.082\linewidth]{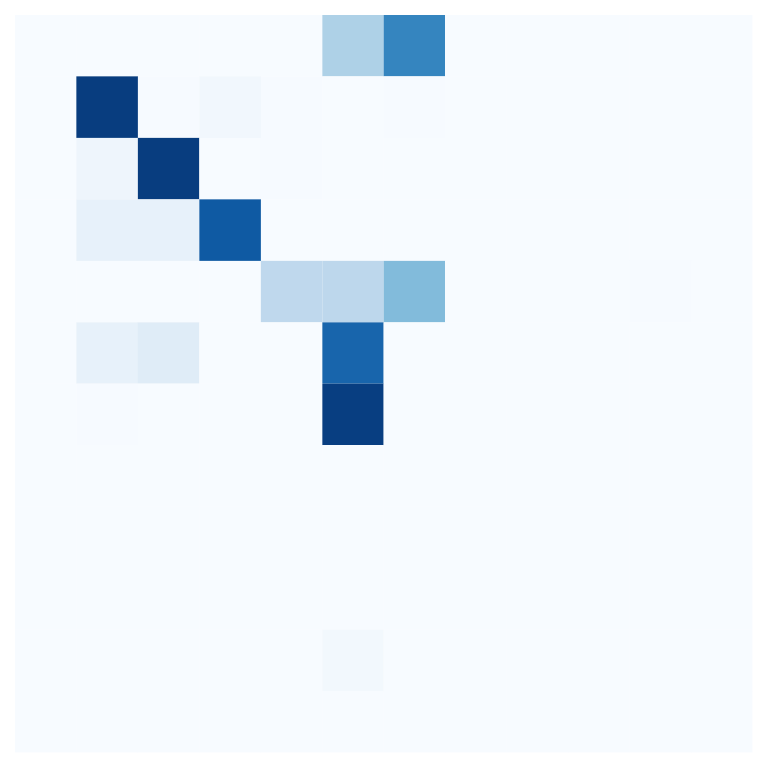}}{\footnotesize\texttt{HSJ-2}}
  \stackunder[5pt]{\includegraphics[width=0.082\linewidth]{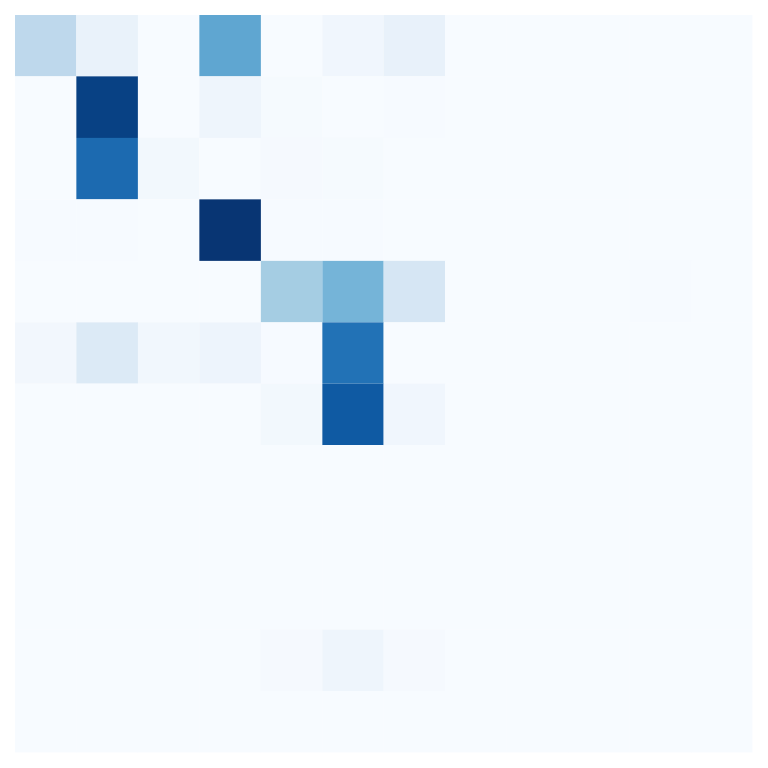}}{\footnotesize\texttt{HSJ-Inf}}
  \stackunder[5pt]{\includegraphics[width=0.082\linewidth]{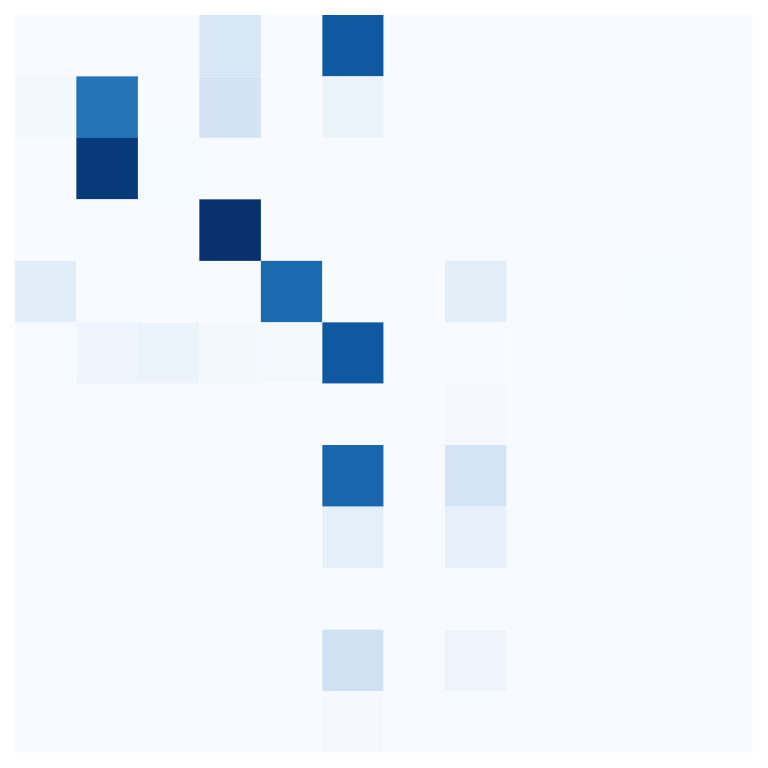}}{\footnotesize\texttt{GeoDA-2}}
  \stackunder[5pt]{\includegraphics[width=0.082\linewidth]{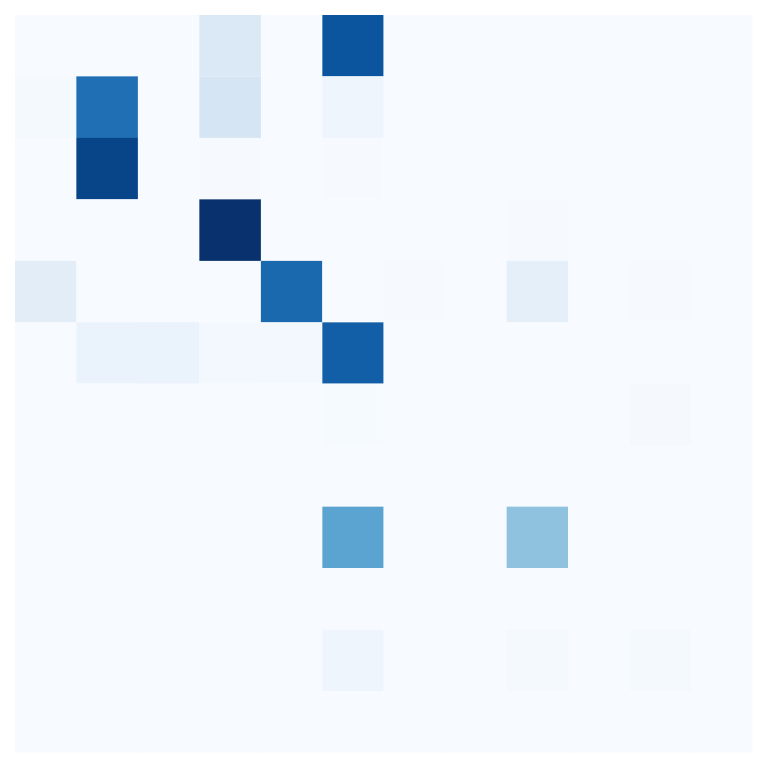}}{\footnotesize\texttt{GeoDA-Inf}}
  \stackunder[5pt]{\includegraphics[width=0.082\linewidth]{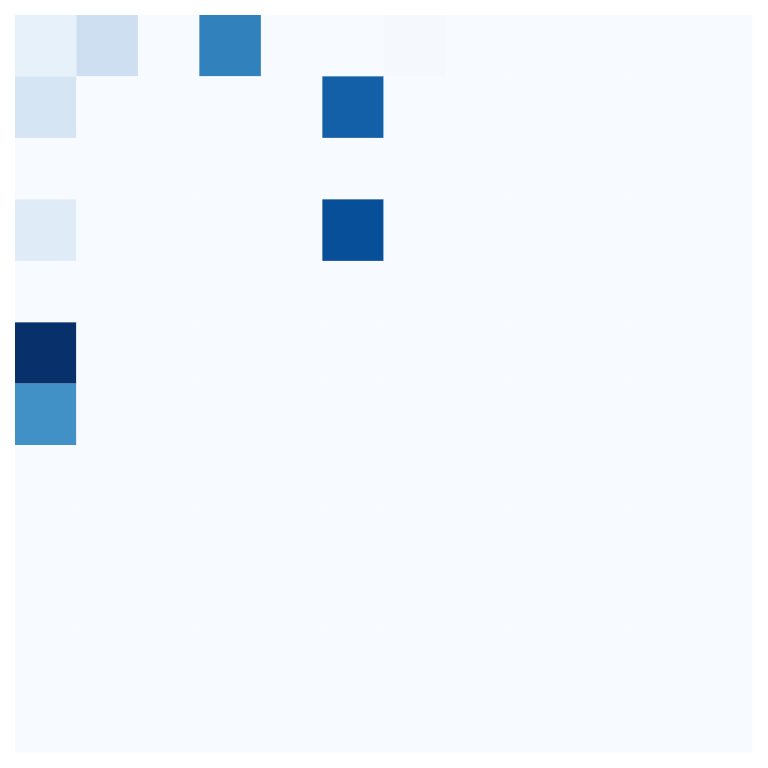}}{\footnotesize\texttt{Square-2}}
  \stackunder[5pt]{\includegraphics[width=0.082\linewidth]{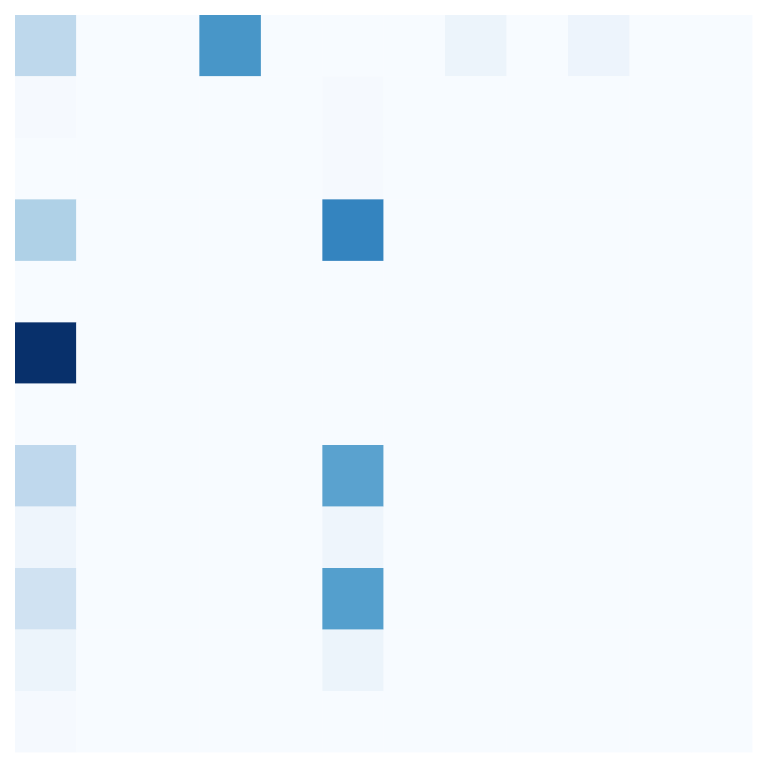}}{\footnotesize\texttt{Square-I}}
  \stackunder[5pt]{\includegraphics[width=0.082\linewidth]{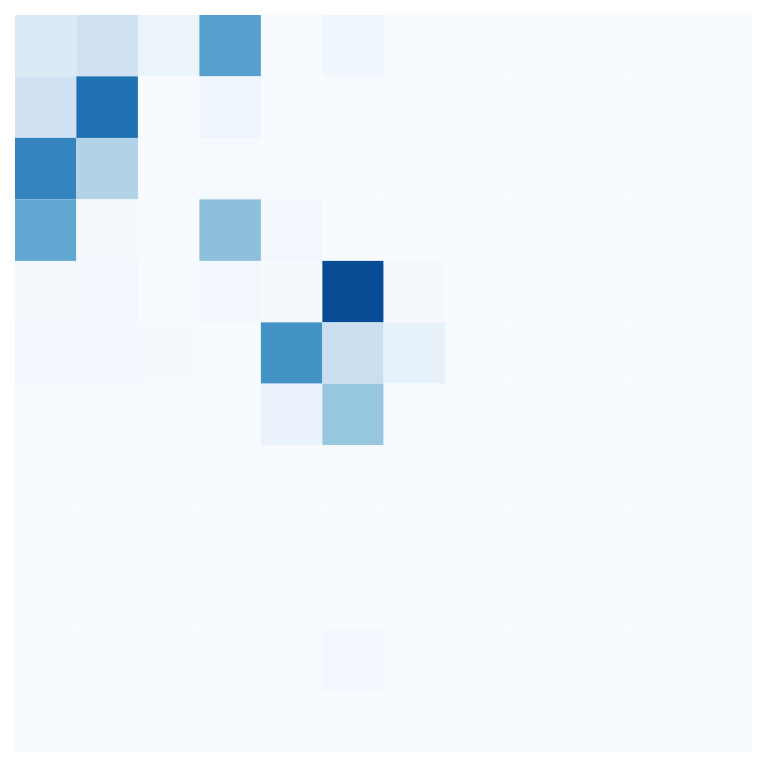}}{\footnotesize\texttt{SignOPT-2}}
  \stackunder[5pt]{\includegraphics[width=0.082\linewidth]{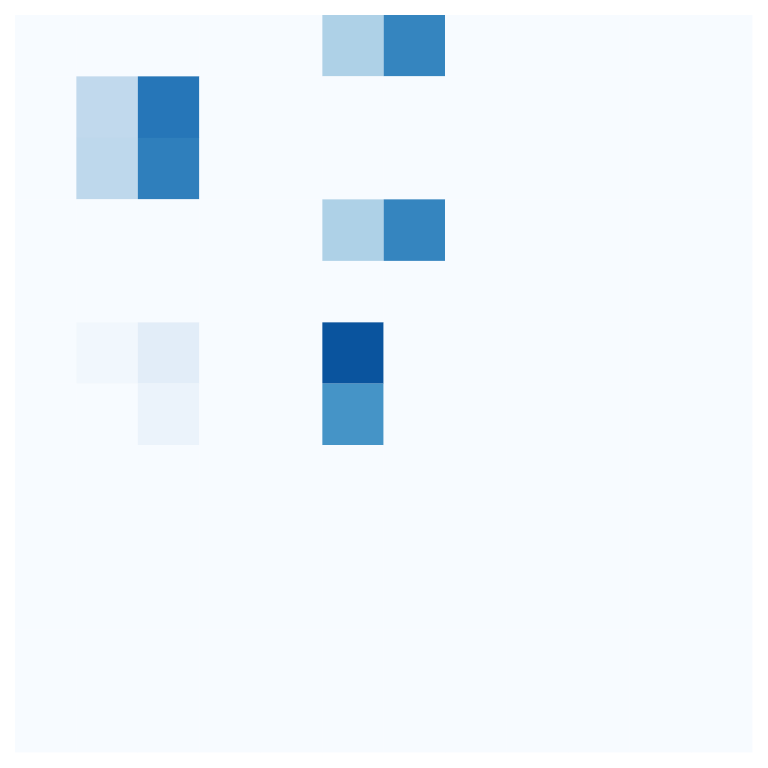}}{\footnotesize\texttt{Bound-2}}
  \stackunder[5pt]{\includegraphics[width=0.082\linewidth]{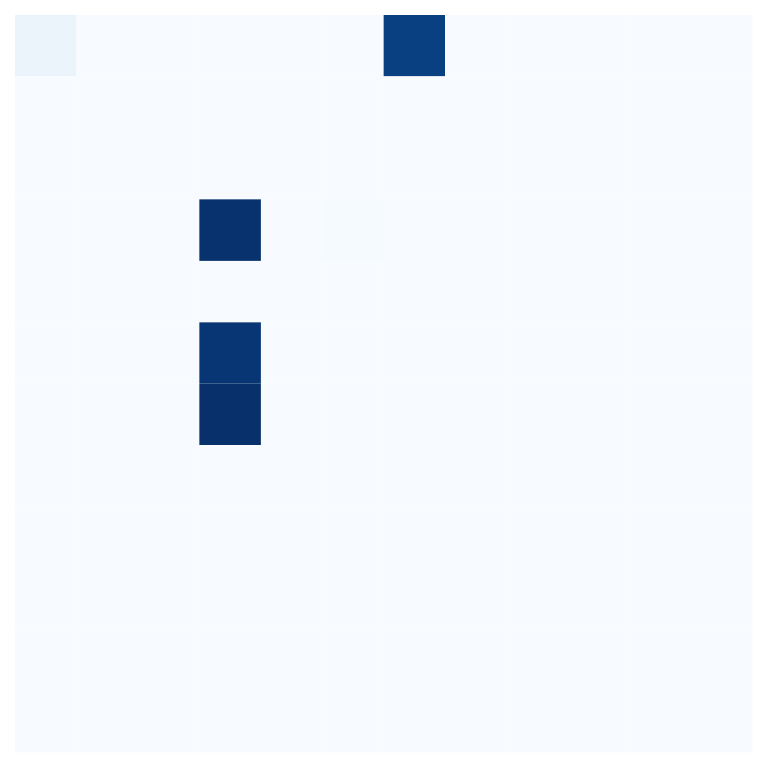}}{\footnotesize\texttt{ECO-2}}
  \stackunder[5pt]{\includegraphics[width=0.082\linewidth]{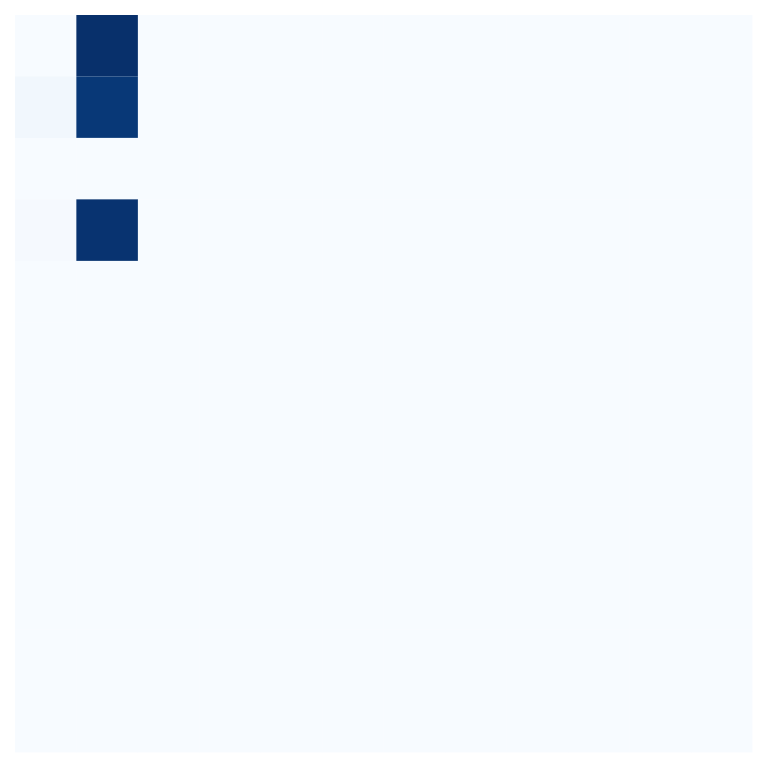}}{\footnotesize\texttt{NES-2}}
  \stackunder[5pt]{\includegraphics[width=0.082\linewidth]{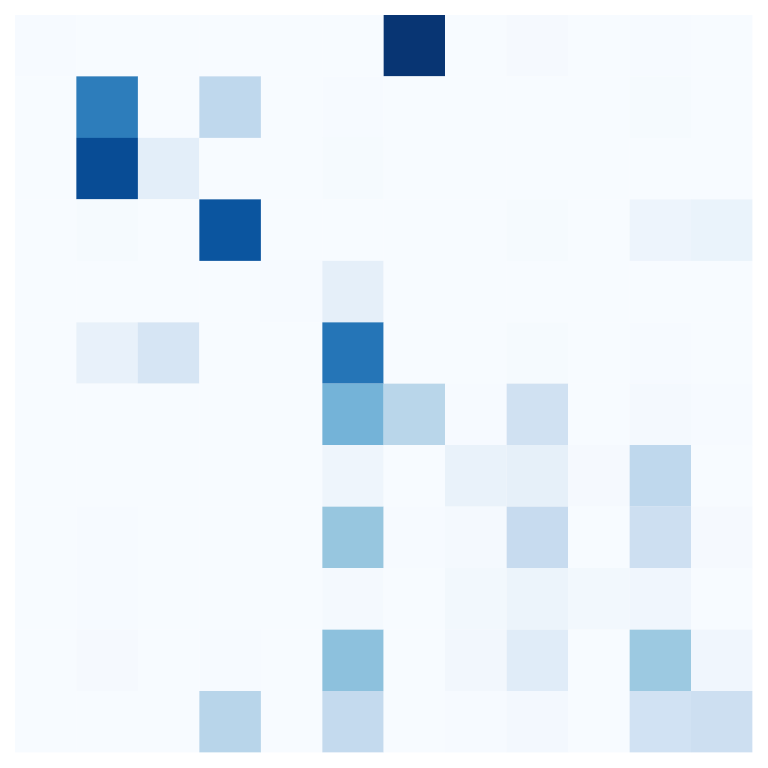}}{\footnotesize\texttt{RayS-Inf}}
  \caption{Fingerprints of the 11 attack variants we studied. Detailed numbers and variances can be found in \Cref{app:visualize:fingerprint}.}
  \label{fig:fp-dense}
\end{figure*}

\subsection{Modeling Black-box Attack Procedures}
\label{sec:hmm:procedure}

As the last step of our effort to model black-box attacks using HMMs, we now define the probability distribution $\ProcedureDist$ of each procedure \PROC{P}. To reduce the modeling complexity, we abstract attack procedures into two types: \emph{noise procedures} and \emph{pattern procedures}. Each type has a distinct definition of PMFs that characterizes how they generate queries. This abstraction lays the foundation for our later, detailed instantiation of specific attack instances in \Cref{sec:method:enroll:discover}.

\paragraph{Noise Procedures}
This type of procedure involves adding i.i.d.\ noise to the pixels of its input. That is, for a given noise procedure and $\bx_t\sim\ProcedureDist$, the pixels of its change $\bdelta_t\coloneqq\bx_t-\bx_{t-1}$ are independently drawn from an identical distribution $\ProcedureDist^\prime$. Thus, the PMF of such a procedure, representing its emission probability, can be defined as
\begin{equation}
\label{eq:noise}
	P(\bx_t | \bx_{t-1})\coloneqq
	  \Pr_{\delta_i\iid\ProcedureDist^\prime}\ProbBr{\delta_1, \delta_2, ..., \delta_d}
	= \prod_{i=1}^d\Pr_{\delta_i\iid\ProcedureDist^\prime}\br{\delta_i},
\end{equation}
where $\s{\delta_i}_{i=1}^d$ are the pixel values of $\bdelta_t$.

The underlying pixel-level distribution $\ProcedureDist^\prime$ varies for different instances of noise procedures. We will explain how to discover and instantiate such procedures in \Cref{sec:method:enroll:discover}.

\paragraph{Pattern Procedures}
This type of procedure induces changes correlating with specific patterns. To model this, we first use a pattern matching metric $\Metric:\DD\to[0,1]$ to measure the similarity between the change $\bdelta_t$ and a pattern template $\btau$. This metric is then transformed into a probability measure, similar to noise procedures in \Cref{eq:noise}.

For calculating the similarity, we employ the Pearson correlation coefficient, a standard template matching metric:
\begin{equation}
\label{eq:pattern}
	\Metric(\bdelta_t, \btau) \coloneqq \abs{R(\bdelta_t, \btau)},
\end{equation}
where the sign is ignored because we are not interested in the correlation's direction, and $R$ is defined as:
\begin{equation}
\label{eq:corr}
	R\p{\bz_1, \bz_2} = \frac{(\bz_1-\bar{\bz}_1)(\bz_2-\bar{\bz}_2)}{\norm{\bz_1-\bar{\bz}_1}\norm{\bz_2-\bar{\bz}_2}}\in[-1, 1].
\end{equation}

To transform this metric into a probability distribution, we enforce a truncated exponential distribution on $\Metric$:
\begin{equation}
\label{eq:truncated}
	P(\bx_t | \bx_{t-1})\coloneqq
	c \cdot \frac{1}{{|\mathbb{D}|^{1-\Metric(\bdelta_t, \btau)}}},
\end{equation}
where $c$ is the normalization constant. As the metric $\Metric(\bdelta_t, \btau)$ approaches one, indicating a matched pattern, the probability converges to one, and conversely, approaches zero for lower scores. The intuition of this transformation and other definitions of the metric are discussed in \Cref{app:formal}.

The underlying pattern $\btau$ varies for different instances of pattern procedures. We will explain how to discover and instantiate such patterns in \Cref{sec:method:enroll:discover}.

\paragraph[?]{Why not modeling the actual procedures}
We focus on modeling the \textit{observable} attack procedures instead of the \textit{actual} procedures defined by black-box attacks. This choice stems from the
impracticability of modeling the actual procedures.
\emph{First}, modeling the actual procedures typically requires access to the attack's source code, which is usually unavailable, especially for novel attacks.
\emph{Second}, the behavior of an internal procedure can be inherently inconsistent, influenced by its progression and random internal states. For instance, a line search procedure may exhibit varying behaviors throughout its execution, hindering consistent characterization.
\emph{Lastly}, the sheer output space $|\DD|=256^d$ renders techniques like Monte Carlo simulations ineffective for estimating the distribution.

\section{System Design}
\label{sec:method}

In this section, we introduce the detailed design of \SEA{} and how it incorporates the HMM-based modeling of black-box attacks. Specifically, \SEA{} takes as input an adversarial example $\bx^\prime$ and history queries $Q$ to execute the following workflow:
\begin{enumerate}[leftmargin=*]
	\item \textbf{Extract} attack trace from history queries (\Cref{sec:method:extract}).
	\item \textbf{Attribute and Explain} the attack behavior (\Cref{sec:method:match}).
	\item \textbf{Fingerprint and Share} the attack behavior (\Cref{sec:method:fingerprint}).
\end{enumerate}

Eventually, \SEA{} outputs a fingerprint for each attack incident, which is then shared for wider attribution and analysis. A preview of these fingerprints is shown in \Cref{fig:fp-dense}. To facilitate this workflow, \SEA{} maintains several key resources:
\begin{itemize}[leftmargin=*]
	\item A \textbf{procedure database} $\mathbb{P}=\s{\PROC{P}_i}_{i=1}^m$, where each procedure $\PROC{P}_i$ corresponds to a distribution $\mathcal{P}_i$ with PMF $P_i(\bx)$, aligning with the emission probability defined in \Cref{sec:hmm:procedure}.
	\item An \textbf{attack database} $\AttackDB=\s{\Attack_i}_{i=1}^M$, where each attack $\Attack_i$ is represented by its transition matrix $\TransMat$ in \Cref{sec:hmm:attack}.
	\item A \textbf{fingerprint database} $\FF=\s{\FP_1, \FP_2, ...}$ that enables intelligence sharing, where each fingerprint $\FP_i$ is associated with an attack incident. We explain its format in \Cref{sec:method:fingerprint}.
\end{itemize}

\subsection{Extracting Attack Trace}
\label{sec:method:extract}

\SEA{} needs to extract the attack trace $X$ from history queries for subsequent analysis, where the last query $\bx_n$ is the adversarial example $\bx^\prime$ that triggers the security alarm. One may obtain the trace by filtering the source account, but we focus on the more general threat model without account information.

\paragraph{Theoretical Limitations}
We first note that a perfect extraction of the attack trace is impossible. Although prior research has identified the similarity among queries from black-box attack~\cite{blacklight,stateful-detection}, the attack may still produce heavily distorted queries that are uncorrelated with the final adversarial example. For example, decision-based attacks usually start with querying the model using pure random noise~\cite{hsj}. It is impossible to associate such queries with the underlying attack.

\paragraph{Identifying Similar Queries}
Given these limitations, we aim to identify queries similar to the adversarial example $\bx^\prime$, thus extracting an \textit{imperfect yet informative} attack trace $X$. While identifying similar queries is challenging in real-time detection defenses due to false positives, forensic analysis has the advantage of hindsight --- knowing the final adversarial example. This allows us to consider any query with significant correlation to $\bx^\prime$ as potentially part of the attack.

To implement this, \SEA{} employs the DBSCAN clustering algorithm, setting the minimum cluster size to one and using correlation as the distance metric. The process begins with the adversarial example $\bx^\prime$ and incrementally includes historical queries that demonstrate a high correlation with any query already identified as part of the trace, formalized as
\begin{align*}
    &\texttt{Initialize: }X \coloneqq \Set{ \bx^\prime}\\
    &\texttt{Repeat: }
    X \gets X \cup \Set{ \bx \in Q \given \exists\ \bz \in X,\  R(\bz, \bx) \geq r },
\end{align*}
where the correlation coefficient $R$ and a system threshold $r$ manage the balance between trace completeness and the risk of including benign queries. Our empirical analysis in \Cref{sec:exp:ablation} shows that the standard threshold $r=0.5$ in template matching yields the best performance.

\subsection{Attributing and Explaining Attacks}
\label{sec:method:match}

With the extracted attack trace $X$, \SEA{} now progresses to attributing this trace to a known attack and explaining its behavior, utilizing the HMM framework's capabilities.

\paragraph{Attributing Attacks}
Attributing the trace to a known attack is analogous to model selection in HMMs. We compare the trace against a set of predefined HMMs, each representing a different known attack, to identify the most likely source:
\begin{equation}\label{eq:match-pattern}
	\Attack^* = \argmax{}_{\Attack\in\AttackDB}\ \Prob{X \given \Attack},
\end{equation}
where $\Prob{X \given \Attack}$ is the likelihood of the sequence $X$ under the HMM parameterized by $\Attack$. This model selection is efficiently computed using the Forward algorithm~\cite{hmmbook}.

\paragraph{Explaining Attacks}
Upon identifying the most likely attack $\Attack^*$, our next goal is to explain its behavior by deducing the sequence of procedures that likely led to the observed trace. This task aligns with the decoding problem in HMMs, which seeks to uncover the most probable sequence of hidden states (i.e., procedures) that resulted in the observed sequence:
\begin{equation}\label{eq:explain-pattern}
	Y^* = \argmax{}_{Y\in\PP^n}\ \Prob{Y \given X, \Attack^*},
\end{equation}
where $\Prob{Y \given X, \Attack^*}$ is the likelihood of procedure sequence $Y=\s{\PROC{P}^{(t)}}_{t=1}^n$ given the trace $X$ and the attack $\Attack^*$. This problem is efficiently solved by the Viterbi algorithm~\cite{viterbi}.

\subsection{Fingerprinting and Sharing Attacks}
\label{sec:method:fingerprint}

While matching and explaining attack traces enable in-depth local analysis, sharing this knowledge with other entities is crucial for broader cybersecurity efforts. \SEA{} addresses this by generating a fingerprint for each observed attack incident, facilitating the dissemination of attack intelligence.

\paragraph{Fingerprinting Attacks}
The key to \SEA{}'s fingerprinting process is the use of the transition matrix $\TransMat$ from the HMM framework. Each attack trace's pattern is encapsulated in its unique transition matrix, which serves as the fingerprint $\FP$:
\begin{equation}
\FP \coloneqq \hat{A} = \argmax{}_{A\in\RR^{m\times m}}\ \Prob{X \given A}.
\end{equation}

The transition matrix $\hat{A}$, representing the fingerprint, is derived using the EM algorithm~\cite{hmmbook}, which maximizes the likelihood of $\hat{A}$ given the observed trace $X$. Over time, as \SEA{} gathers multiple fingerprints for each attack, a comprehensive database of these fingerprints is established, enhancing the accuracy and scalability of attack attribution.

\paragraph{Sharing and Comparing Fingerprints}
The ability to share and compare fingerprints is a crucial aspect of \SEA{}'s design. Consider two fingerprints: $\FP_1$ from a known attack instance and $\FP_2$ from a current trace. Their similarity can be computed using cosine similarity:
\begin{equation}
\mathtt{cos\text{-}sim}(\FP_1, \FP_2) = \frac{\FP_1\cdot\FP_2}{\norm{\FP_1}\norm{\FP_2}},
\end{equation}
where fingerprints are first flattened into vectors. Additionally, an observed trace can be attributed to an attack with the most similar average fingerprint:
\begin{equation}
\label{eq:fp-cos}
\Attack^* = \argmax{}_{\Attack\in\AttackDB}\ \mathtt{cos\text{-}sim}\p*{\FP, \sum_{\FP^\prime\in\FF_\Attack}\frac{\FP^\prime}{|\FF_\Attack|}},
\end{equation}
where $\FP$ is the fingerprint being analyzed, and $\FF_\Attack\subseteq\FF$ is the collection of known fingerprints for a specific attack $\Attack$.

\paragraph[?]{Why not neural networks}
While neural networks might excel in performance, they often lack explainability and shareability. HMM-based fingerprints not only retain high explainability but also facilitate straightforward sharing and comparison across different cybersecurity entities. We also discuss why the model outputs are not trusted or utilized in \Cref{app:discuss}.

\section{System Implementation}
\label{sec:method:enroll}

In this section, we describe the last step for \SEA{}'s functionality: the enrollment of known black-box attacks into procedure and attack databases. This enrollment process is also applicable for analysts to add new attacks to \SEA{}.

\begin{figure}[t]
    \centering
    \subfloat[Attack Steps (Spectrum)]{\includegraphics[width=0.48\linewidth]{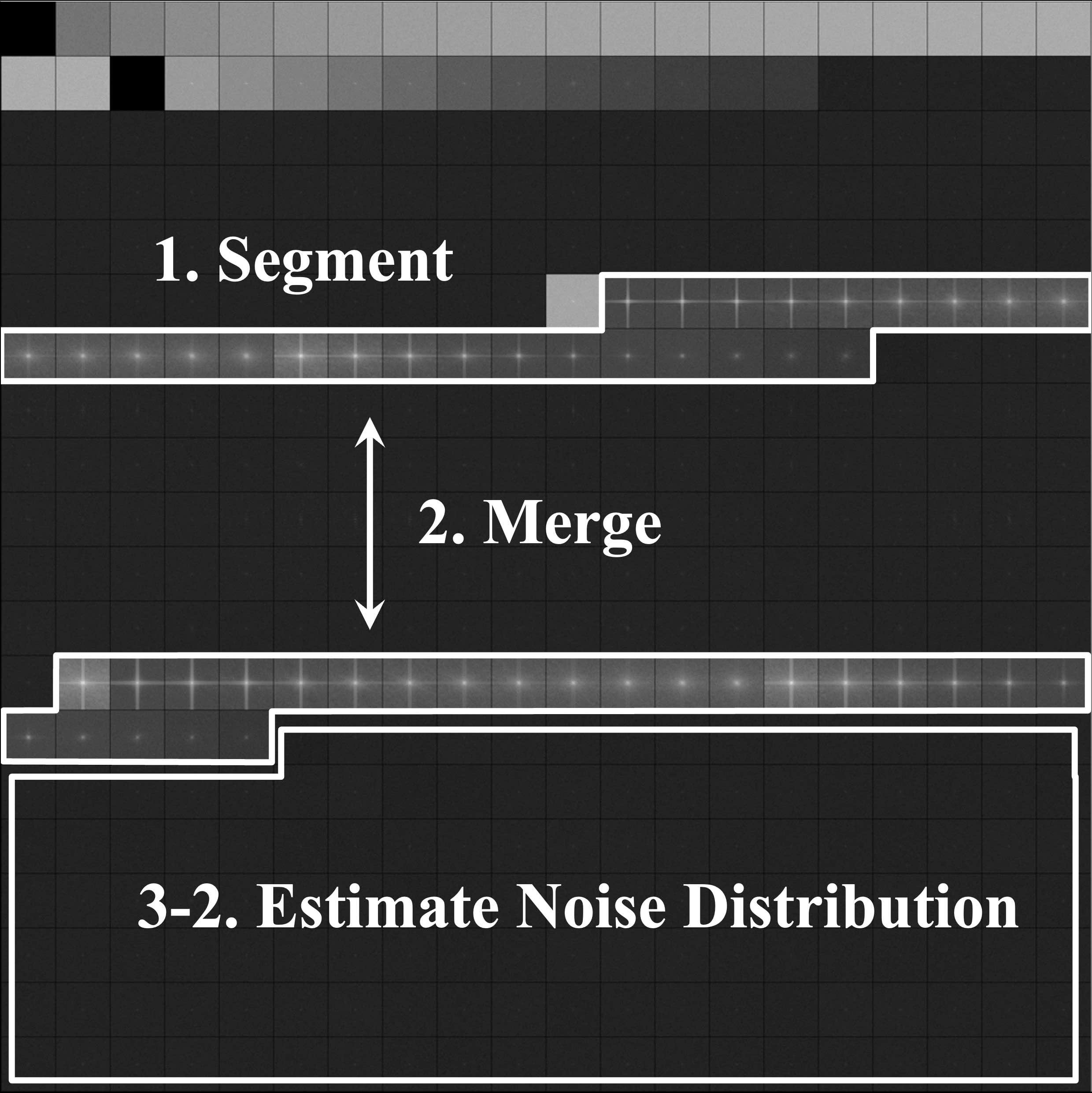}\label{fig:demo:cluster-1}}
    \hfill
    \subfloat[Attack Steps (Binarization)]{\includegraphics[width=0.48\linewidth]{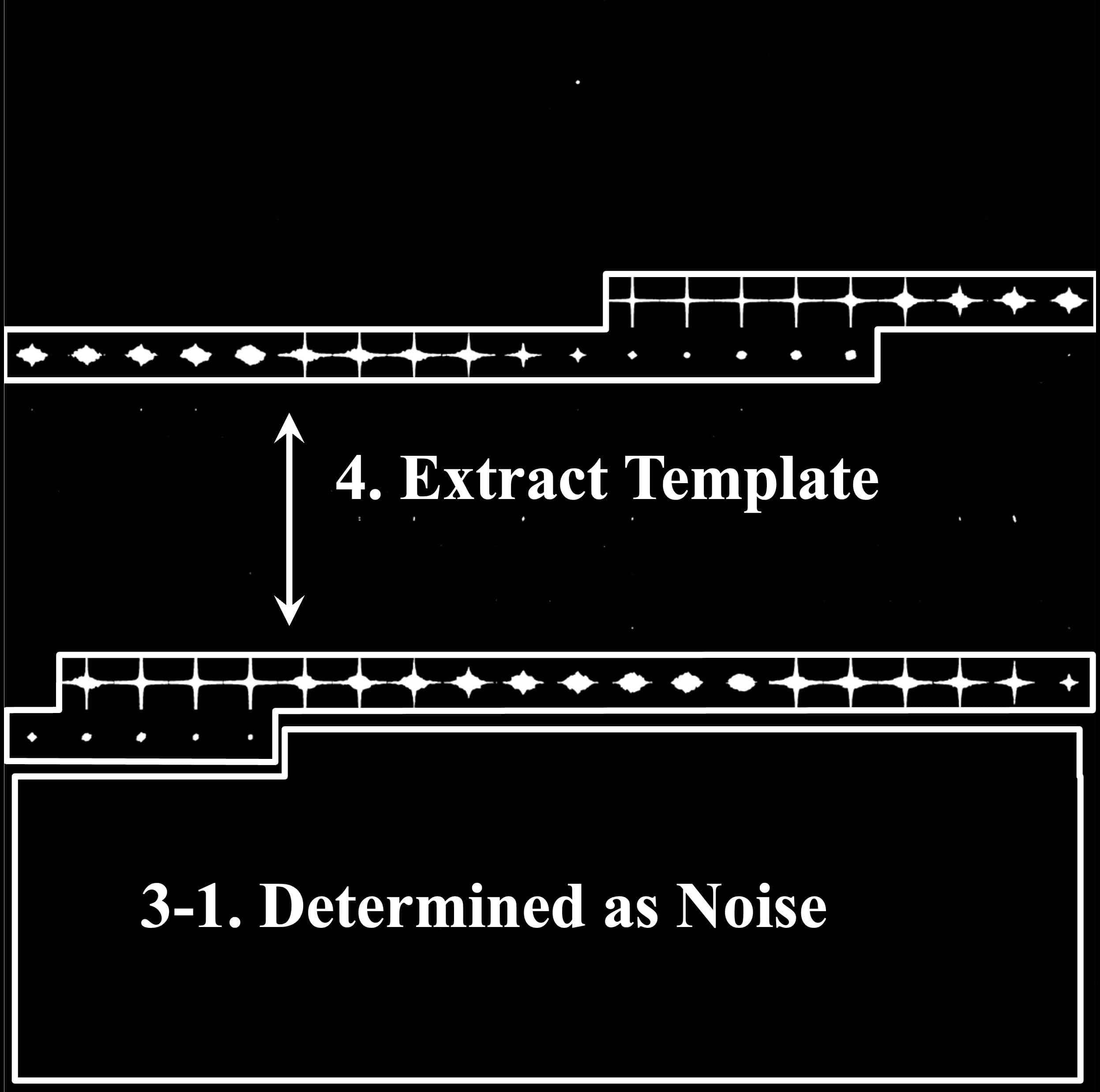}\label{fig:demo:cluster-2}}
    \caption{Depicting the procedure discovery and modeling process based on per-query changes of the \texttt{GeoDA-2} attack.}
    \label{fig:demo:cluster}
\end{figure}

\begin{table*}[tbp]
\footnotesize
\centering
\caption{Procedures Discovered from Analyzing the Traces of 11 Attack Variants.}
\label{tab:procedures}
\begin{tabular}{clll}
\toprule
ID          & Procedure           & Discovered From     & Description   \\ \midrule
\PROC{NULL} & \PROC{Empty}        & -                   & A query is identical to its previous one, caused by quantization or duplicated queries. \\
\PROC{N1} & \PROC{Noise-1}        & \texttt{HSJ-2}      & The difference between normalized standard Gaussian noise times a large constant.  \\
\PROC{N2} & \PROC{Noise-2}        & \texttt{HSJ-2}      & The difference between normalized standard Gaussian noise times a small constant.  \\
\PROC{N3} & \PROC{Noise-3}        & \texttt{GeoDA-2}    & The difference between non-normalized standard Gaussian noise.  \\
\PROC{N4} & \PROC{Noise-4}        & \texttt{GeoDA-2}    & The difference between non-normalized standard Gaussian noise in the DCT subspace.  \\
\midrule[0.3pt]
\PROC{P1} & \PROC{Pattern-1}      & \texttt{GeoDA-2}    & A gradient update with horizontal and vertical spectral peaks.  \\
\PROC{P2} & \PROC{Pattern-2}      & \texttt{GeoDA-Inf}  & A gradient update with horizontal and vertical spectral peaks (with a high central spectral power). \\
\PROC{P3} & \PROC{Pattern-3}      & \texttt{Square-Inf} & A per-query change with horizontal spectral peaks.              \\
\PROC{P4} & \PROC{Pattern-4}      & \texttt{RayS-Inf}   & A per-query change with vertical spectral peaks (with a high central spectral power).              \\
\PROC{P5} & \PROC{Pattern-5}      & \texttt{RayS-Inf}   & A per-query change with vertical spectral peaks.              \\
\PROC{LS} & \PROC{Pattern-LS}     & -                   & A generic line search step, where two per-query changes have the same unsigned direction.              \\
\PROC{IMG} & \PROC{Pattern-IMG} & -                   & An interpolation step between the clean and perturbed images.  \\
\bottomrule
\end{tabular}
\end{table*}

\subsection{Discovering and Modeling Procedures}
\label{sec:method:enroll:discover}

The first task in implementing \SEA{} is to create the procedure database $\PP$ with known attack procedures. To enable modeling the attack on its \textit{first} incident, we propose a semi-automated process to discover procedures from a single attack trace. Despite this, analysts are free to define their own procedures directly if gained access to the attack's source code.

\paragraph{Discovering Procedures}
Without source code access, we rely on clustering per-query changes in the attack trace to identify procedures, where each cluster corresponds to a candidate procedure, as illustrated in \Cref{fig:demo:cluster-1}. We adopt a segment-and-merge strategy to obtain the clusters without deciding the number of clusters. We present the algorithm sketch below and defer its details to \Cref{app:impl} for brevity.

First, we \emph{segment} the trace into sub-sequences at change points, i.e., points of abrupt variations that signal transitions between different states~\cite{changepoint}. We identify change points by detecting per-query changes that are dissimilar from the previous step. Formally, the change points are steps where the MSE distance of the per-query changes $\bdelta_i$ exceeds a system threshold $\tau_{\mathrm{segment}}$. That is, if $\mathtt{MSE}(\bdelta_i - \bdelta_{i-1}) > \tau_{\mathrm{segment}}$.

Then, we \emph{merge} sub-sequences that are statistically similar. Formally, two sub-sequences are merged if the MSE distance between their central points, denoted as $\bar{\bdelta}_i$ and $\bar{\bdelta}_j$, is below a system threshold $\tau_{\mathrm{merge}}$. That is, if $\mathtt{MSE}(\bar{\bdelta}_i - \bar{\bdelta}_j) < \tau_{\mathrm{merge}}$.

Finally, this process separates the trace into clusters of per-query changes, where each cluster $\CC=\s{\bdelta_{i_1}, \bdelta_{i_2}, ...}$ contains queries (or their per-query changes) determined as coming from the same candidate procedure. For example, \Cref{fig:demo:cluster-1} depicts the merging of two line search sub-sequences. The choice of system thresholds is discussed in \Cref{app:impl}.

\paragraph{Modeling Procedures}
Once clusters representing candidate procedures are identified, the next step is to determine their procedure type and estimate the emission probabilities. 

\textit{Noise procedures} lead to per-query changes that resemble white noise in the spectrum space, exhibiting a near-constant power spectral density. This is because they add i.i.d.\ noise to each query, where per-query changes will only retain the noise difference. We confirm this characterization by binarizing the power spectral density and observing a near-zero $\ell_0$ norm. To define the emission probability of these noise procedures, we perform Kernel Density Estimation (KDE) over the clustered per-query changes. This KDE estimates the pixel-level noise distribution $\mathcal{P}^\prime$, thereby instantiating the emission probability in \Cref{eq:noise}, as illustrated in Step-3 of \Cref{fig:demo:cluster-2}.

\textit{Pattern procedures} exhibit peaks in their average spectral density of per-query changes. For these procedures, we extract pattern templates by averaging the binarized power spectral density of per-query changes within each cluster. These templates instantiate $\btau$ in the emission probability in \Cref{eq:pattern}, as illustrated in Step-4 of \Cref{fig:demo:cluster-2}.

\subsection{Enrolling Procedures and Attacks}
\label{sec:method:enroll:enroll}

Having modeled the candidate procedures, the next task in \SEA{}'s implementation is the enrollment of these procedures and corresponding attacks into the system's databases.

\paragraph{Enrolling Procedures}
Given a set of candidate procedures $\PP^\prime=\s{\PROC{P}_i}_{i=1}^k$, each characterized by its PMF $P_i(\bx)$ derived from the output per-query changes $\CC_i$, we incorporate them into \SEA{}'s procedure database. A naive approach would be enrolling all candidate procedures. However, this could unnecessarily complicate the HMM and potentially impact its generalizability. To address this, we implement a greedy pruning algorithm to select procedures from $\PP^\prime$ for enrollment.

This algorithm evaluates each candidate procedure for its contribution to the HMM's descriptive accuracy. It assesses whether a candidate procedure $\PROC{P}_i$ offers a significant improvement over those already in the database $\PP$, quantified as:
\begin{equation}
	\texttt{gain} \coloneqq \frac{1 - \Prob{\CC_i \given \PROC{P}_i}}{1 - \max_{\PROC{P}\in\PP}\Prob{\CC_i \given \PROC{P}}},
\end{equation}
where $\Prob{\CC_i \given \PROC{P}_j}=\Pi_{\bdelta\in\CC_i}P_j(\bdelta)$ is the likelihood of cluster $\CC_i$ under procedure $\PROC{P}_j$. A candidate procedure is incorporated into the database only if it offers a \texttt{gain} exceeding 10\%.

\paragraph{Enrolling Attacks}
With the procedure database established, we integrate the attacks themselves. This is done by estimating each attack's transition matrix $\TransMat$ from observed traces, similar to fingerprinting each attack in \Cref{sec:method:fingerprint}. \Cref{tab:procedures} showcases the procedures we discovered from known attacks. \Cref{fig:fp-dense} visualizes the transition matrices of these attacks. One could observe how variants of the same attack appear more visually similar to each other than to different attacks.

\begin{figure*}[t]
\vspace{-1em}
\begin{minipage}{0.74\linewidth}
  \centering
  \subfloat[\texttt{ImageNet}]{\includegraphics[width=0.32\linewidth]{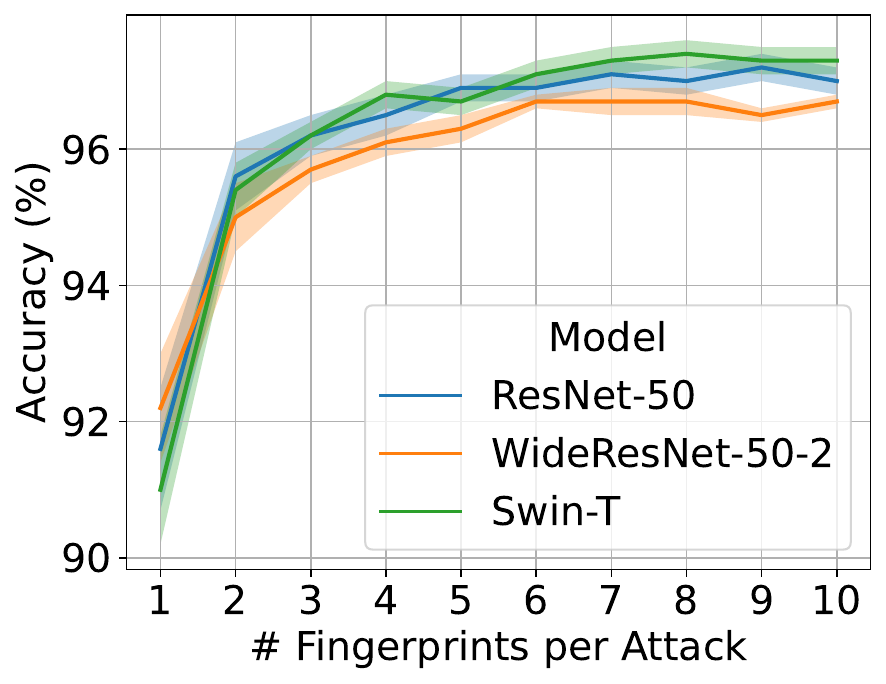}\label{fig:fp-attr-imagenet}}\hfill
  \subfloat[\texttt{CelebA}]{\includegraphics[width=0.335\linewidth]{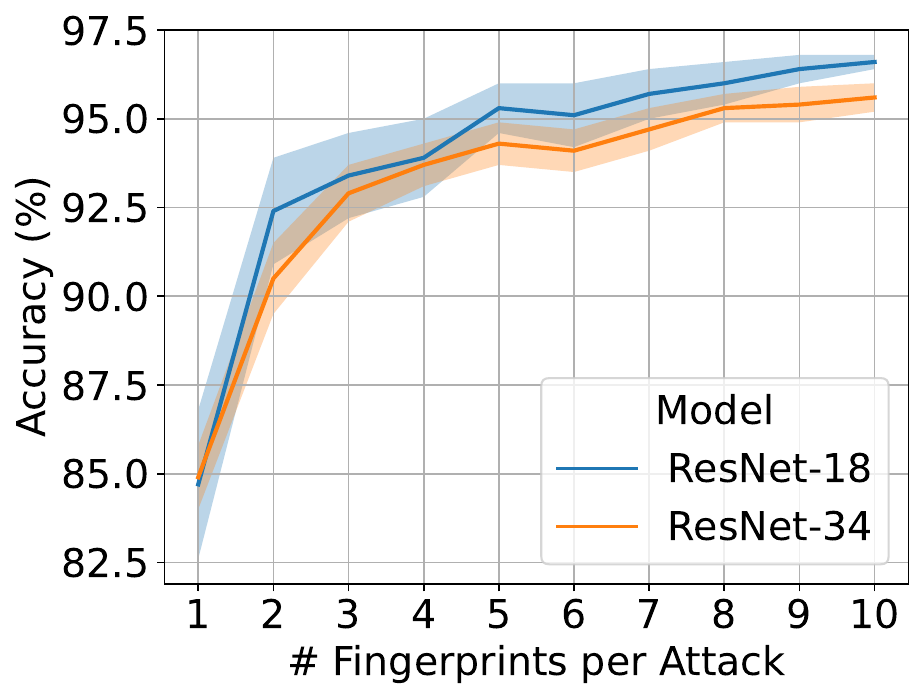}\label{fig:fp-attr-celeba}}\hfill
  \subfloat[\texttt{CIFAR-10}]{\includegraphics[width=0.32\linewidth]{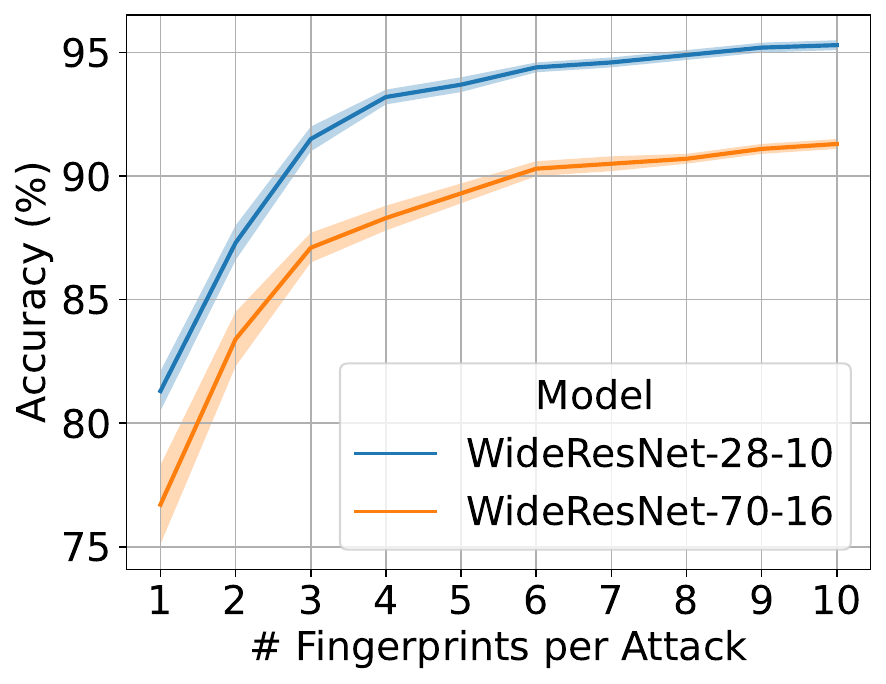}\label{fig:fp-attr-cifar10}}\hfill
  \caption{Top-1 accuracy of refined attribution when more fingerprints are available.}
  \label{fig:fp-attr}
\end{minipage}
\begin{minipage}{0.24\linewidth}
  \centering
  \includegraphics[width=\linewidth]{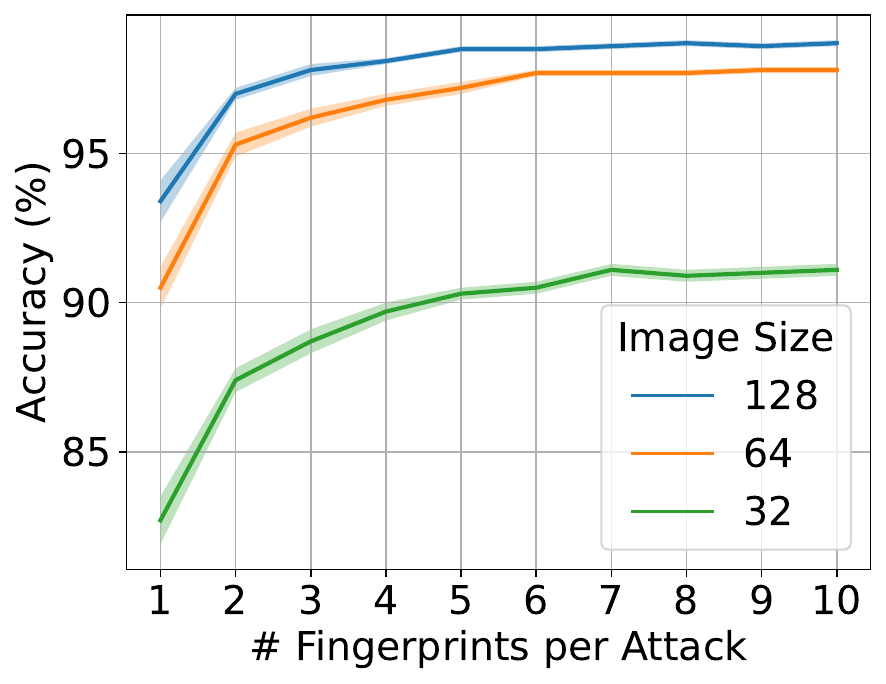}
  \caption{Performance of attribution on down-scaled queries.}
  \vspace{-1.1em}
  \label{fig:fp-attr-rescale}
\end{minipage}
\end{figure*}

\section{Evaluations}
\label{sec:exp}

We evaluate \SEA{} using a variety of attacks, datasets, and models. We outline the experiments and their key findings:
\begin{enumerate}[leftmargin=*]
	\item \textbf{Attribution and Sharing.} \SEA{} can fingerprint an attack on its first incident and accurately recognize its subsequent incidents on multiple datasets and models.
	\item \textbf{Explanation.} \SEA{} can explain the attack's behavior at a per-query granularity, which leads us to discover specific implementation variants across different attack libraries.
	\item \textbf{Robustness.} \SEA{} is robust against several adaptive strategies aiming to evade forensics. It can also accurately recognize such strategies after observing the incidents.
    \item \textbf{Generalization.} \SEA{} can generalize to other domains like task classification. It can fingerprint and recognize textual black-box attacks even if such attacks are similar in spirit.
\end{enumerate}

\subsection{Setup}
\label{sec:exp:setup}

We focus on applying \SEA{} to analyze query-based black-box attacks against image classification tasks.

\paragraph{Datasets and Models}
We consider three image classification datasets: \texttt{ImagetNet}~\cite{imagenet}, \texttt{CelebA}~\cite{celeba}, and \texttt{CIFAR10}~\cite{cifar10}. For \texttt{CIFAR10}, we randomly select 500 images from the test set as the clean images to attack. For \texttt{ImageNet} and \texttt{CelebA}, we randomly select 200 images due to their higher computation costs. Further details of the datasets and models we use are summarized in \Cref{tab:attribution-second} and \Cref{app:exp:dataset}.

\paragraph{Attacks}
We consider 11 black-box attacks as outlined in \Cref{fig:fp-dense}, where \texttt{2} and \texttt{Inf} indicate each attack's optimized norm distance. We use attack implementations from the open-source library \texttt{ART}~\cite{art} and \texttt{BBB}~\cite{bbb}, with default parameters stated in the original paper or recommended by the library. We set the perturbation budget $\epsilon$ to $4/255$ for \LL{\infty} and $10$ for \LL{2} and run each attack until it succeeds or reaches 1000 queries. Further details of these attacks can be found in \Cref{app:exp:attack}.

\paragraph{System Configuration}
We initialize the databases of \SEA{} with only one trace from each attack. This setting resembles a challenging scenario where \SEA{} needs to fingerprint each attack on its \textit{first} incident and recognize their \textit{second} or later incidents. We also evaluate the long-term performance with more traces available. We quantify performance by the accuracy of attributing traces to their ground-truth source attacks. The extraction's precision and recall are measured by mixing the trace with 1000 images randomly chosen from \texttt{ImageNet}.

\subsection{Evaluation of Attribution and Sharing}
\label{sec:exp:attribute}

We first evaluate the performance of \SEA{} in attributing attacks and the stability of the fingerprints it generates.

\paragraph{Attributing the Second Incident}
We start with a challenging scenario where \SEA{} has only observed a \textit{single} incident of each attack and needs to attribute the attack's second incident. This scenario tests \SEA{}'s capability to identify and respond to attacks in fast-paced cybersecurity environments. Specifically, we randomly select one trace from each attack to form the attack database, and then employ \SEA{} to attribute the remaining traces from all attacks. We repeat the process 100 times and report the accuracy in \Cref{tab:attribution-second}. Overall, \SEA{} can attribute attacks with over 90\% Top-1 and 95\% Top-3 accuracy. The Top-1 accuracy has a slight decline on the low-resolution \texttt{CIFAR10} because it contains less information for attribution.

\paragraph{Attributing Further Incidents}
As \SEA{} observes more incidents, it enhances its understanding of each attack's probability distribution. This allows for a refined attribution by comparing a newly acquired fingerprint with each attack's averaged fingerprints, as described in \Cref{eq:fp-cos}. It tests \SEA{}'s capability to improve its performance over long-term forensic analysis. As shown in \Cref{fig:fp-attr}, there is a rapid increase in Top-1 accuracy as more fingerprints are collected, reaching over 95\% for \texttt{ImageNet} and 90\% for \texttt{CIFAR10} datasets.

\begin{table}[t]
\footnotesize
\centering
\caption{Attributing Attacks on Their Second Incident}
\label{tab:attribution-second}
\begin{tabular}{cccc}
\toprule
Dataset                               & Model              & Top-1 Acc (\%) & Top-3 Acc (\%) \\ \midrule
\multirow{3}{*}{\texttt{ImagetNet}} & \texttt{ResNet-50} & 92.5 ± 0.5     & 97.2 ± 0.6     \\
                                   & \texttt{WRN-50-2}  & 92.8 ± 0.8     & 97.6 ± 0.5     \\
                                   & \texttt{Swin-T}    & 91.0 ± 0.3     & 97.1 ± 0.2     \\ \midrule[0.3pt]
\multirow{2}{*}{\texttt{CelebA}}    & \texttt{ResNet-18} & 91.6 ± 1.8     & 93.3 ± 1.3     \\
                                   & \texttt{ResNet-34} & 92.3 ± 1.3     & 95.2 ± 0.9     \\ \midrule[0.3pt]
\multirow{2}{*}{\texttt{CIFAR10}}   & \texttt{WRN-28-10} & 82.6 ± 0.9     & 95.7 ± 0.4     \\
                                   & \texttt{WRN-70-16} & 80.1 ± 0.5     & 92.7 ± 0.3     \\ \bottomrule
\end{tabular}
\end{table}

\begin{figure*}[t]
  \centering
  \subfloat[\texttt{SignOPT-2}]{\includegraphics[width=0.7\linewidth]{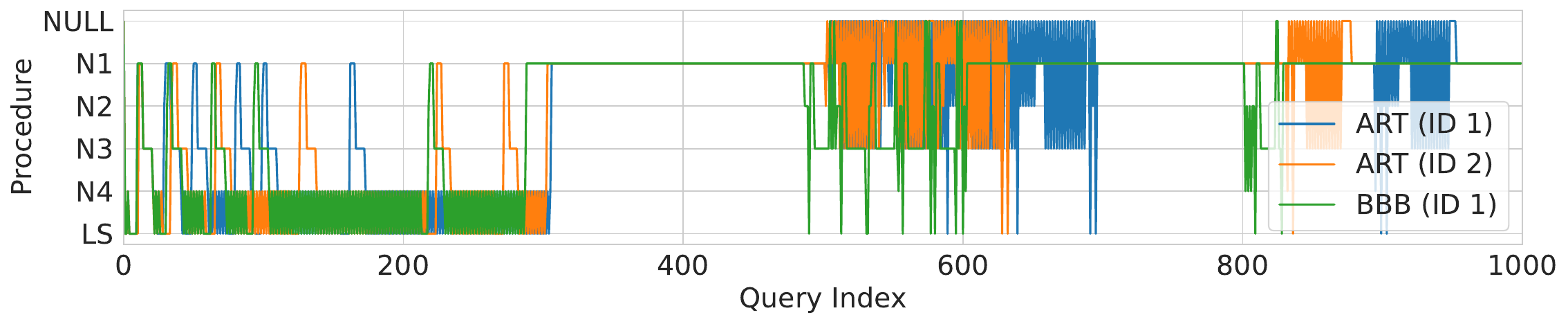}\label{fig:explain:opt}}
  \subfloat[\texttt{Square-2}]{\includegraphics[width=0.29\linewidth]{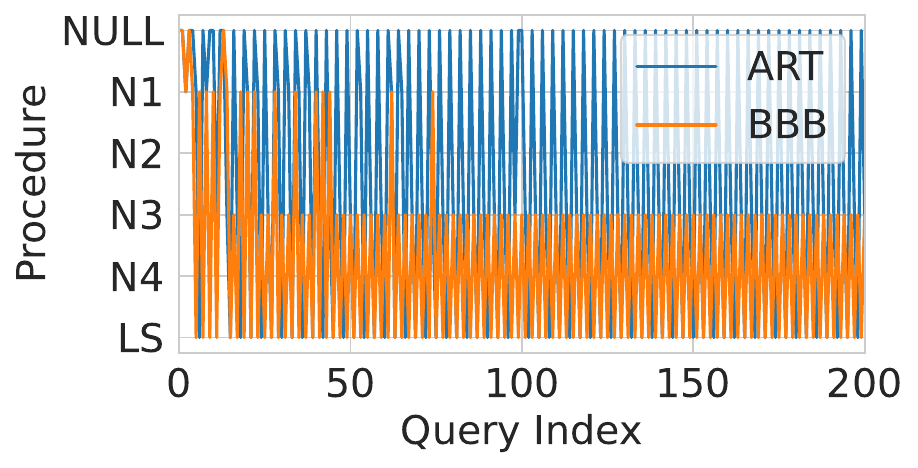}\label{fig:explain:square}}
  \caption{Decoded behavior of attacks from \texttt{ART} and \texttt{BBB}. \textbf{Left:} Traces of two different inputs from \texttt{ART} exhibit consistent behavior. But traces of the same input (ID 1) from \texttt{ART} and \texttt{BBB} exhibit discrepancies after the 500th query. \textbf{Right:} Two traces of the same input from \texttt{ART} and \texttt{BBB} exhibit different behaviors almost everywhere.}
\end{figure*}

\paragraph{Fingerprint Stability}
\SEA{}'s utility of intelligence sharing depends on its ability to generate stable and distinguishable fingerprints from an attack's different incidents. To evaluate this, we generated fingerprints from 200 randomly sampled traces for each attack on \texttt{ImageNet} and \texttt{ResNet-50}. It simulates repeated independent observations of the same attack, where each incident should consistently result in a similar fingerprint. \Cref{fig:fp-all} in \Cref{app:visualize:fingerprint} shows that \SEA{} can produce stable fingerprints for each attack with a narrow confidence interval. The $t$-SNE visualization in \Cref{fig:fp-tsne} shows clear clusters of fingerprints. Interestingly, variants of the same attack appear closer to each other than different attacks. We also show that an attack's fingerprints can generalize across different victim models with a minor performance drop in \Cref{app:transfer}.

\subsection{Case Studies of Explanation}
\label{sec:exp:explain}

We then demonstrate \SEA{}'s capability to provide explanations by decoding the behavior of a trace's underlying attack. This feature is vital for forensic analysts to intuitively understand the attack's behavior and validate the attributions.

To evaluate this, we analyze the behavior of the same attack across two open-source libraries, \texttt{ART} and \texttt{BBB}. This comparison helps to identify discrepancies in different implementations of the same attack. We randomly select a clean image from \texttt{ImageNet} and generate attack traces using both libraries. \SEA{} is then applied to decode the underlying procedure sequence of each trace, as described in \Cref{sec:method:match}. This identifies two attacks, each exhibiting different behaviors in these libraries despite identical hyper-parameters. Further inspection of this discrepancy leads us to discover 3 bugs in the \texttt{ART} library.

\paragraph{Explaining \texttt{SignOPT} Attack}
We randomly sample a clean image and generate attack traces using the \texttt{SignOPT-2} attack from both libraries. \SEA{} successfully attributes both traces to the \texttt{SignOPT-2} attack and decodes the procedure sequences.

In \Cref{fig:explain:opt}, the initial 500 queries exhibit consistent behaviors, with a frequent transition between \PROC{N4} and \PROC{LS} ending with a constant \PROC{N1}. This behavior aligns with the attack's algorithm, which employs noise sampling and binary search to initialize the adversarial direction. However, discrepancies emerge after the 500th query, where \texttt{ART} shows a distinct transition to \PROC{NULL}, indicating duplicated queries. After examining its source code, we trace back to the following bug:
\begin{lstlisting}[language=Python]
if (not is_adv(x) and targeted) or (is_adv(x) and not targeted):
   ...  # do something
\end{lstlisting}
where the same query request \texttt{isadv(x)} is executed twice when \texttt{targeted = False}. After fixing this bug, the updated attack trace exhibits the same behavior as \texttt{BBB}.

\paragraph{Explaining \texttt{Square} Attack}
Similarly, we repeat the above process using the \texttt{Square-2} attack from both libraries. \SEA{} successfully attributes both traces to the \texttt{Square-2} attack and decodes the procedure sequences. In \Cref{fig:explain:square}, both traces exhibit a consistent behavior except \texttt{ART}'s frequent transition to \PROC{NULL}. After examining this discrepancy, we trace back to two bugs, where prior query results could have been reused under certain conditions. Detailed analysis can be found in \Cref{app:exp:square}.
The attack trace from \texttt{ART} resembles \texttt{BBB} only after we have fixed both of these two bugs.

\paragraph{Takeaway}
These case studies demonstrate \SEA{}'s ability to provide detailed explanations of attack behavior. In forensic analysis, such an explainability can provide insights into the attack's toolchain and identifiable artifacts, a crucial feature beyond mere attack attribution. This is another benefit of examining the attack's progression instead of the final adversarial example, which cannot encode such artifacts. We reported these bugs to \texttt{ART}, and one of them has been fixed.

\subsection{Ablation Studies}
\label{sec:exp:ablation}

We conduct several ablation studies to evaluate the attribution under different thresholds and corner settings.

\paragraph{Attributing Incomplete Traces}
Our trace extraction uses a fixed threshold $r=0.5$ in \Cref{sec:method:extract}. We now evaluate the extraction and attribution with different values. As shown in \Cref{tab:extract}, while a lower threshold of $r=0.4$ results in less precise extraction, a higher threshold of $r=0.6$ reduces the recall. We thus choose $r=0.5$ for high recall with 100\% precision. We then evaluate the overall performance across various thresholds and aggregate over attacks. \Cref{tab:incomplete} shows that higher thresholds lead to the extraction of incomplete traces, which in turn slightly degrades the attribution performance.

\begin{table}[t]
\footnotesize
\centering
\caption{Trace Extraction with Different $r$ Thresholds}
\label{tab:extract}
\begin{tabular}{lrrrrrr}
\toprule
\multicolumn{1}{c}{Attack} & \multicolumn{3}{c}{Precision (\%)} & \multicolumn{3}{c}{Recall (\%)} \\
      \multicolumn{1}{c}{$r$}             & \multicolumn{1}{c}{0.4}      & \multicolumn{1}{c}{0.5}     & \multicolumn{1}{c}{0.6}   & \multicolumn{1}{c}{0.4}      & \multicolumn{1}{c}{0.5}     & \multicolumn{1}{c}{0.6} \\ \midrule
\texttt{HSJ-2} & 98.9  &  100.0  & 100.0  &   99.9 &  99.8 &   94.3 \\
\texttt{HSJ-Inf} & 98.9  &  100.0  & 100.0  &   99.8 &  99.7 &   82.8 \\
\texttt{GeoDA-2} & 98.9  &  100.0  & 100.0  &  100.0 &  94.0 &   88.6 \\
\texttt{GeoDA-Inf} & 98.8  &  100.0  & 100.0  &   99.5 &  99.4 &   99.2 \\
\texttt{Square-2} & 98.7  &  100.0  & 100.0  &   90.0 &  90.0 &   90.0 \\
\texttt{Square-Inf} & 95.3  &  99.9   & 100.0  &   98.5 &  98.5 &   98.5 \\
\texttt{SignOPT-2} & 98.1  &  99.9   & 100.0  &  100.0 & 100.0 &  100.0 \\
\texttt{Bound-2} & 98.9  &  100.0  & 100.0  &   99.9 &  99.9 &   99.9 \\
\texttt{ECO-2} & 98.9  &  100.0  & 100.0  &  100.0 & 100.0 &  100.0 \\
\texttt{NES-2} & 98.8  &  100.0  & 100.0  &   82.5 &  77.4 &   73.3 \\
\texttt{RayS-Inf} & 98.9  &  100.0  & 100.0  &  100.0 & 100.0 &  100.0 \\ \bottomrule
\end{tabular}
\end{table}

\begin{table}[t]
\footnotesize
\centering
\caption{Performance of Attribution on Incomplete Traces}
\label{tab:incomplete}
\begin{tabular}{ccccc}
\toprule
\multirow{2}{*}{$r$} & \multicolumn{2}{c}{Extraction (\%)} & \multicolumn{2}{c}{Attribution (\%)} \\
                        & Precision  & Recall   & Top-1 Acc & Top-3 Acc \\ \midrule
0.5 & 100.0 ± 0.0 & 96.2 ± 0.2 & 92.5 ± 0.5 & 97.2 ± 0.6 \\
0.6 & 100.0 ± 0.0 & 93.3 ± 0.1 & 91.2 ± 1.2 & 96.0 ± 0.8 \\
0.7 & 100.0 ± 0.0 & 92.5 ± 0.6 & 90.2 ± 1.2 & 95.3 ± 0.8 \\
0.8 & 100.0 ± 0.0 & 90.1 ± 0.9 & 88.3 ± 1.1 & 94.0 ± 0.7 \\
0.9 & 100.0 ± 0.0 & 88.7 ± 1.2 & 83.9 ± 1.0 & 91.1 ± 0.7 \\ \bottomrule
\end{tabular}
\end{table}

\paragraph{Reducing Storage Cost}
Storing history queries introduces an inevitable storage cost. \SEA{} allows for storing down-scaled queries with a minor impact on the attribution performance. To demonstrate this, we evaluate attribution on \texttt{ImageNet} and \texttt{ResNet-50}, with queries down-scaled to $128\times128$, $64\times64$, and $32\times32$. As shown in \Cref{fig:fp-attr-rescale}, smaller image sizes have a clear impact on the performance. This is because the lossy compression inevitably makes the trace harder to distinguish. The availability of more fingerprints partially mitigates this effect, where even $32\times32$ can preserve 90\% Top-1 accuracy. We also measured runtime latency in \Cref{app:runtime}.

\paragraph{Attributing Unknown Attacks}
While the primary objective of \SEA{} is to attribute the attack trace to the most probable \emph{known} attack as we have discussed in \Cref{sec:overview:threat}, one could modify \SEA{} to recognize \emph{unknown} attacks (i.e., attacks that are not similar enough to any of the known attack) by checking if their similarity in \Cref{eq:fp-cos} to known attacks is below a threshold $\tau_\mathrm{un}$. To evaluate this, we remove each attack from the database and check if \SEA{} can attribute it to unknown. For $\tau_\mathrm{un}=0.6$, \SEA{} achieves 46.1\% Top-1 and 96.7\% Top-3 accuracy, where the confusion arises among similar attacks like \texttt{HSJ-2/HSJ-Inf}. Dissimilar attacks like \texttt{Square-2} achieve 100\% Top-2 accuracy; the confusion stems from the inherent ambiguity of attributing unknown attacks when they resemble the behavior of known attacks.

The ability of attributing unknown attacks further validates the benefits of modeling attacks through procedures. As long as an unknown attack's changes map to similar distributions in the procedure database, \SEA{} can describe their patterns in the space of known procedures and choose a top-similar attack. Once this unknown attack’s pattern is added to the database, \SEA{} ensures that it becomes the new top-similar choice.

\subsection{\SEA{} for Textual Black-box Attacks}
\label{sec:exp:text}

\SEA{} generalizes to textual black-box attacks, which perturb an input at the word level~\cite{textfooler,pwws,genetic,pso} or the character level~\cite{textbugger,deepwordbug}, through addition, deletion, and replacement. As these perturbations are discrete and directly extractable from the queries, it is straightforward to treat them as procedures. We thus characterize textual attacks using the Categorical HMM, a discrete HMM that is simpler than what we used to characterize the vision attacks in \Cref{sec:hmm:attack}. 

To demonstrate this approach, we consider 6 black-box textual attacks: \texttt{TextFooler}~\cite{textfooler}, \texttt{PSO}~\cite{pso}, \texttt{PWWS}~\cite{pwws}, \texttt{TextBugger}~\cite{textbugger}, \texttt{Genetic}~\cite{genetic}, and \texttt{DeepWordBug}~\cite{deepwordbug}. These attacks involve a common set of 3 word-level and 3 character-level transformations, where each level contains addition, deletion, and replacement. We use \texttt{difflib} to parse the per-query changes into discrete observations, i.e., the transformation and the edit distance at each step. 

We devise 6 procedures to model the transitions, thus characterizing each attack with a 6$\times$6 matrix, similar to our modeling of vision attacks in \Cref{sec:hmm:attack}. We then apply \SEA{} to the above attacks and perform attribution on the \texttt{SST-2} dataset~\cite{sst} and \texttt{RoBERTa} model~\cite{roberta}, where we collect 800 query sequences for each attack. \SEA{} achieves over 70\% Top-1 and 90\% Top-2 accuracy as it gathers 5--10 fingerprints. We discuss more details and extension to LLMs in \Cref{app:exp:text}.

\subsection{Adaptive Attacks}
\label{sec:exp:adaptive}

We consider adaptive attacks that are aware of our system and attempt to evade the analysis. Given that \SEA{} conducts its forensic analysis post-attack, it possesses the intrinsic capability to learn from and integrate new adaptive strategies into its database. To simulate this scenario, we focus on the adaptive attack's \textit{first} and \textit{second} incidents. This allows us to evaluate \SEA{}'s effectiveness \textit{before} and \textit{after} it has observed and reacted to a particular adaptive strategy.

\paragraph{Sending Dummy Queries}
A straightforward obfuscation strategy is hiding the attack trace within a large number of dummy queries, whose predictions are ignored. Such dummy queries are correlated with the final adversarial example so \SEA{} will have to include them in the attack trace.

A practical instance of this strategy is the \textit{random duplication of queries}. Our explainability evaluation in \Cref{sec:exp:explain} incidentally covers this scenario, where the \texttt{ART} library generates duplicate queries in the \texttt{SignOPT} and \texttt{Square} attacks. \SEA{} can accurately attribute and explain these attacks, even when they have included duplicated queries.

We also consider dummy queries \textit{perturbed by Gaussian noise}. Here, each batch of queries $\s{\bx_t}$ is randomly mixed with its noisy counterpart $\s{\bx_t+\boldsymbol{\eta}_t}$, where $\boldsymbol{\eta}_t\sim\mathcal{N}(\mat{0}_d, \sigma^2\mat{I}_d)$ is the added Gaussian noise. This method is particularly likely to obfuscate the \texttt{HSJ-2} attack, which inherently uses Gaussian noise to estimate gradients. To evaluate this, we vary the noise level $\sigma$ and analyze the attribution accuracy in \Cref{tab:attribution-dummy}.

On the adaptive attack's first incident, \SEA{} can only attribute around 59\% of such traces to \texttt{HSJ-2}, although a 95\% Top-3 accuracy bounds the confusion. This decrease is expected because the attack has effectively changed its behavior. However, once \SEA{} incorporates the newly observed attack's fingerprint into its database, it recognizes this variant on the second incident with 100\% Top-1 accuracy.

\begin{table}[t]
\footnotesize
\centering
\caption{Attribution of \texttt{HSJ-2} with Dummy Noisy Queries}
\label{tab:attribution-dummy}
\begin{tabular}{lcccc}
\toprule
\multirow{2}{*}{Noise ($\sigma$)} & \multicolumn{2}{c}{First Incident (\%)} & \multicolumn{2}{c}{Second Incident (\%)} \\
 & Top-1 Acc & Top-3 Acc & Top-1 Acc & Top-3 Acc \\ \midrule
0.00  &  93.3 ± 0.9  &  98.9 ± 0.1 & 100.0 ± 0.0  & 100.0 ± 0.0      \\
0.05  &  58.4 ± 3.3  &  95.8 ± 0.7  & 100.0 ± 0.0  & 100.0 ± 0.0      \\
0.10  &  56.1 ± 3.1  &  93.7 ± 0.7  & 100.0 ± 0.0  & 100.0 ± 0.0      \\
0.15  &  59.1 ± 3.9  &  94.3 ± 0.8  & 100.0 ± 0.0  & 100.0 ± 0.0      \\
0.20  &  58.5 ± 4.0  &  94.4 ± 0.7  & 100.0 ± 0.0  & 100.0 ± 0.0      \\ \bottomrule
\end{tabular}
\end{table}

\begin{table}[t]
\footnotesize
\centering
\caption{Attribution of \texttt{GeoDA-2} with Increased Noise}
\label{tab:attribution-increase-noise}
\begin{tabular}{lcccc}
\toprule
\multirow{2}{*}{Noise ($\sigma$)} & \multicolumn{2}{c}{First Incident (\%)} & \multicolumn{2}{c}{Second Incident (\%)} \\
         & Top-1 Acc & Top-3 Acc & Top-1 Acc & Top-3 Acc \\ \midrule
0.0002  &  97.1 ± 1.4  & 100.0 ± 0.0  & 100.0 ± 0.0  & 100.0 ± 0.0      \\
0.002   &  95.7 ± 2.2  & 100.0 ± 0.0  & 100.0 ± 0.0  & 100.0 ± 0.0      \\
0.02    &  83.2 ± 6.2  & 100.0 ± 0.0  & 100.0 ± 0.0  & 100.0 ± 0.0      \\
0.2     &  80.1 ± 3.1  & 100.0 ± 0.0  & 100.0 ± 0.0  & 100.0 ± 0.0      \\
2.0     &  56.7 ± 8.6  &  99.8 ± 0.2  & 100.0 ± 0.0  & 100.0 ± 0.0      \\ \bottomrule
\end{tabular}
\end{table}

\paragraph{Increasing Noise Magnitude}
Given that \SEA{} focuses on the noise distribution of per-query changes, adaptive attacks may seek to obfuscate their identity by adjusting the noise magnitude used in gradient estimation. To evaluate this, we examine how an increased noise magnitude affects the attribution of a \texttt{GeoDA-2} attack in \Cref{tab:attribution-increase-noise}, where we incrementally amplify the noise magnitude $\sigma$ by up to a factor of $10^4$.

Initially, when \SEA{} encounters the modified \texttt{GeoDA-2} for the first time, there is a noticeable decrease in Top-1 accuracy. This indicates that the adaptive strategy succeeds in obfuscating the attack's identity. However, the effectiveness of this obfuscation is limited due to the high Top-3 accuracy.

After adding this variant's fingerprint to the database, \SEA{} recognizes it with 100\% accuracy. This suggests that while adjusting noise magnitude can provide temporary obfuscation, consistently deceiving \SEA{} remains hard. To effectively evade the system, an attack must alter not just the distribution of individual queries but also its overall behavioral pattern.

\paragraph{Increasing Learning Rates}
Attacks may also seek to obfuscate their identity by increasing the learning rate, which alters the per-query changes. We focus on \texttt{NES-2}, \texttt{HSJ-2}, and \texttt{HSJ-Inf} attacks, as they explicitly control their learning rates.
For \texttt{NES-2}, we gradually increase its learning rate from 1 to 1000, beyond which the attack completely fails. Despite this, \SEA{} consistently demonstrates 100\% precision and recall in extracting the entire query sequence and accurately attributes all modified traces to the \texttt{NES-2} attack.

For \texttt{HSJ-2} and \texttt{HSJ-Inf}, which use binary search to adjust their step sizes, we remove their binary search and apply a fixed learning rate. Surprisingly, \SEA{} maintains a 98\% accuracy as we increase the learning rate to 1000. We confirm by visualization that the attack's behavior is still dominated by a different binary search process. Further removal of this process makes the attack unreasonable and significantly changes its behavior. But once \SEA{} adds this variant's fingerprint to the database, it can recognize the attack with 100\% accuracy.

\paragraph{Applying Image Transformation}
Given that \SEA{} characterizes per-query changes, attacks may attempt to complicate its ability to track such changes by applying transformations to their queries, such as shifting and rotation. Although our threat model focuses on norm-bounded attacks, where such transformations are currently intractable, we show that \SEA{} can effectively recognize this obfuscation without adaptation.

We introduce randomized rotation to each query, with angles $\theta_t$ uniformly distributed between $-10^\circ$ and $10^\circ$. For such queries, our standard extraction leads to around 99\% precision and 65--75\% recall, as detailed in \Cref{tab:transform}. When \SEA{} gathers $k=5$ fingerprints for each attack, it can recognize this technique with over 90\% accuracy. Indeed, attacks that employ this rotation technique essentially form a new category of attacks, which \SEA{} is able to identify thereafter consistently.

\paragraph{Mixed Attacks}
An attacker may also attempt to confuse \SEA{} by mixing two attacks against the same image in one incident. In \Cref{app:mixed-attacks}, we \textit{randomly} mix the attack trace from \texttt{HSJ-2} with each of the other 10 attacks on the same image. We show that once \SEA{} has observed these mixed attacks and added their fingerprints to the database, it can then effectively associate every mixed attack's second incident to its observed first incident with over 90\% Top-3 accuracy.

\begin{table}[t]
\footnotesize
\caption{Extraction and Attribution of the Image Transformation Adaptive Attack when $k$ Traces are Available}
\label{tab:transform}
\centering
\begin{tabular}{lcccc}
\toprule
\multicolumn{1}{c}{\multirow{2}{*}{Attack}} & \multicolumn{2}{c}{Extraction (\%) } & \multicolumn{2}{c}{Attribution (\%)} \\
                        & Precision       & Recall       & $k=1$          & $k=5$          \\ \midrule
\texttt{HSJ-2}          & 99.4 ± 0.3      & 69.2 ± 4.6   & 63.4 ± 2.5     & 90.4 ± 0.9     \\
\texttt{HSJ-Inf}        & 99.1 ± 0.4      & 67.3 ± 4.7   & 70.5 ± 2.2     & 94.0 ± 0.7     \\
\texttt{GeoDA-2}        & 99.2 ± 0.3      & 61.1 ± 5.3   & 67.2 ± 2.8     & 92.7 ± 0.7     \\ 
\texttt{GeoDA-Inf}      & 99.3 ± 0.3      & 67.3 ± 4.7   & 78.3 ± 2.2     & 96.9 ± 0.5     \\ 
\texttt{SignOPT-2}      & 99.2 ± 0.4      & 68.9 ± 4.5   & 61.5 ± 2.4     & 92.1 ± 0.9     \\
\texttt{Square-2}       & 96.5 ± 1.8      & 73.9 ± 4.7   & 70.6 ± 2.4     & 92.7 ± 0.9     \\ 
\texttt{Square-Inf}     & 97.8 ± 1.3      & 75.4 ± 4.8   & 71.5 ± 2.4     & 94.2 ± 0.8     \\ 
\texttt{Boundary-2}     & 99.2 ± 0.4      & 64.3 ± 4.4   & 82.6 ± 2.3     & 93.6 ± 0.4     \\ 
\texttt{ECO-2}          & 99.1 ± 0.4      & 76.6 ± 4.6   & 71.6 ± 2.2     & 93.6 ± 0.9     \\ 
\texttt{RayS-Inf}       & 98.9 ± 0.5      & 45.7 ± 6.1   & 54.8 ± 2.5     & 92.6 ± 0.9     \\
\texttt{NES-2}          & 99.3 ± 0.3      & 73.6 ± 4.7   & 70.2 ± 2.4     & 94.4 ± 0.7     \\  \bottomrule
\end{tabular}
\end{table}

\paragraph{Takeaway}
While adaptive attacks may initially hinder the analysis, they have fundamentally changed the nature of the attack. As a result, \SEA{} begins to recognize them as distinct attack types and effectively fingerprints their subsequent incidents. This leaves attackers at a disadvantage: they cannot retroactively alter the trace once it has been observed. 

\section{Discussions}
\label{sec:discuss}

\paragraph{Limitations}
\SEA{} is a forensic system that aims to analyze attacks after they have occurred. Thus, \SEA{} shares several key limitations with any forensic system. For example, \SEA{} is not applicable when privacy or storage outweighs security, where storing private queries is prohibited. However, sharing attack patterns is unlikely to compromise privacy because they are abstract representations generated based only on attacker queries, and no user data is involved. Moreover, \SEA{} cannot be actively involved when the attack has never been found; this usually means the attack has not caused noticeable damage.

\paragraph{Capability Boundary}
The fundamental assumption of \SEA{} is that the per-query changes are produced by some procedure and can be measured by some probability distribution. For example, changes consisting of Gaussian noise can be characterized by a well-defined PMF, defining a specific procedure. Therefore, our modeling of attacks can only encapsulate attacks that produce changes measurable by some PMF. \SEA{} cannot analyze attacks that can succeed with uncorrelated queries, but such attacks are not used in the literature. \SEA{} cannot analyze transferability attacks, as they would reduce to the open question of robustly distinguishing white-box adversarial examples. We will discuss this issue with more details in \Cref{app:discuss}.

\paragraph{Manual Overhead}
We described two approaches to enroll new attacks in \Cref{sec:method:enroll}, which require manual intervention to analyze source code or model the procedures. Our semi-automated approach to discovering the procedures involves some amount of manual intervention. One author enrolled the attacks in our experiments one trace at a time, where adding a single trace takes a few minutes. Manual overhead arises from visually confirming the clustering results and including the automatically extracted PMF and template in the code. Note that there are no technical challenges to automate this further. We opt not to automate this process to ensure the evaluations are controlled and valid.
\section{Conclusion}
\label{sec:conclusion}

\SEA{} is a security system designed to analyze black-box attacks against ML systems for forensic purposes and enable human-explainable intelligence sharing. It fills the long missing respond stage in the NIST cybersecurity framework for ML security. We demonstrate on a variety of settings that \SEA{} can characterize an attack on its first incident and accurately recognize subsequent incidents. More importantly, the provided attribution is understandable by humans. Future work may explore the application of \SEA{} to different purposes of attacks, such as fingerprinting the purpose of attacks and refining its trace preservation and extraction process.

\section*{Acknowledgements}
We thank all anonymous reviewers for their insightful comments and feedback. This work is partially supported by the DARPA GARD program under agreement number 885000 and the NSF through awards CNS-1942014 and CNS-2247381.

\bibliographystyle{plain}
{\bibliography{bib/defense,bib/attack,bib/misc}}

\FloatBarrier
\appendix
\subsection{More Theoretical Analysis}
\label{app:formal}

\subsubsection*{Modeling Image Interpolation}

In \Cref{sec:hmm:procedure}, we define the template matching score $\Metric$ using Pearson correlation coefficient as follows:
\begin{equation}
	\Metric(\bdelta_t, \btau) \coloneqq \abs{R(\bdelta_t, \btau)},
\end{equation}
where $\bdelta_t$ is the per-query changes and $\btau$ is the template that we extracted from the binarized power spectral density when enrolling new attacks in \Cref{sec:method:enroll:enroll}.

However, this definition may not cover all scenarios, such as the image line search step that interpolates between the clean image $\bx$ and the candidate adversarial example $\bx^{\prime\prime}$. In this case, the per-query changes $\bdelta$ would correlate with the clean image and the final adversarial example.

To model this procedure, we let $\btau\coloneqq\bx^\prime$ and follow the same definition as other pattern procedures. This way, we can characterize procedures that produce changes correlated with the adversarial example. This refers to the \PROC{IMG} procedure.

\paragraph{Modeling General Line Search}
Some attacks may also apply linear search over the factor $a$ multiplied by a fixed noise $\boldsymbol{\eta}$:
\begin{equation}
\begin{aligned}
	\bx_{t-1} &= \bx + a_1 \cdot \boldsymbol{\eta} \\
	\bx_{t} &= \bx + a_2 \cdot \boldsymbol{\eta},
\end{aligned}
\end{equation}
where the per-query change $\bdelta_t$ would become noise. Instead of matching $\bdelta_t$ to a pre-defined template, we match its linear relationship with the previous per-step change $\bdelta_{t-1}$ using cosine similarity, written as
\begin{equation}
	\Metric(\bdelta_t, \bdelta_{t-1}) 
	\coloneqq \mathtt{cos\text{-}sim}(\bdelta_t, \bdelta_{t-1})
	= \frac{\bdelta_t\cdot\bdelta_{t-1}}{\norm{\bdelta_t}\norm{\bdelta_{t-1}}},
\end{equation}
where both changes are flattened into vectors.

In this way, we can model the general line search procedure \PROC{LS} in the same framework as defined in \Cref{eq:truncated}. The only exception is that the probability will now condition on two previous queries $P(\bx_t|\bx_{t-1},\bx_{t-2})$.

\paragraph{Converting Similarity Score to Probability}
In \Cref{eq:truncated}, we transform the above similarity score to a probability distribution by enforcing a truncated exponential distribution on $\Metric$, defined as
\begin{equation}\label{eq:trunc2}
	P(\bx_t | \bx_{t-1})\coloneqq
	c \cdot \frac{1}{|\mathbb{D}|^{1-\Metric(\bdelta_t, \btau)}},
\end{equation}
where $c$ is the normalization constant. We now explain the intuition behind this definition step by step.

First of all, one may want to regard the similarity score $\Metric$ as an off-the-shelf probability. However, this is not a valid probability measure because it does not sum to one over the entire input space $\DD$.

The second option is dividing this score by $|\DD|=256^d$ so it sums up to one and therefore becomes a valid probability distribution. However, this option would prevent the activation of pattern procedures in the face of other noise procedures because the resulting probability will never approach 1, even with a perfect matching. The same problem applies when normalizing the score with softmax.

Finally, we enforce a truncated exponential distribution on $\Metric$, so its probability concentrates on a perfect matching yet decreases to zero exponentially upon a bad matching. 
One may derive this by following a similar idea as noise procedures, where the probability applies independently to each single pixel. That is, for a given similarity score $\Metric$, we assume each pixel $\s{\delta_i}_{i=1}^d$ of $\bdelta_t$ will have a probability of $256^\Metric/256$. Thus, the probability of $\bdelta_t$ can be defined as
\begin{equation}
	P(\bx_t | \bx_{t-1}) \coloneqq \Pi_{i=1}^d\Prob{\delta_i} 
	= \frac{256^{d\Metric}}{256^d} 
	= |\DD|^{\Metric-1},
\end{equation}
which resembles \Cref{eq:trunc2} after normalization.

\subsection{More Details of System Implementation}
\label{app:impl}

We summarize our algorithm for discovering procedures from a given trace $X$ in \Cref{alg:clustering}. We set the thresholds to $\tau_\mathrm{segment}=500$ and $\tau_\mathrm{merge}=250$, which are chosen empirically from held-out traces to semi-automate the enrollment process. Note that all segmented sub-sequences are included to estimate the PMFs or patterns, and no further selection process is involved. We tried perturbing the thresholds and found the system not sensitive to the threshold values. The performance of this implementation detail is evaluated along with the entire system.

\paragraph{Procedure Coverage}
This approach allows us to discover 12 procedures outlined in \Cref{tab:procedures}. In particular, we observed that procedures and their underlying patterns discovered from one attack have the ability to cover other attacks. For example, the \texttt{P1} procedure discovered from the \texttt{GeoDA-2} attack was able to describe most of the other attacks in \Cref{fig:fp-all}, as evident from the transitions to the \texttt{P1} procedure in other attacks such as \texttt{HSJ-2} and \texttt{HSJ-Inf}. This observation shows that the patterns we discovered from chosen attacks have sufficient coverage, primarily due to the procedure-level similar nature of black-box attacks. As for textual attacks, we have shown that all of them share the same set of procedures in \Cref{app:exp:text}.

%Note that all comparisons are conducted over the per-query change's power spectral density.

\begin{algorithm}[htb]
\caption{Discover Procedures}
\label{alg:clustering}
\begin{algorithmic}[1]
\Require The attack trace $X=\s{\bx_i}_{i=1}^n$.
\Ensure Clusters of sub-traces $\XX$, where each cluster $\CC\in\XX$ contains sub-sequences from the same procedure.

\State $\triangleright$ Get the power spectral density of per-query changes
\State $\s{\bdelta_i}_{i=1}^{n-1} \gets \s{\texttt{FFT2}(\bx_{i+1} - \bx_i)}_{i=1}^{n-1}$

\State $\XX \gets \varnothing$
\Comment Initialize the set of clusters

\State $\CC \gets \varnothing$
\Comment Initialize one cluster

\For{$i \gets 1$ to $n-2$}
    \State $\CC \gets \CC \cup \s{\bdelta_i}$
  	\Comment Procedure continues
  	
    \If{$\mathtt{MSE}(\bdelta_i - \bdelta_{i+1}) > \tau_\textrm{segment}$}
    	\State $\XX \gets \XX \cup \s{\CC}$
        \Comment Procedure completes
    	\State $\CC \gets \varnothing$
    \EndIf
\EndFor

\State $\XX \gets \XX \cup \s{\CC}$
\Comment The last procedure

\For{$i \gets 2$ to $|\XX|$} 
    \State $\s{\bar{\bdelta}_j}_{j=1}^i \gets \sum_{\bdelta\in\XX_j}\bdelta/|\XX_j|$ \Comment Compute means
    \State $\CC^* \gets \argmin_{\CC_j\in\XX_{1:i}} \mathtt{MSE}(\bar{\bdelta}_i-\bar{\bdelta}_j)$ \Comment Find closest
    \If{$\mathtt{MSE}(\bar{\bdelta}_i-\bar{\bdelta}^{*}) < \tau_{\mathrm{merge}}$}
        \State $\CC^* \gets \CC^* \cup \XX_i$ \Comment Merge if close enough
    \EndIf
\EndFor

\State \textbf{return} $\XX$
\end{algorithmic}
\end{algorithm}

\subsection{Datasets and Models}
\label{app:exp:dataset}

We provide more details about the datasets and models we used for evaluation.

\paragraph{ImageNet} We randomly select 200 images of size $224\times224$ from the validation split as the clean image of all attacks we consider. We use three different models, including \texttt{ResNet-50}, \texttt{WRN-50-2}, \texttt{Swin-T}. These models are pre-trained by TorchVision\footnote{\url{https://pytorch.org/vision/stable/models.html}} on the ImageNet dataset, with Top-1 accuracy of 	
80.858\%, 81.602\%, and 81.474\%, respectively.

\paragraph{CelebA} We randomly select 200 images and rescale the images to $224\times224$ from the validation split as the clean image of all attacks. We use two models \texttt{ResNet-18} and \texttt{ResNet-34}. These models are pre-trained on the CelebA dataset to predict the \texttt{Mouth\_Slightly\_Open} and the \texttt{Gender} attribute with accuracy 92.4\% and 95.3\%.

\paragraph{CIFAR10} We randomly select 500 images of size $32\times32$ from the test split as the clean image of all attacks. We use two models \texttt{WRN-28-10} and \texttt{WRN-70-16}. These models are provided by RobustBench~\cite{robustbench} with Top-1 accuracy of 94.78\% and 95.54\%. In particular, the second model attains robust accuracy 84.98\% under \LL{2} norm perturbation $\epsilon=0.5$~\cite{cifar10-model}.

\subsection{Attack Configurations}
\label{app:exp:attack}

We use \texttt{ART} for the first four attacks and \texttt{BBB} for the last four.

\paragraph{HSJ~\cite{hsj}} This is the improved Boundary~\cite{boundary} attack. At each iteration, it refines the perturbation by estimating the gradient direction and using binary search to approach the decision boundary.

\paragraph{GeoDA~\cite{geoda}} This attack is similar to HSJ~\cite{hsj}, but leverages the geometry property of decision boundary and sub-space Gaussian noise to improve the gradient estimation.

\paragraph{SignOPT~\cite{signopt}} This attack solves a zeroth-order optimization problem by estimating the sign of each step.

\paragraph{Square~\cite{square}} This score-based attack adds squares to the image to estimate the loss.

\paragraph{Boundary~\cite{boundary}} This is the first decision-based attack without estimating the gradients. It finds an adversarial example close to the decision boundary and iteratively refines it to mislead the model without needing access to gradients.

\paragraph{ECO~\cite{eco}} This score-based attack optimizes adversarial perturbation by splitting the image into blocks.

\paragraph{RayS~\cite{rays}} This \LL{\infty} decision-based attack reformulates the continuous problem of finding the closest decision boundary into a discrete problem that does not require any zeroth-order gradient estimation.

\paragraph{NES~\cite{nes}} This score-based attack uses natural evolution strategies to improve gradient estimation.

\subsection{Explain the Square Attack}
\label{app:exp:square}

We first outline the sketch of \texttt{ART}'s implementation of the \texttt{Square-2} attack as follows.
\begin{lstlisting}[language=Python]
for i in range(num_iters):
   # Do we continue?
   score = query(x_adv)
   if success(score):
      break
   
   # Get adv noise
   score = query(x_adv)
   noise = get_noise(score)
   x_adv_new = x_adv + noise
   
   # Does it improve?
   score = query(x_adv_new)
   if better(score):
      x_adv = x_adv_new
\end{lstlisting}

We find that the above code will send duplicate queries to the model in two cases:
\begin{enumerate}[leftmargin=*]
	\item The first \texttt{query} is only used for checking if the current adversarial example is already successful. This output can be reused in the next \texttt{query}, where the attack attempts to get the score from the model. This duplication makes around 50\% of the total queries unnecessary.
	\item The last \texttt{query} determines if the new candidate adversarial example could obtain a lower score. If the candidate cannot improve the attack, the adversarial example will not be updated; hence the same output can be reused in the next iteration's first query. This duplication is triggered less frequently than the previous one. 
\end{enumerate}

\subsection{Runtime Latency}
\label{app:runtime}

\SEA{}'s runtime latency has three separate parts: extracting the attack trace, generating a fingerprint, and matching a fingerprint. We measure the latency averaged over 100 \texttt{ImageNet} traces of 1K queries on a Linux server with Intel Xeon CPUs. We opt not to use GPU or CPU parallelization, although most computations are independent for each query and thus highly parallelizable. Extracting the trace from 2K history queries takes 7.41s, which involves data loading and computing the correlation between images. Generating a fingerprint takes 23.01s, which involves data loading and computing the emission probabilities. Matching two fingerprints involves computing their similarity and takes 22.05ms.

\subsection{Attribution on Textual Attacks}
\label{app:exp:text}

We consider 6 black-box textual attacks: \texttt{TextFooler}~\cite{textfooler}, \texttt{TextBugger}~\cite{textbugger}, \texttt{Genetic}~\cite{genetic}, \texttt{PSO}~\cite{pso}, \texttt{PWWS}~\cite{pwws}, and \texttt{DeepWordBug}~\cite{deepwordbug}. These attacks involve a common set of transformations, including 3 word-level and 3 character-level transformations, where each level contains addition, deletion, and replacement. We use \texttt{difflib} to parse the per-query changes into discrete observations, i.e., the transformation and the edit distance at each step. For simplicity, we devise the following 6 procedures to model the transition:
\begin{enumerate}[leftmargin=*]
	\item \texttt{AW - AddWord}: Add a word to the sentence.
	\item \texttt{DW - DelWord}: Delete a word from the sentence.
	\item \texttt{RW - RepWord}: Replace a word with another one.
	\item \texttt{RC - RepChar}: Replace a character with another one.
	\item \texttt{CO - Compose}: A composite of the above.
	\item \texttt{EQ - \ \ Equal}: Same as the previous query.
\end{enumerate}

\begin{figure}[t]
	\centering
	\includegraphics[width=0.7\linewidth]{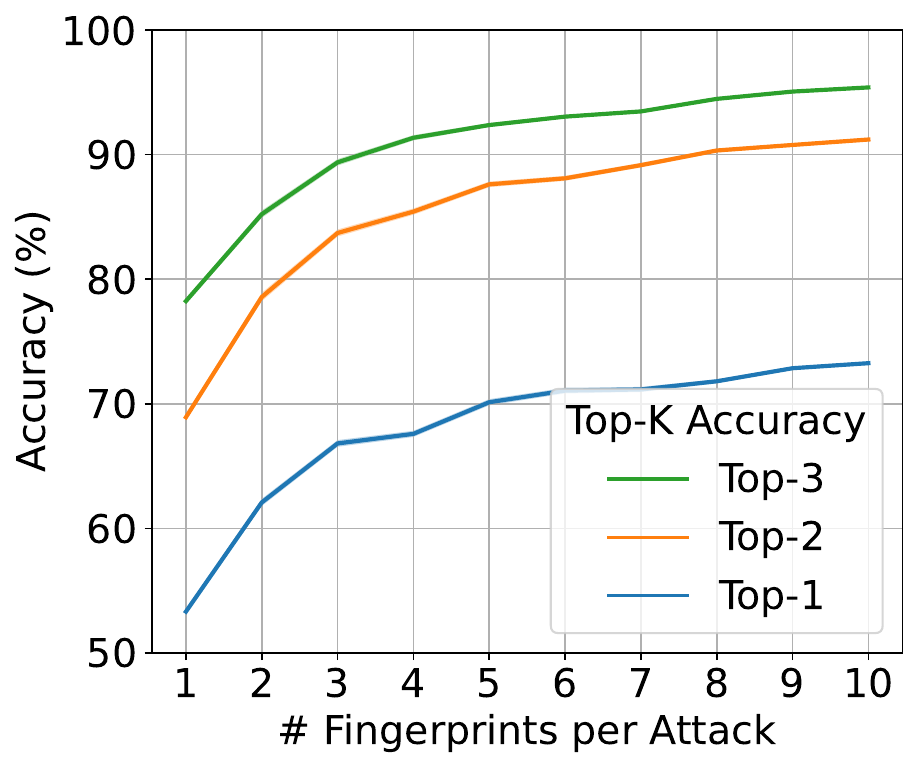}
	\caption{Top-$k$ accuracy of attributing textual attacks.}
	\label{fig:textual}
\end{figure}

As such, each textual attack can be represented by a 6$\times$6 matrix, similar to our modeling of vision attacks in \Cref{sec:hmm:attack}. We then apply \SEA{} to the above attacks and perform attribution on the \texttt{SST-2} dataset~\cite{sst} and \texttt{RoBERTa} model~\cite{roberta}, where we have collected 800 query sequences for each attack. As shown in \Cref{fig:textual}, \SEA{} achieves over 70\% Top-1 and 90\% Top-2 accuracy as it gathers 5--10 fingerprints.

We also observed that the overall attribution performance of textual attacks is slightly worse than that of vision attacks. The reason is that existing textual black-box attacks share the same set of transformations and only differ subtly in their methods for determining the word-replacement order. Thus, \textit{while existing textual attacks are different in their algorithmic methods, they do not exhibit diverse enough behaviors to become distinct attacks.} This interesting observation shows that textual attacks are similar in spirit and might warrant more investigation. We visualize their fingerprints in \Cref{fig:fp-text-all} and find all attacks heavily rely on word replacement.

\paragraph{Extension to LLMs}
\SEA{} can extend to LLM attacks similar to textual attacks in \Cref{sec:exp:text}. For attacks optimizing the same query with measurable changes, \SEA{} can analyze based on the types of per-query changes and their transition probabilities. However, \SEA{} may not directly apply to attacks producing drastically different queries like paraphrasing. In such cases, \SEA{} would need to analyze the traces in the embedding space.

\subsection{Adaptive Attacks: Mixed Attacks}
\label{app:mixed-attacks}

An attacker may attempt to mix two attacks against the same image in one incident. In this case, the attacker is effectively creating a new mixed attack, where separating the two base attacks is hard. The reason is that many attacks share the same queries; it is impossible to attribute an observed query to one base attack versus the other without side information like accounts. Thus, \SEA{} seeks to analyze such a new mixed attack, rather than attempting to differentiate and recognize the base attacks on the same image.

To emulate the most challenging scenario, we \textit{randomly} mix the attack trace from \texttt{HSJ-2} with each of the other 10 attacks on the same image. The mixture preserves the order of queries within each attack, as each attack still produces its queries sequentially. Then, we use \SEA{} to fingerprint and recognize these ensemble traces on their first two incidents. However, it is worth emphasizing that the attack's identity here is \emph{ambiguous} because the attack is no longer \texttt{HSJ-2} or any of the mixed attacks. We measure the attribution of \texttt{HSJ-2} in \Cref{tab:collusion} only for demonstration purposes.
 
Unsurprisingly, when \SEA{} encounters the mixed attack for the first time, most of the attacks are no longer recognized as \texttt{HSJ-2}. The only exceptions are attacks that are originally similar to \texttt{HSJ-2}, such as \texttt{HSJ-Inf} and \texttt{RayS-Inf}. This observation precisely reflects \SEA{}'s design of analyzing the behavior rather than merely the queries, as such mixed attacks no longer share similar behavior as the original attack.

However, once \SEA{} has observed these mixed attacks and added their fingerprints to the database, it can then effectively associate every mixed attack's second incident (to its observed first incident) with over 70\% Top-1 and 90\% Top-3 accuracy.

Finally, the attacker may also attempt to mix the same attack against two different images. However, this mixture is ineffective because the resulting two attack traces will be separated when we extract the attack trace, as these two adversarial examples have a higher correlation with their own corresponding trace, respectively. It is also possible for the attacker to leverage multiple accounts to issue the query, but \SEA{} has never used such information, and the overall attack trace will not change when the attacker switches from one account to multiple accounts. When the attacker wants to mix more base attacks, \SEA{} can automatically obtain their patterns as such attacks are just mixing the queries of base attacks, which \SEA{} can achieve by itself after knowing the base attacks.

\begin{table}[t]
\footnotesize
\caption{Attribution of the Mixed Adaptive Attack}
\label{tab:collusion}
\centering
\begin{tabular}{lrrll}
\toprule
\multirow{2}{*}{Attack} & \multicolumn{2}{c}{First Incident (\%)} & \multicolumn{2}{c}{Second Incident (\%)} \\
                        & Top-1 Acc     & Top-3 Acc     & Top-1 Acc     & Top-3 Acc      \\ \midrule
\texttt{HSJ-2}          & 95.5 ± 1.4    & 99.9 ± 0.0    & 83.8 ± 2.8    & 99.1 ± 0.6       \\
\texttt{HSJ-Inf}        & 23.4 ± 4.2    & 98.0 ± 0.5    & 91.6 ± 2.0    & 99.6 ± 0.2       \\
\texttt{GeoDA-2}        &  9.9 ± 1.3    & 58.3 ± 1.6    & 78.2 ± 2.0    & 97.1 ± 0.6       \\ 
\texttt{GeoDA-Inf}      &  6.8 ± 1.4    & 46.6 ± 2.2    & 92.4 ± 1.2    & 99.1 ± 0.2       \\ 
\texttt{SignOPT-2}      &  2.1 ± 0.7    &  7.5 ± 1.6    & 78.7 ± 2.6    & 96.5 ± 1.0       \\
\texttt{Square-2}       &  0.0 ± 0.0    &  0.0 ± 0.0    & 70.9 ± 3.6    & 97.0 ± 1.4       \\ 
\texttt{Square-Inf}     &  0.0 ± 0.0    &  0.1 ± 0.1    & 74.6 ± 2.9    & 96.3 ± 0.8       \\ 
\texttt{Boundary-2}     &  0.7 ± 0.2    & 52.3 ± 4.5    & 87.3 ± 2.6    & 97.1 ± 1.2       \\ 
\texttt{ECO-2}          &  0.8 ± 0.4    & 11.2 ± 2.3    & 77.3 ± 2.7    & 97.2 ± 1.0       \\ 
\texttt{RayS-Inf}       & 39.9 ± 4.0    & 94.2 ± 1.4    & 79.9 ± 2.3    & 97.8 ± 0.5       \\
\texttt{NES-2}          &  0.1 ± 0.1    &  3.1 ± 0.7    & 92.1 ± 2.2    & 99.5 ± 0.5       \\  \bottomrule
\end{tabular}
\end{table}

\subsection{The Fingerprint's Transferability Across Models}
\label{app:transfer}

For the fingerprints to work in practice, they should exhibit transferability across different models. That means, the fingerprints of an attack should be generally agnostic to the victim models. This transferability is expected-by-design because \SEA{} is designed to characterize an attack's per-query behavior, which is itself model-agnostic due to the nature of black-box attacks.

We evaluate this transferability by matching the fingerprints across three different models in ImageNet. Specifically, we randomly pick 1 and 10 fingerprints for each attack on the source models to form the reference fingerprint, and match the other 100 fingerprints generated for each attack on the target models. The results with 100 repeats are shown in \Cref{tab:transfer-1,tab:transfer-2}. As we can observe, the fingerprints can generalize well across different models and maintain a high accuracy. The minor discrepancy is because some models are more susceptible to attacks, hence the attack may succeed before issuing more queries that demonstrate its full-fledged behavior.

\subsection{More Discussions}
\label{app:discuss}

\paragraph{Benign and Adversarial Misclassifications}
A benign input $\bx^\prime$ may also cause a misclassification. We can differentiate benign and adversarial inputs by their relationship to historical queries. For example, query-based black-box attacks have to send a sequence of visually similar queries to construct the final adversarial example $\bx^\prime$, and SEA extracts such a sequence by identifying queries similar to $\bx^\prime$ in \Cref{sec:method:extract}. However, a benign misclassification $\bx^\prime$ is unlikely to find many similar history queries, otherwise it would be suspicious and thus considered as a potential attack that is worth investigating.

This observation has also been utilized by existing black-box defenses such as Blacklight. Therefore, if we cannot identify many similar history queries to a misclassification $\bx^\prime$, either by Blacklight or SEA’s trace extraction, we can regard $\bx^\prime$ as a benign misclassification instead of an adversarial example.

\begin{table}[t]
\caption{Transferability Accuracy (\# Fingerprints = 1)}
\label{tab:transfer-1}
\centering
\footnotesize
\begin{tabular}{@{}ccccc@{}}
\toprule
\multicolumn{2}{c}{Source \textbackslash{} Target} & \texttt{ResNet-50} & \texttt{WRN-50-2} & \texttt{Swin-T} \\ \midrule
\multirow{2}{*}{\texttt{ResNet-50}} & Top-1 & 91.94 ± 0.68 & 85.13 ± 0.76 & 85.52 ± 0.72 \\
                                    & Top-3 & 97.58 ± 0.29 & 93.56 ± 0.35 & 93.09 ± 0.39 \\ \midrule
\multirow{2}{*}{\texttt{WRN-50-2}}  & Top-1 & 83.85 ± 0.63 & 91.34 ± 0.73 & 91.37 ± 0.72 \\
                                    & Top-3 & 90.59 ± 0.76 & 97.49 ± 0.26 & 97.84 ± 0.21 \\ \midrule
\multirow{2}{*}{\texttt{Swin-T}}    & Top-1 & 84.29 ± 1.00 & 90.41 ± 0.81 & 90.73 ± 0.85 \\
                                    & Top-3 & 92.71 ± 0.86 & 97.03 ± 0.34 & 97.76 ± 0.26 \\ \bottomrule
\end{tabular}
\end{table}

\begin{table}[t]
\caption{Transferability Accuracy (\# Fingerprints = 10)}
\label{tab:transfer-2}
\centering
\footnotesize
\begin{tabular}{@{}ccccc@{}}
\toprule
\multicolumn{2}{c}{Source \textbackslash{} Target} & \texttt{ResNet-50} & \texttt{WRN-50-2} & \texttt{Swin-T} \\ \midrule
\multirow{2}{*}{\texttt{ResNet-50}} & Top-1 & 97.15 ± 0.15 & 89.03 ± 0.17 & 90.01 ± 0.16 \\
                                    & Top-3 & 99.04 ± 0.12 & 95.06 ± 0.15 & 94.90 ± 0.11 \\ \midrule
\multirow{2}{*}{\texttt{WRN-50-2}}  & Top-1 & 87.98 ± 0.13 & 96.84 ± 0.12 & 97.47 ± 0.13 \\
                                    & Top-3 & 90.01 ± 0.25 & 98.85 ± 0.11 & 99.22 ± 0.08 \\ \midrule
\multirow{2}{*}{\texttt{Swin-T}}    & Top-1 & 87.87 ± 0.15 & 96.84 ± 0.13 & 97.58 ± 0.11 \\
                                    & Top-3 & 97.24 ± 0.45 & 98.99 ± 0.11 & 99.32 ± 0.09 \\ \bottomrule
\end{tabular}
\end{table}

\paragraph{Real-World Scenarios}
As we discussed in \Cref{sec:overview:threat}, our threat model applies to scenarios where security outweighs storage and privacy. This usually means the cost of inability to diagnose security breaches is high or even unacceptable. Examples of these scenarios would include:
\begin{itemize}
	\item \textit{Medical Diagnosis:} The attacker may perturb a medical image so the diagnosis model gives a false claim, and then exploits this false evidence to conduct phishing attacks. While storing medical images is costly and highly privacy-sensitive, they are critical for regulatory compliance and potential legal liability.
	\item \textit{Facial Access Control:} The attacker may perturb their facial images to evade an online or on-device access control system. When a subsequent security incident is reported, it is critical that we have all the queries logged to investigate the initial breach and collect legal evidence.
	\item \textit{Spam Email Detection:} When a successful evasion is reported, it is critical for the email service to inspect all history emails, and identify and ban all malicious accounts associated with the attack trace. Without storing such data, it will be hard for us to identify and ban the accounts controlled by the attacker.
\end{itemize}

\paragraph{Generalization to Transferability Attacks}
Our threat model does not work for transferability attacks. Our main consideration is that transferability attacks can leverage the gradient of a surrogate model to generate near-optimal attack queries, which may succeed with only a few queries. Differentiating such queries would reduce to detecting white-box adversarial examples. Unfortunately, this remains an open question given the lessons learned from white-box defenses -- classifying white-box adversarial examples would require a similar capability of designing an adversarially robust white-box defense. We thus decide to exclude this attack from our threat model to avoid over-complicating the problem.

\paragraph{Incorporating Model Responses}
It is worth noting that the model response is a valuable signal to leverage. During our initial exploration, we observed that the model response alone could demonstrate some patterns of how the queries jump back and forth around the decision boundary. However, we eventually decided to rely solely on the input space signal for two reasons: (1) such fingerprints may not work on other models and may even leak the information of private models; (2) since we are dealing with adversarial examples, the model outputs are less trustworthy and stable than input-space differences.

\subsection{More Visualizations}
\label{app:visualize}
\label{app:visualize:fingerprint}

\begin{figure}[t]
    \centering
    \includegraphics[width=\linewidth]{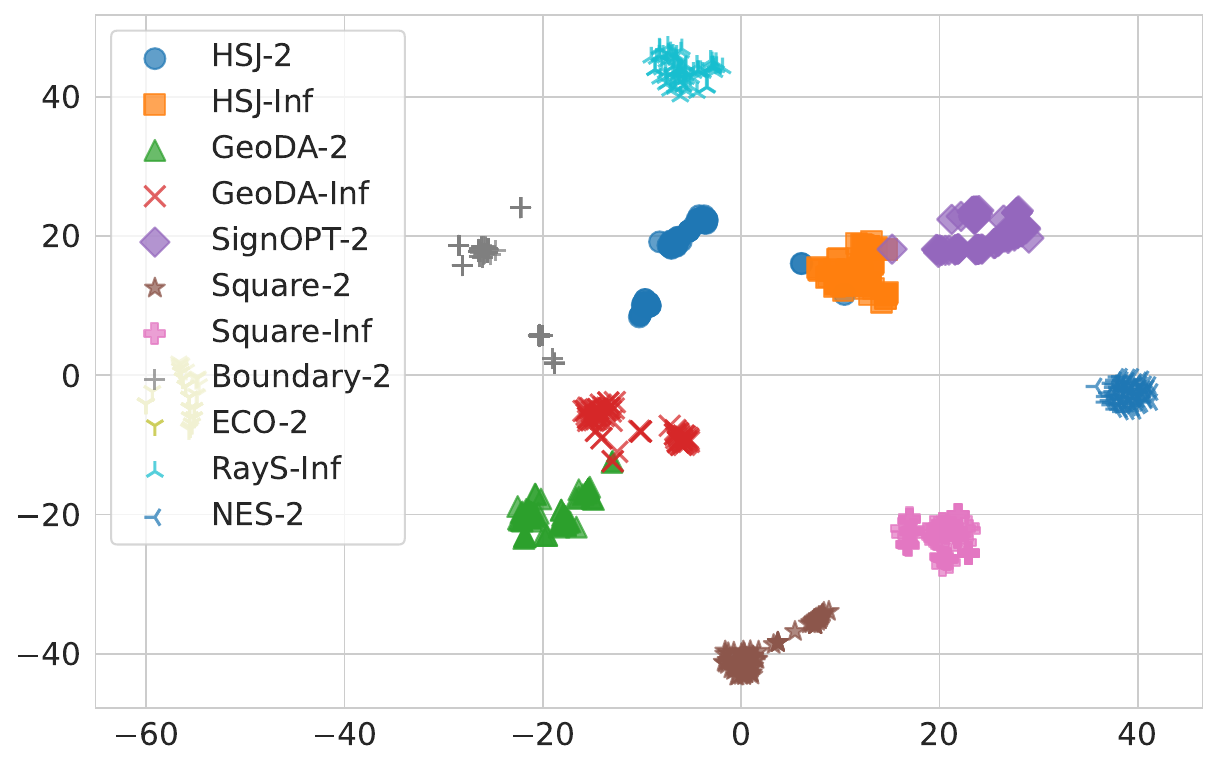}
    \caption{$t$-SNE visualization of each attack's fingerprints.}
    \label{fig:fp-tsne}
\end{figure}

\begin{figure*}[h!]
  \centering
  \captionsetup[subfigure]{labelformat=empty}
    \subfloat[\texttt{TextFooler}]{\includegraphics[width=0.16\linewidth]{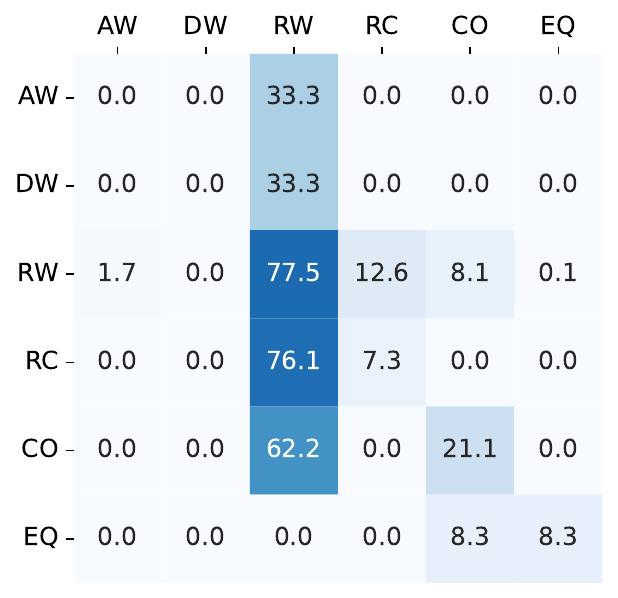}}
  \subfloat[\texttt{TextBugger}]{\includegraphics[width=0.16\linewidth]{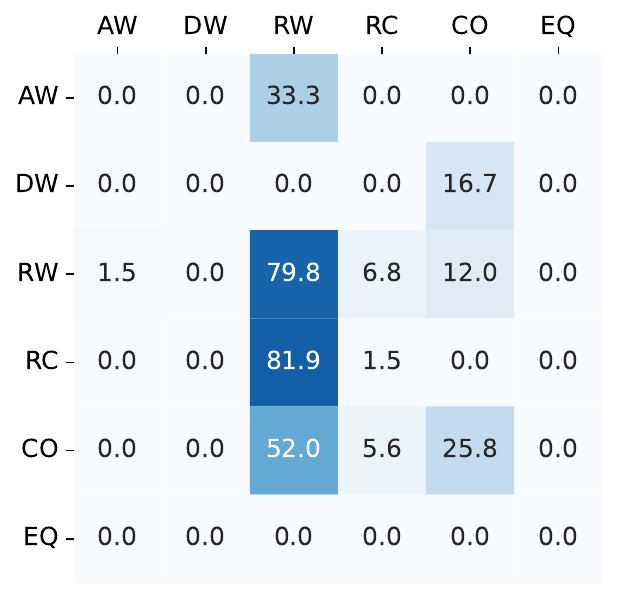}}
  \subfloat[\texttt{DeepWordBug}]{\includegraphics[width=0.16\linewidth]{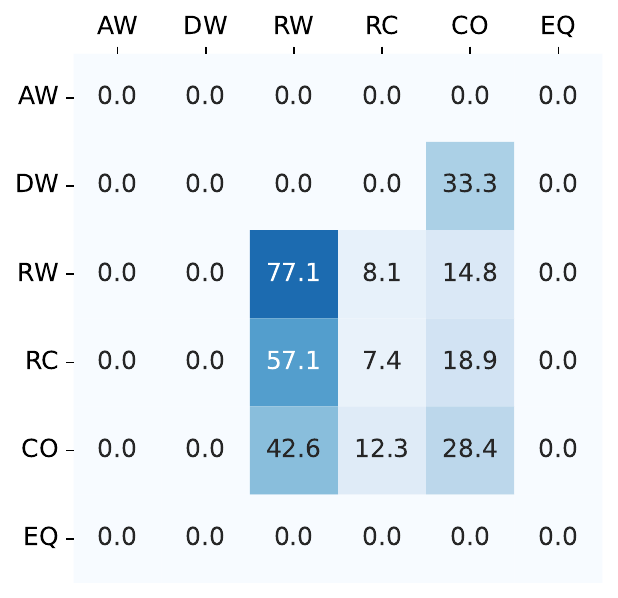}}
  \subfloat[\texttt{Genetic}]{\includegraphics[width=0.16\linewidth]{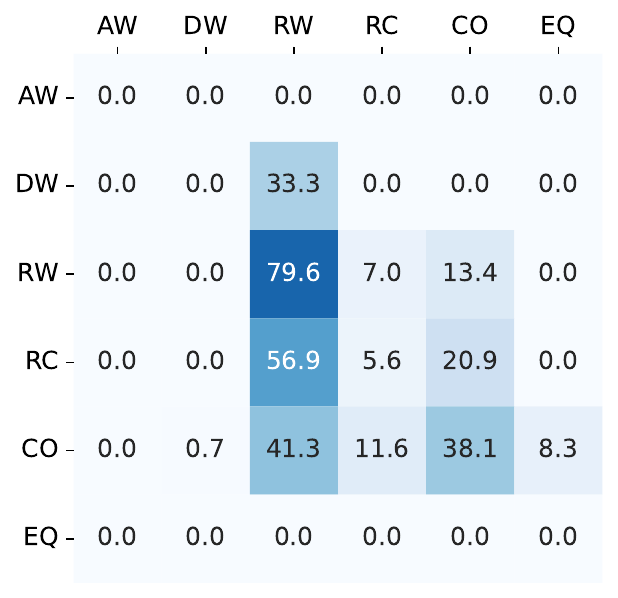}}
  \subfloat[\texttt{PWWS}]{\includegraphics[width=0.16\linewidth]{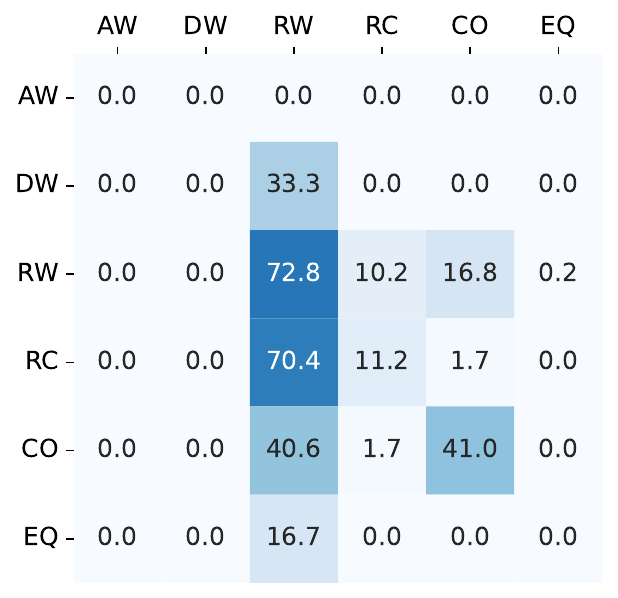}}
  \subfloat[\texttt{PSO}]{\includegraphics[width=0.16\linewidth]{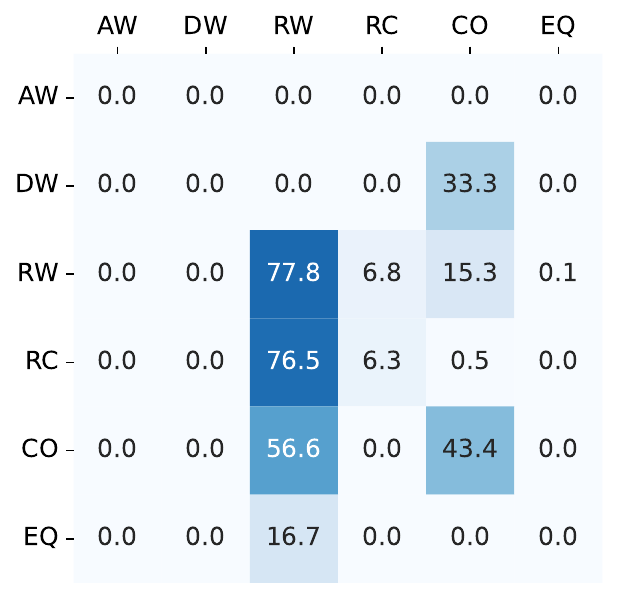}}

  \subfloat[\texttt{TextFooler}]{\includegraphics[width=0.16\linewidth]{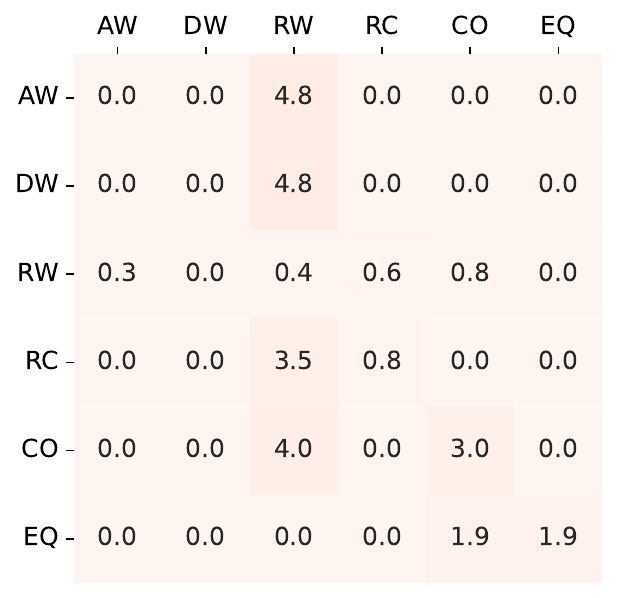}}
  \subfloat[\texttt{TextBugger}]{\includegraphics[width=0.16\linewidth]{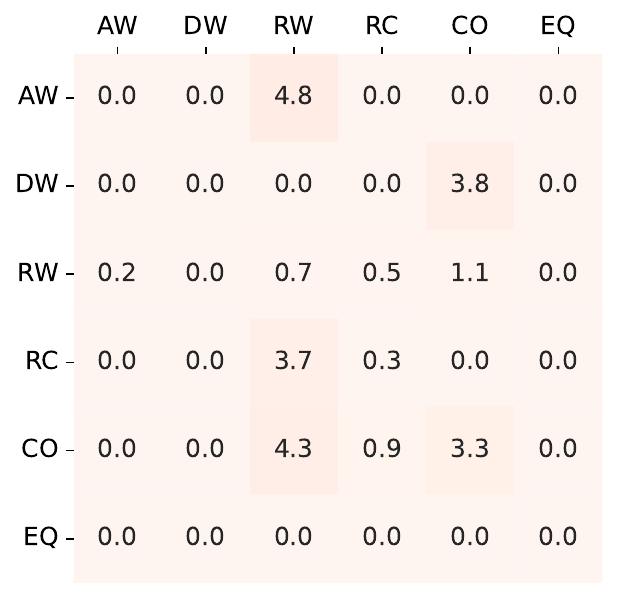}}
  \subfloat[\texttt{DeepWordBug}]{\includegraphics[width=0.16\linewidth]{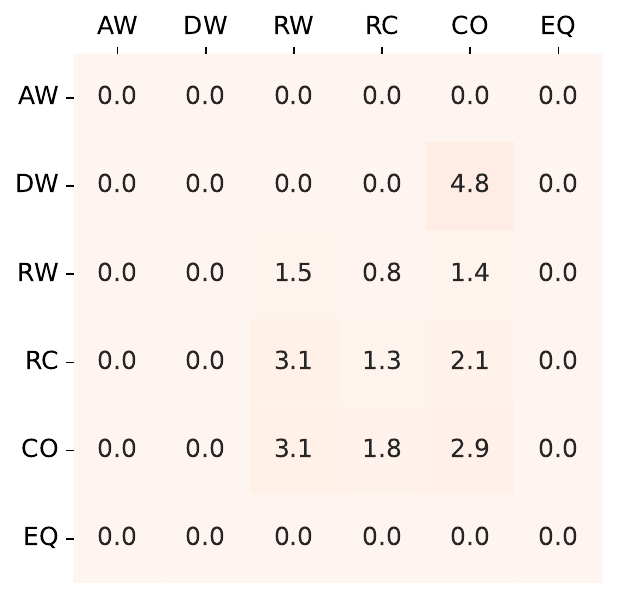}}
  \subfloat[\texttt{Genetic}]{\includegraphics[width=0.16\linewidth]{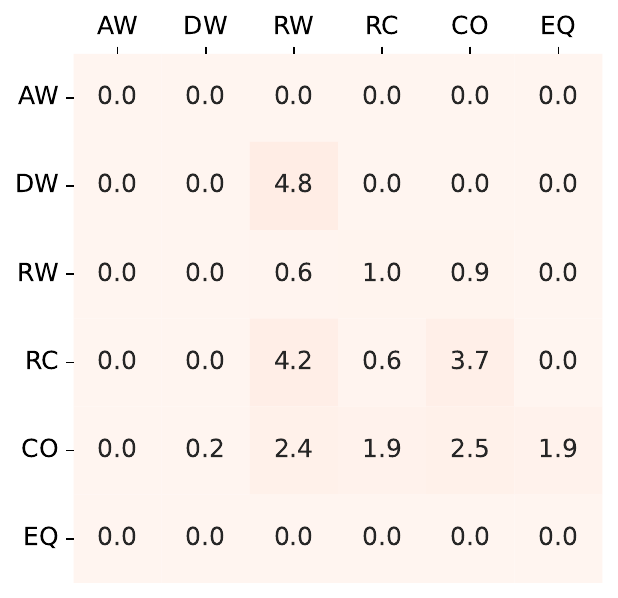}}
  \subfloat[\texttt{PWWS}]{\includegraphics[width=0.16\linewidth]{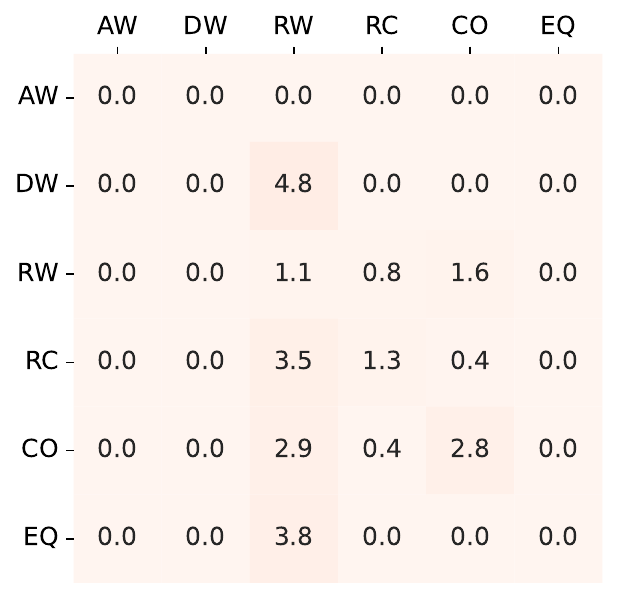}}
  \subfloat[\texttt{PSO}]{\includegraphics[width=0.16\linewidth]{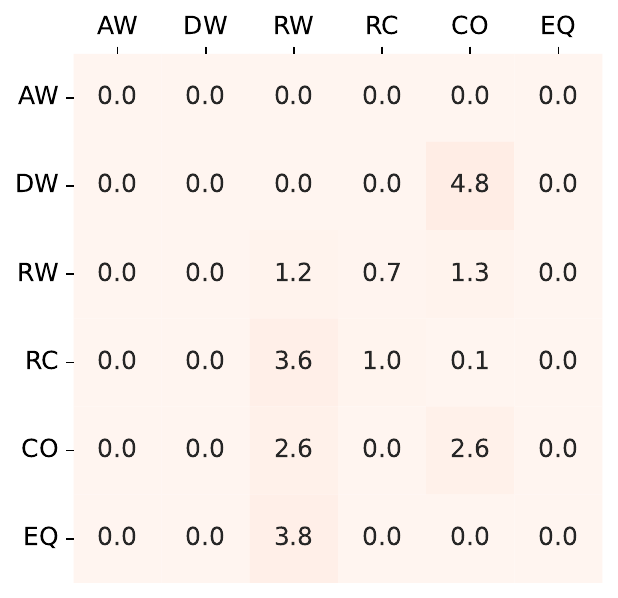}}
  
  \subfloat[\texttt{HSJ-2}]{\includegraphics[width=0.16\linewidth]{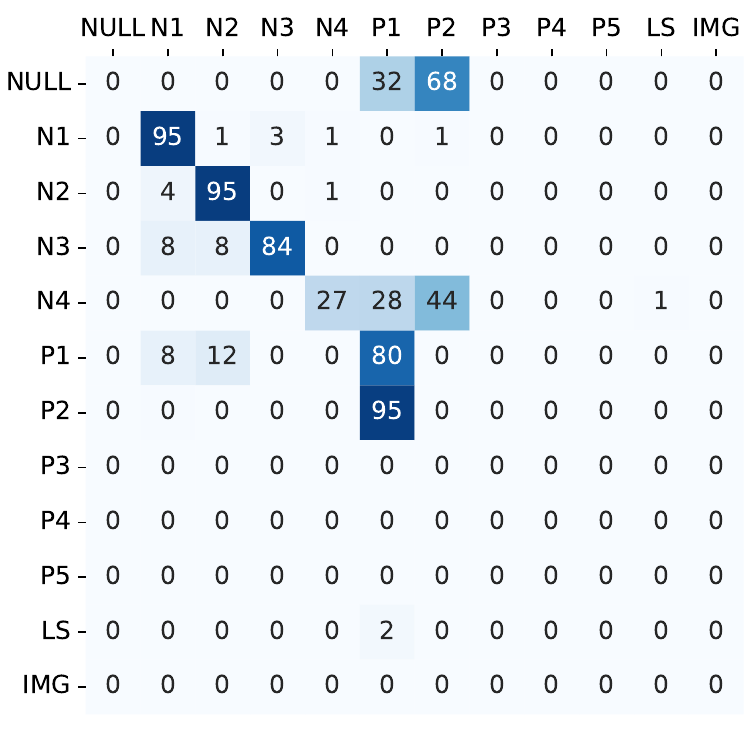}}
  \subfloat[\texttt{HSJ-Inf}]{\includegraphics[width=0.16\linewidth]{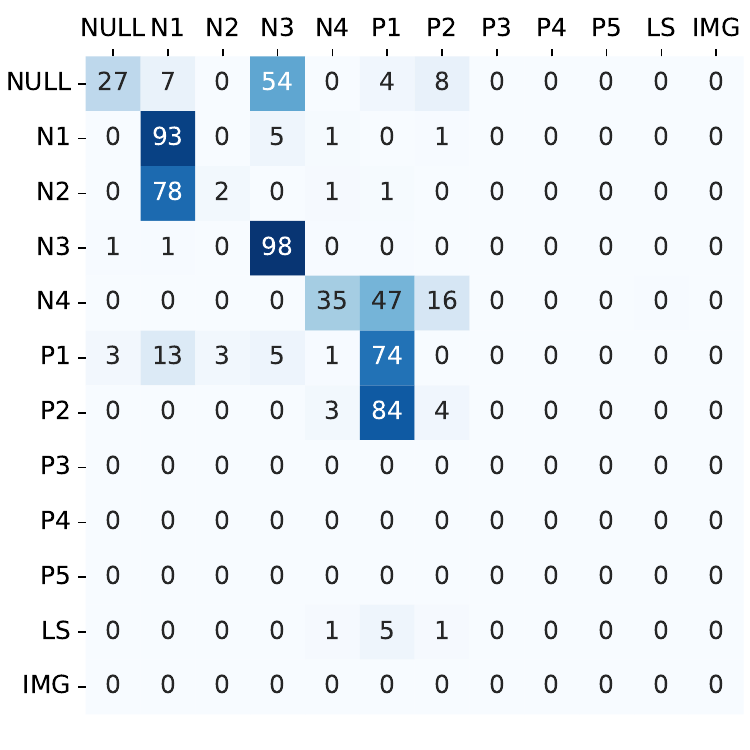}}
  \subfloat[\texttt{GeoDA-2}]{\includegraphics[width=0.16\linewidth]{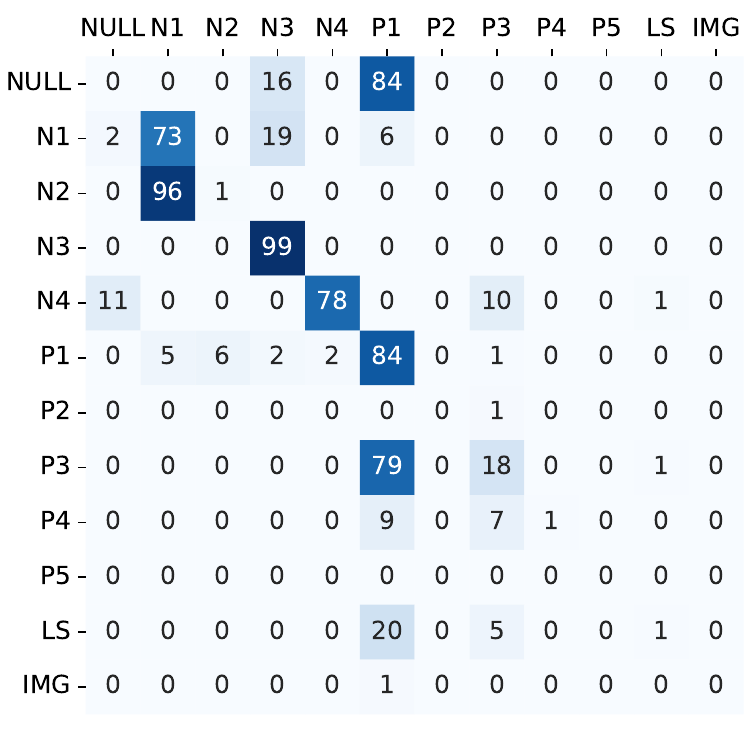}}
  \subfloat[\texttt{GeoDA-Inf}]{\includegraphics[width=0.16\linewidth]{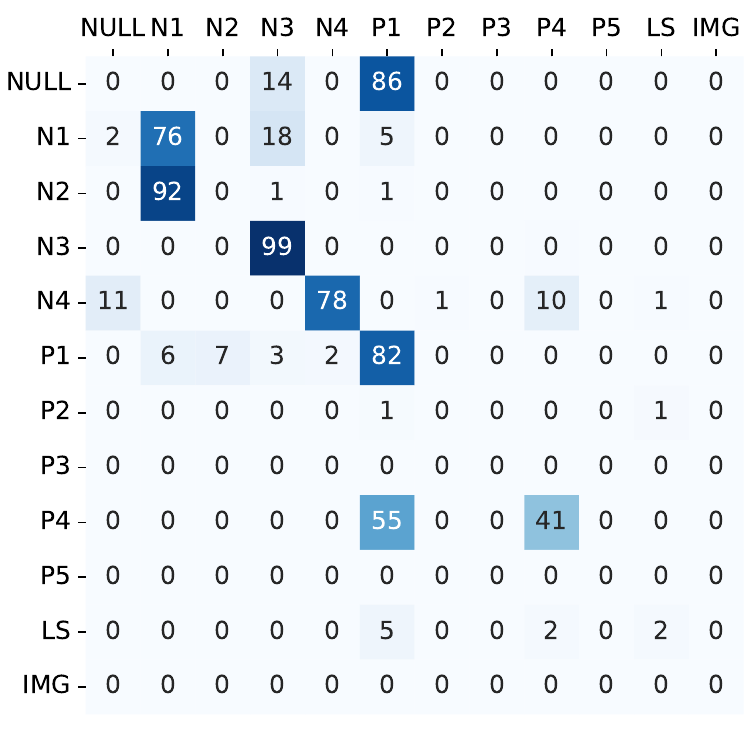}}
  \subfloat[\texttt{Square-2}]{\includegraphics[width=0.16\linewidth]{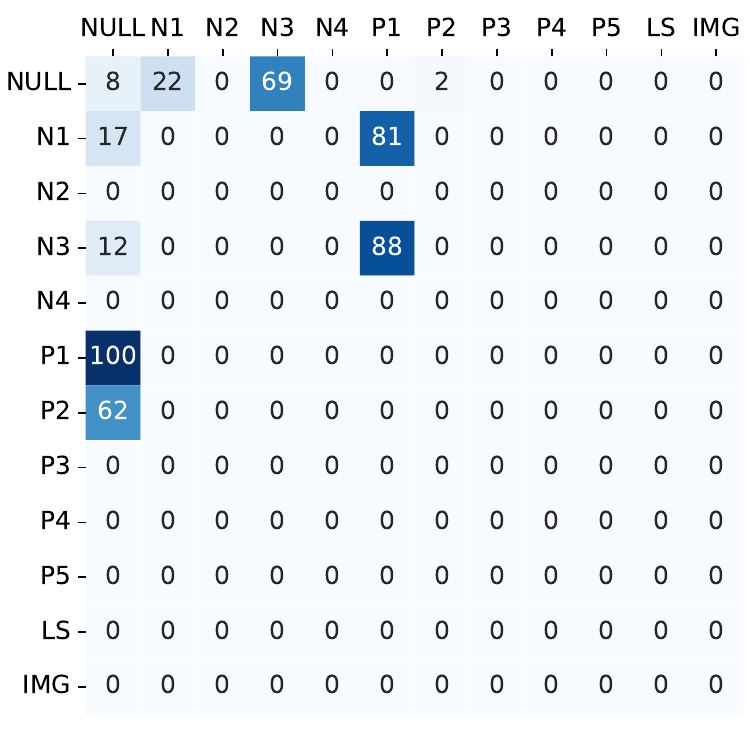}}
  \subfloat[\texttt{Square-Inf}]{\includegraphics[width=0.16\linewidth]{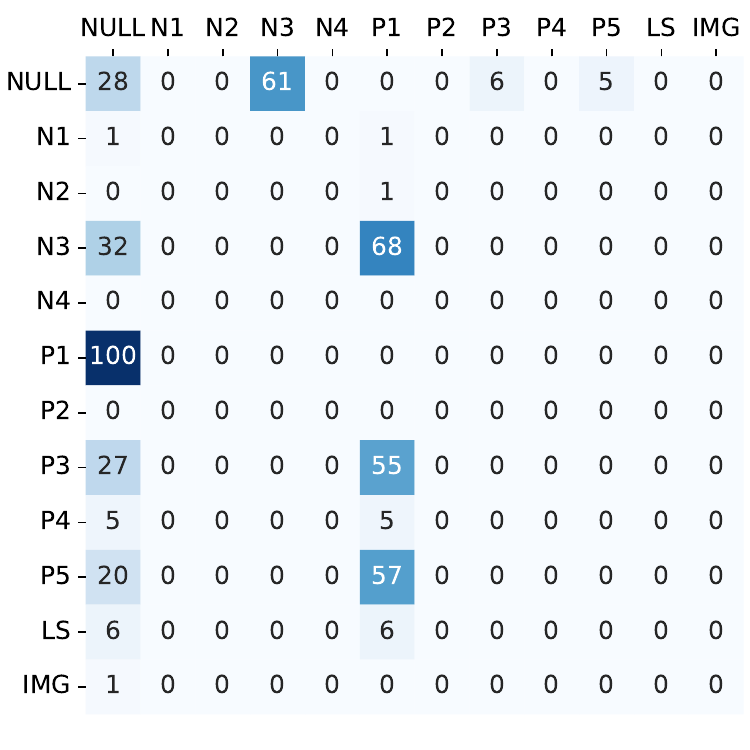}}

  \subfloat[\texttt{HSJ-2}]{\includegraphics[width=0.16\linewidth]{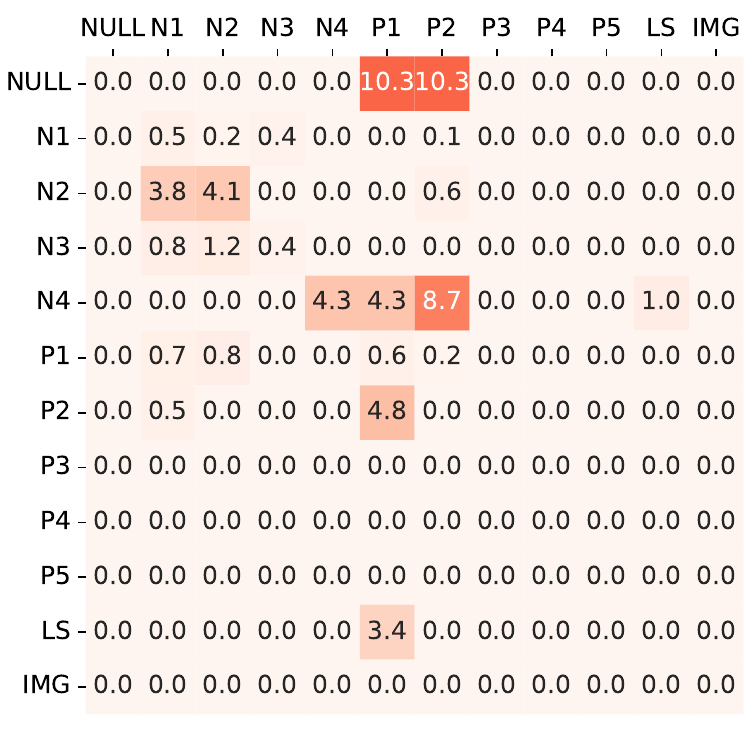}}
  \subfloat[\texttt{HSJ-Inf}]{\includegraphics[width=0.16\linewidth]{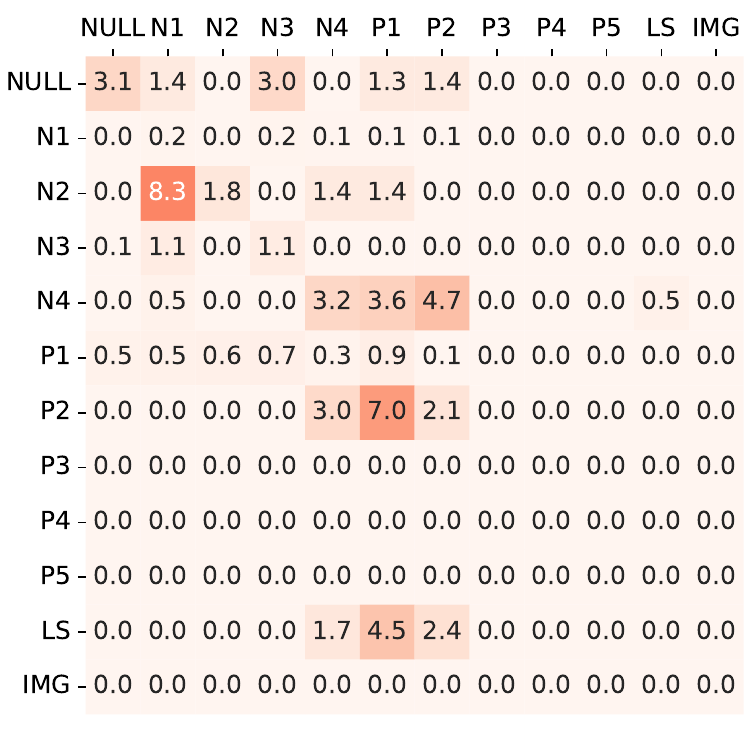}}
  \subfloat[\texttt{GeoDA-2}]{\includegraphics[width=0.16\linewidth]{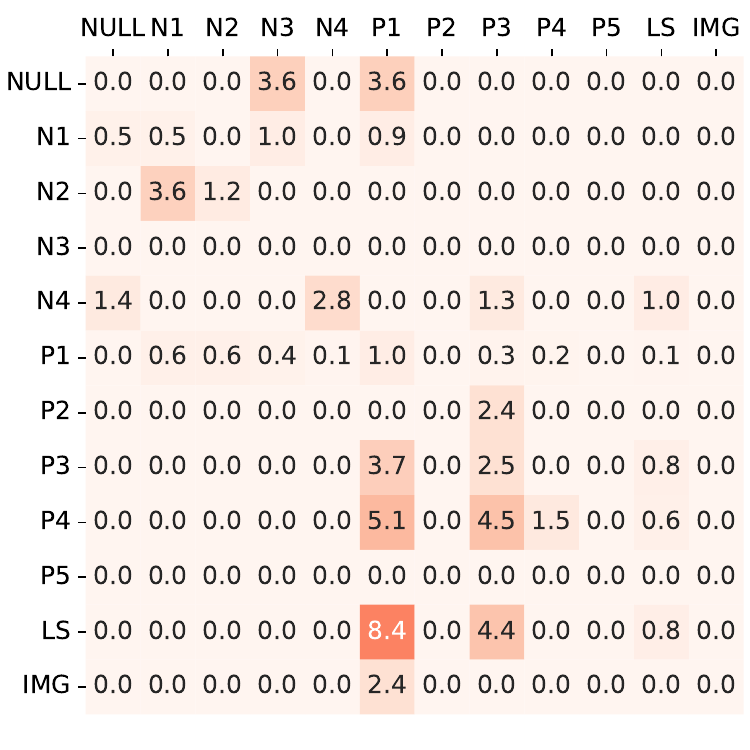}}
  \subfloat[\texttt{GeoDA-Inf}]{\includegraphics[width=0.16\linewidth]{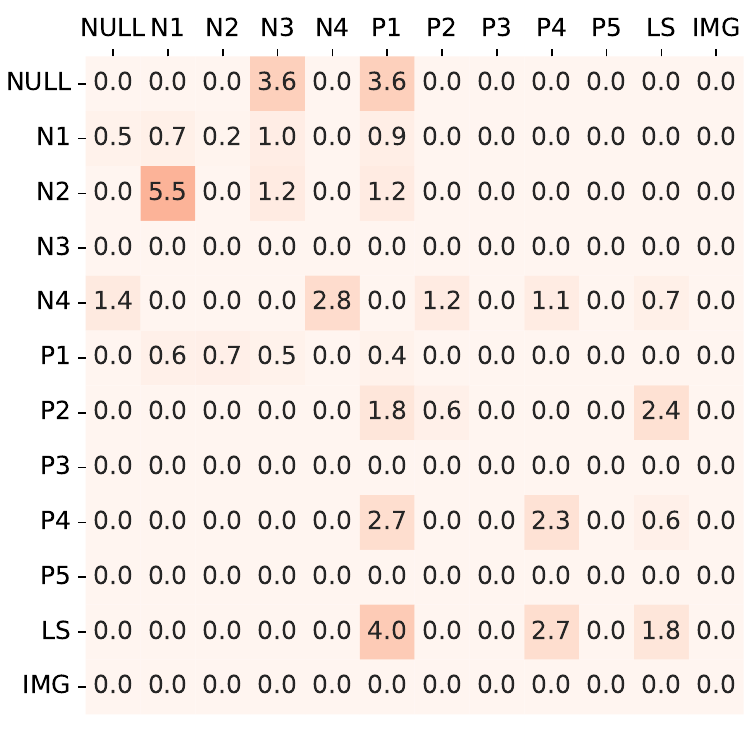}}
  \subfloat[\texttt{Square-2}]{\includegraphics[width=0.16\linewidth]{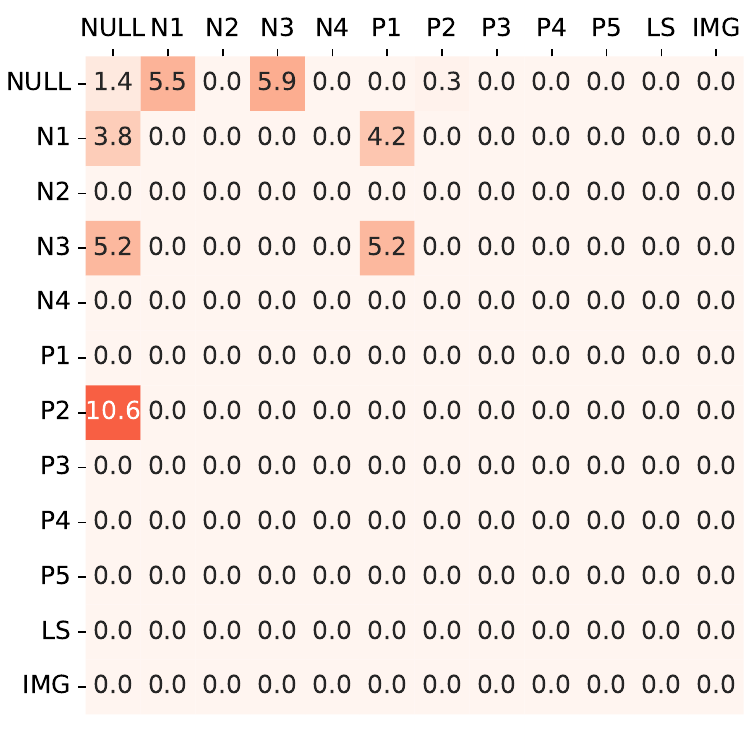}}
  \subfloat[\texttt{Square-Inf}]{\includegraphics[width=0.16\linewidth]{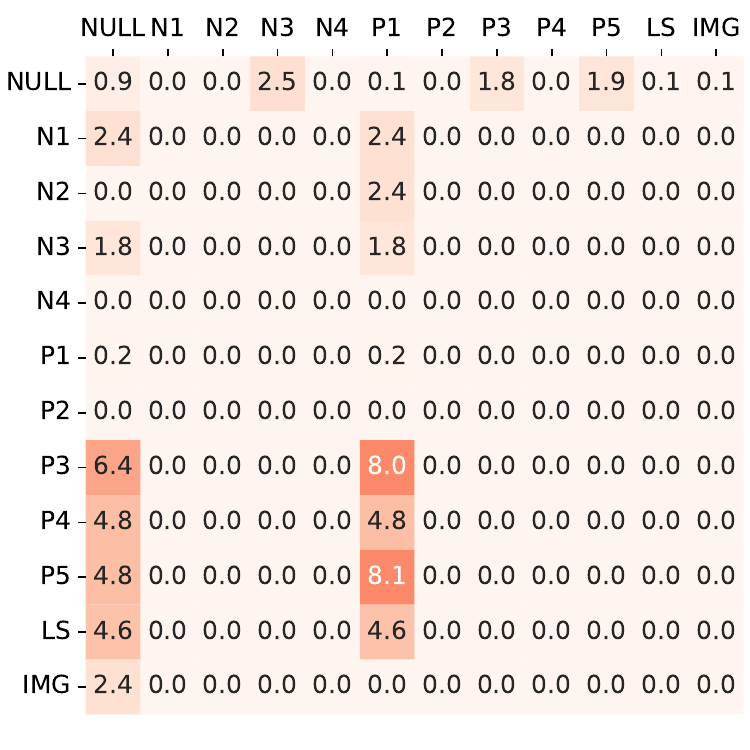}}

  \subfloat[\texttt{NES-2}]{\includegraphics[width=0.16\linewidth]{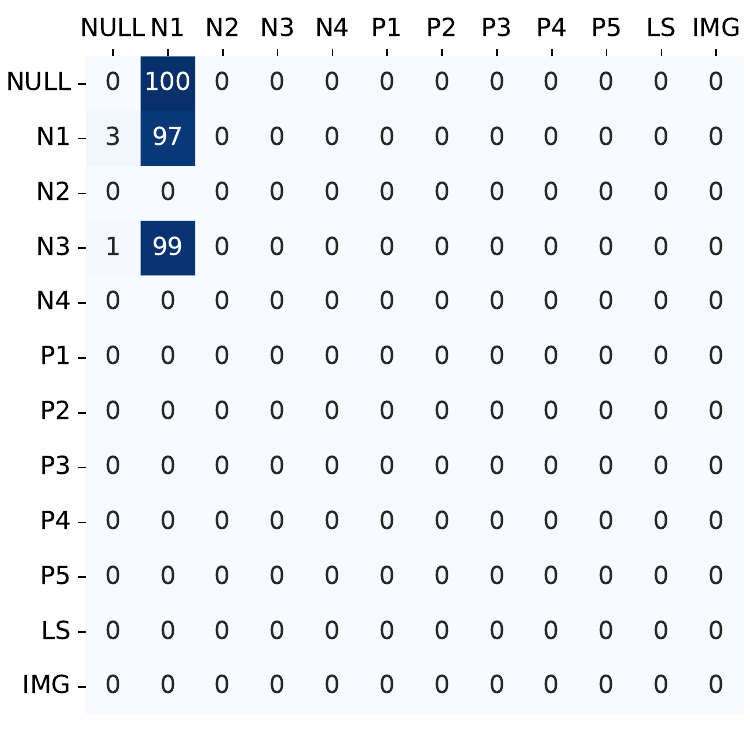}}
  \subfloat[\texttt{Boundary-2}]{\includegraphics[width=0.16\linewidth]{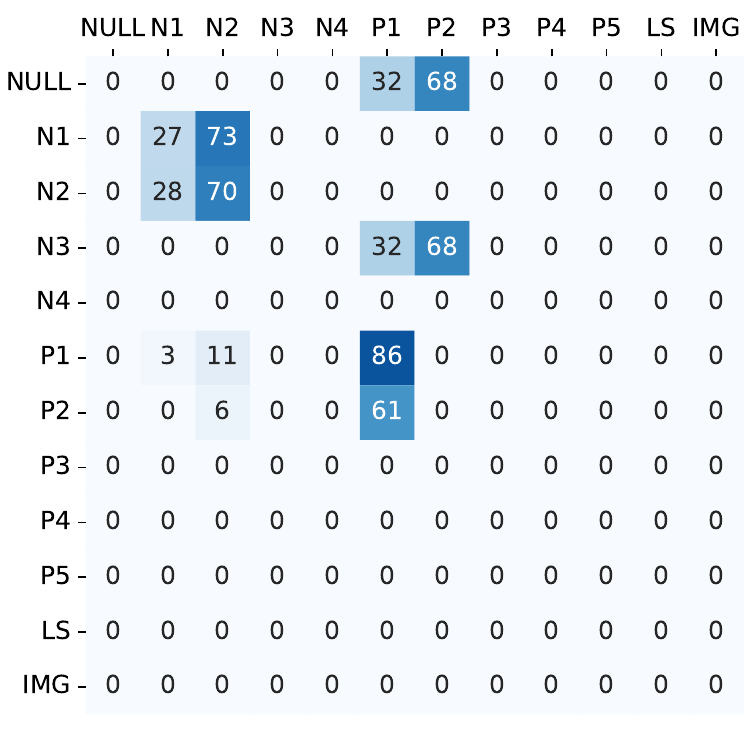}}
  \subfloat[\texttt{SignOPT-2}]{\includegraphics[width=0.16\linewidth]{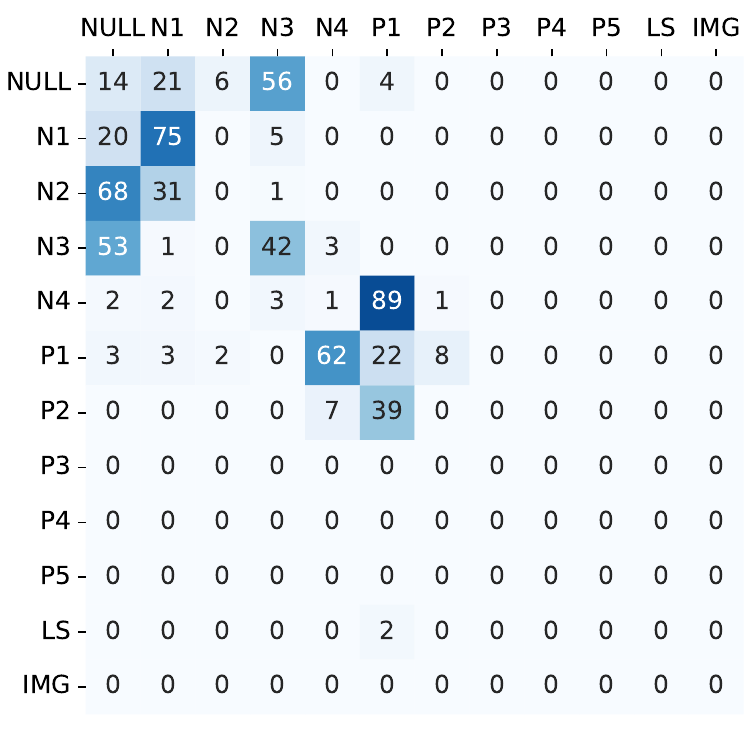}}
  \subfloat[\texttt{ECO-2}]{\includegraphics[width=0.16\linewidth]{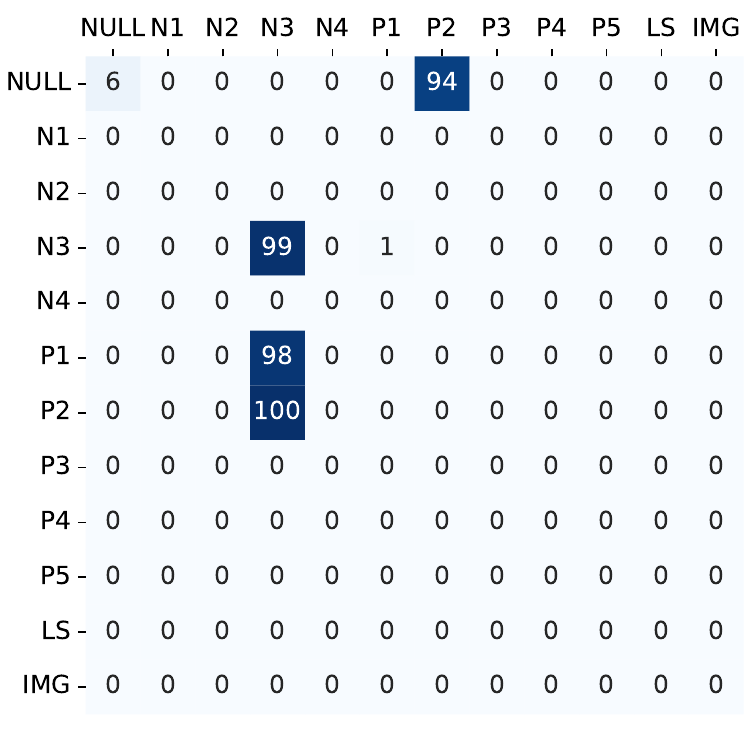}}
  \subfloat[\texttt{RayS-Inf}]{\includegraphics[width=0.16\linewidth]{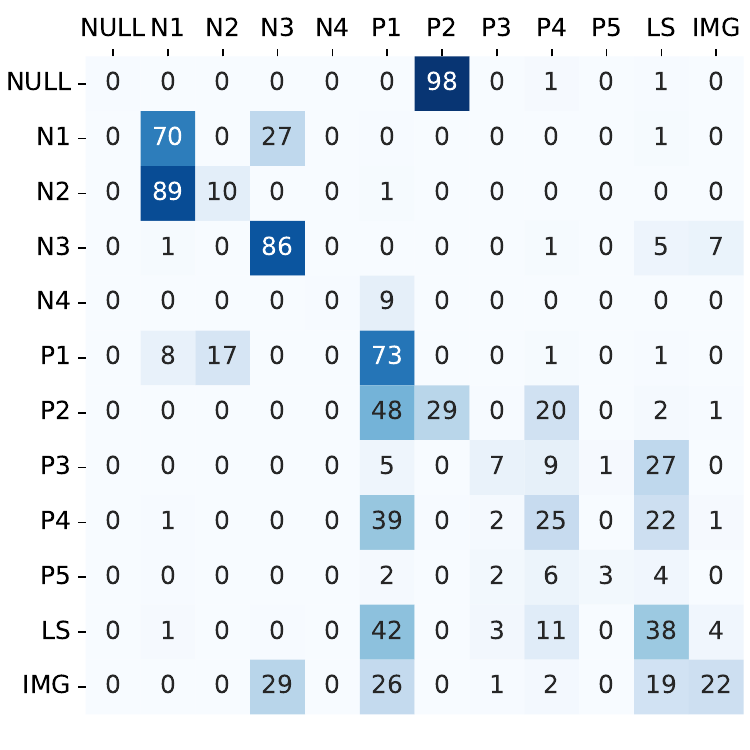}}

  \subfloat[\texttt{NES-2}]{\includegraphics[width=0.16\linewidth]{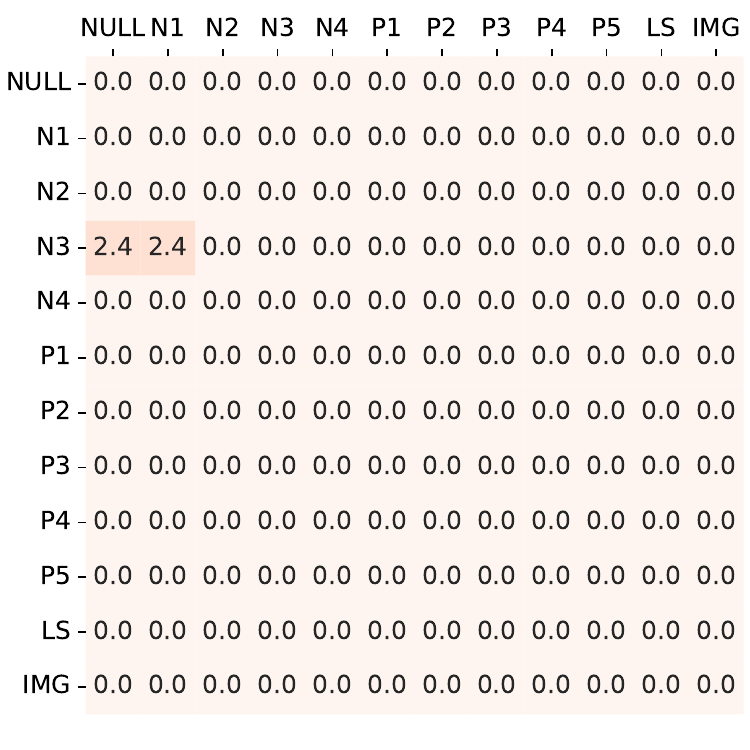}}
  \subfloat[\texttt{Boundary-2}]{\includegraphics[width=0.16\linewidth]{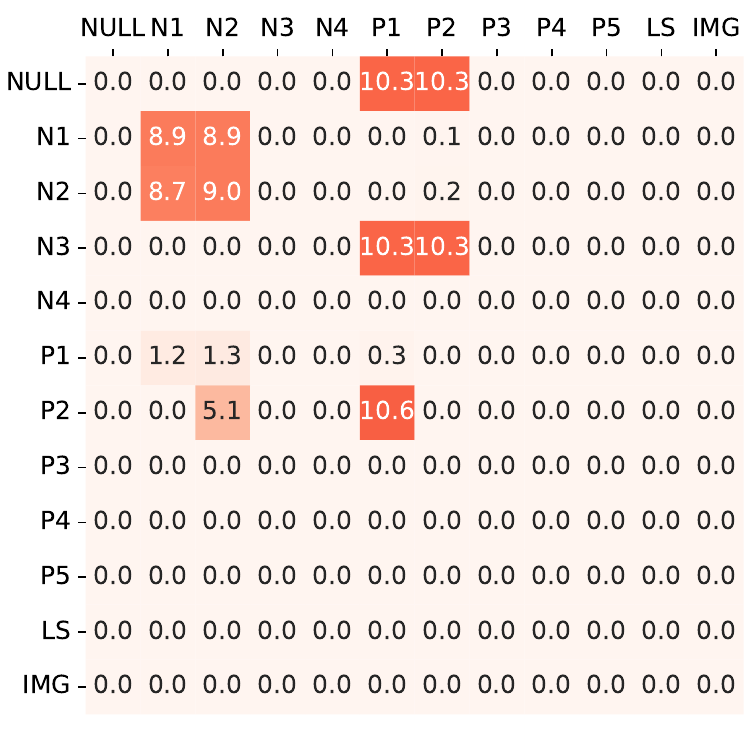}}
  \subfloat[\texttt{SignOPT-2}]{\includegraphics[width=0.16\linewidth]{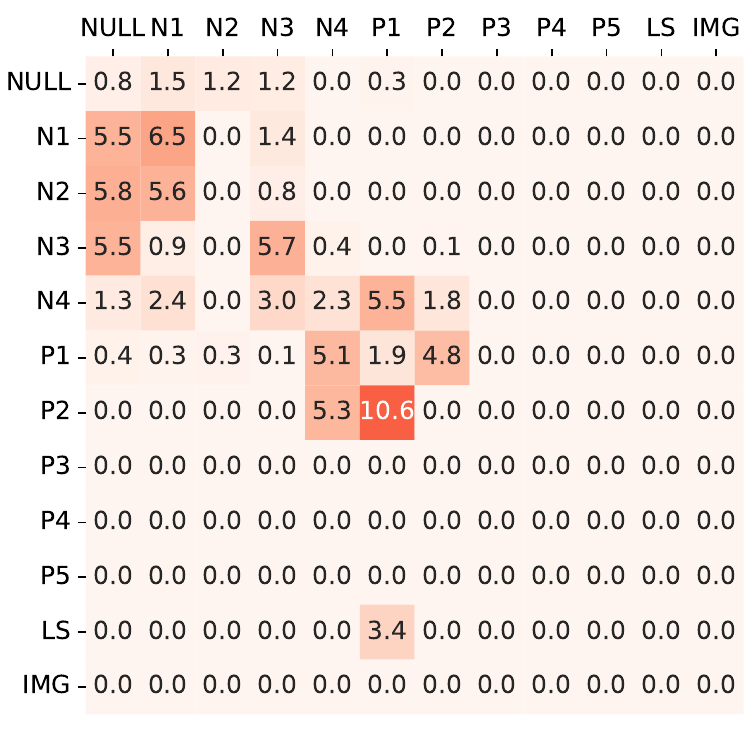}}
  \subfloat[\texttt{ECO-2}]{\includegraphics[width=0.16\linewidth]{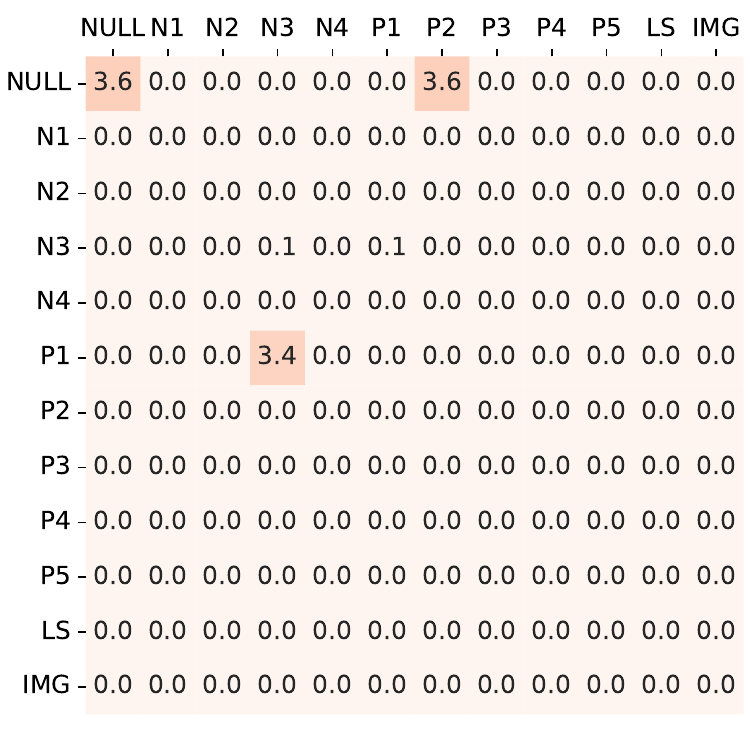}}
  \subfloat[\texttt{RayS-Inf}]{\includegraphics[width=0.16\linewidth]{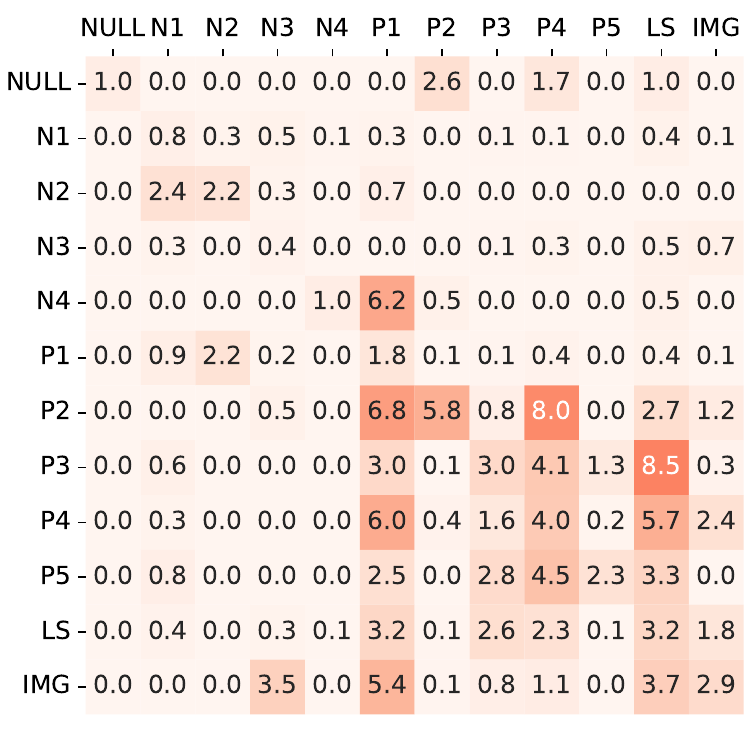}}

  \caption{The average (blue) and 95\% confidence interval (orange) of fingerprints for each textual and vision attack.}
  \label{fig:fp-all}
  \label{fig:fp-text-all}
\end{figure*}

We visualize the distribution of all fingerprints in \Cref{fig:fp-tsne} and their average and 95\% confidence interval for each attack in \Cref{fig:fp-text-all,fig:fp-all}.

%\newpage

\end{document}